%% file: main.tex
\documentclass[english]{wiley-article}
\usepackage{xcolor}
\usepackage{float}
\usepackage{textcomp}
\usepackage{bbding}
\usepackage{ifsym}
\usepackage{mathtools}
\usepackage{amsmath}
\usepackage{graphicx}
\usepackage[authoryear]{natbib}

\makeatletter

\providecommand{\tabularnewline}{\\}
\newcommand{\lyxdot}{.}

\usepackage{siunitx}

\title{A Kernel Stein Test for Comparing Latent Variable Models}

\author[1]{Heishiro Kanagawa}
\author[2,3]{Wittawat Jitkrittum}
\author[4]{Lester Mackey}
\author[5]{Kenji Fukumizu}
\author[1]{Arthur Gretton}

\affil[1]{Gatsby Computational Neuroscience Unit, University College London, London, UK}
\affil[2]{Max Planck Institute for Intelligent Systems, T{\"u}bingen, Germany}
\affil[3]{Google Research, New York, USA}
\affil[4]{Microsoft Research New England, Cambridge, MA, USA}
\affil[5]{The Institute of Statistical Mathematics, Tokyo, Japan}

\corraddress{Heishiro Kanagawa, Gatsby Computational Neuroscience Unit, University College London, London, UK}
\corremail{heishiro.kanagawa@gmail.com}

\runningauthor{Kanagawa et al.}

\usepackage{thmtools}
\usepackage{thm-restate}
\usepackage{physics}
\usepackage{xcolor}
\usepackage[ruled,vlined,linesnumbered]{algorithm2e}
\newtheorem{assumption}{Assumption}
\usepackage{lipsum}
\usepackage{capt-of}
\usepackage{wrapfig}
\usepackage{xr}
\newcommand{\headcell}{\cellcolor{black!20}}
\newcommand{\theadmd}[1]{\bfseries #1\rule[-1.2ex]{0pt}{2em}}

\@ifundefined{showcaptionsetup}{}{%
 \PassOptionsToPackage{caption=false}{subfig}}
\usepackage{subfig}
\makeatother

\usepackage{babel}
\begin{document}
\global\long\def\score#1{\mathbf{s}_{#1}}%

\global\long\def\hatscore#1{\hat{\mathbf{s}}_{#1}}%

\global\long\def\barscore#1{\bar{\mathbf{s}}_{#1}}%

\global\long\def\tildescore#1{\tilde{{\bf s}}_{#1}}%

\global\long\def\calX{\mathcal{X}}%

\global\long\def\calD{\mathcal{D}}%

\global\long\def\EE{\mathbb{E}}%

\global\long\def\var{\mathrm{Var}}%
\global\long\def\cov{\mathrm{Cov}}%

\global\long\def\calY{\mathcal{Y}}%
\global\long\def\calF{\mathcal{H}}%

\global\long\def\calZ{\mathcal{Z}}%
\global\long\def\calF{\mathcal{F}}%

\global\long\def\inner#1#2{\left\langle #1,#2\right\rangle }%

\global\long\def\la{\langle}%

\global\long\def\ra{\rangle}%
\global\long\def\mmd#1#2{\mathrm{MMD}(#1,#2)}%

\global\long\def\mmdsq#1#2{\mathrm{MMD}^{2}(#1,#2)}%

\global\long\def\mmdsqhat#1#2{\mathrm{\mathrm{\widehat{\mathrm{\mathrm{MMD}^{2}}}}}(#1,#2)}%

\global\long\def\norm#1{\left\Vert #1\right\Vert }%

\global\long\def\Verts#1{\lVert#1\rVert}%

\global\long\def\bignorm#1{\big\Vert#1\big\Vert}%

\global\long\def\Normal#1#2{\mathcal{N}(#1,#2)}%

\global\long\def\ksdhpm{\widehat{S_{p_{m}}^{2}}}%

\global\long\def\ksdhp{\widehat{S_{p}^{2}}}%

\global\long\def\ksdsqh#1#2{\widehat{\mathrm{\mathrm{KSD}}^{2}}\left(#1\Vert#2\right)}%

\global\long\def\ksdsq#1#2{\mathrm{KSD}^{2}\left(#1\Vert#2\right)}%

\global\long\def\ksd#1#2{\mathrm{KSD}\left(#1\Vert#2\right)}%

\global\long\def\dto{\overset{d}{\to}}%
\global\long\def\pto{\overset{p}{\to}}%

\global\long\def\parder#1{\frac{\partial}{\partial#1}}%

\global\long\def\diag{\mathrm{diag}}%

\global\long\def\Pr{\mathrm{Pr}}%

\global\long\def\austat{U_{n,m}}%
\global\long\def\ustat{U_{n}}%

\global\long\def\unmp{U_{n,m}^{(p)}}%
\global\long\def\unmq{U_{n,m}^{(q)}}%

\global\long\def\unp{U_{n}^{(p)}}%
\global\long\def\unq{U_{n}^{(q)}}%

\global\long\def\un#1{U_{n}^{(#1)}}%

\global\long\def\ksdpsq#1{S_{p}^{2}(#1)}%
\global\long\def\ksdqsq#1{S_{q}^{2}(#1)}%

\global\long\def\diff#1#2{#1-#2}%

\global\long\def\bfone{\mathbf{1}}%

\global\long\def\bfzero{\mathbf{0}}%

\global\long\def\data{R}%

\global\long\def\steinop#1{\mathcal{T_{{#1}}}}%

\global\long\def\iidsim{\overset{\mathrm{i.i.d.}}{\sim}}%

\global\long\def\indsim{\overset{\mathrm{ind.}}{\sim}}%

\begin{frontmatter} \maketitle
\begin{abstract}
We propose a kernel-based nonparametric test of relative goodness
of fit, where the goal is to compare two models, both of which may
have unobserved latent variables, such that the marginal distribution
of the observed variables is intractable. The proposed test generalizes
the recently proposed kernel Stein discrepancy (KSD) tests (Liu et
al., 2016, Chwialkowski et al., 2016, Yang et al., 2018) to the case
of latent variable models, a much more general class than the fully
observed models treated previously. The new test, with a properly
calibrated threshold, has a well-controlled type-I error. In the case
of certain models with low-dimensional latent structure and high-dimensional
observations, our test significantly outperforms the relative Maximum
Mean Discrepancy test, which is based on samples from the models and
does not exploit the latent structure.

\end{abstract}
\end{frontmatter}

\section{Introduction\label{sec:intro}}

A major approach to statistical modeling is the use of variables representing
quantities that are unobserved but thought to underlie the observed
data: well-known instances include probabilistic PCA \citep{Roweis1997EM,TipBis02},
factor analysis \citep[see, e.g.,][]{Basilevsky1994Statistical},
mixture models \citep[see, e.g.,][]{GilRicSpi95}, topic models for
text \citep{BleNgJor03}, and hidden Markov models (HMMs) \citep{Rabiner89}.
The hidden structure in these generative models serves multiple purposes:
it allows interpretability and understanding of model features (e.g.,
the topic proportions in a latent Dirichlet allocation (LDA) model
of text), and it facilitates modeling by leveraging simple low-dimensional
dynamics of phenomena observed in high dimensions (e.g., HMMs with
a low dimensional hidden state). Statistical modelers ultimately use
such models to reason about the data; thus, in order to guarantee
the validity of the inference, tools for comparing models and evaluating
model fit are required. 

This article addresses the problem of evaluating and comparing generative
probabilistic models, in cases where the models have a latent variable
structure, and the marginals over the observed data are intractable.
In this scenario, one strategy for evaluating a generative model is
to draw samples from it and to compare these samples to the modeled
data using a two-sample test: for instance, \citet{LloGha15} use
a test based on the maximum mean discrepancy (MMD) \citep{GreBorRasSchSmo2012}.
This approach has two disadvantages, however: it is not computationally
efficient due to the sampling step, and it does not take advantage
of the information that the model supplies, for instance the dependence
relations among the variables.

Recently, an alternative model evaluation strategy based on Stein's
method \citep{Stein1972bound,Chen1975Poisson,Stein1986Approximate,Barbour1988Steins,Goetze1991rate}
has been proposed, which directly employs a closed-form expression
for the unnormalized model. Stein's method is a technique from probability
theory developed to prove central limit theorems with explicit rates
of convergence \citep[see, e.g.,][]{ross2011}. The core of Stein's
method is that it characterizes a distribution with a \emph{Stein
operator}, which, when applied to a function, causes the expectation
of the function to be zero under the distribution. For our purposes,
we will use the result that a model-specific Stein operator may be
defined, to construct a measure of the model's discrepancy.  Notably,
Stein operators may be obtained without computing the normalizing
constant.

Stein operators have been used to design integral probability metrics
(IPMs) \citep{Mueller1997Integral} to test the goodness of fit of
models. IPMs specify a \emph{witness function} which has a large difference
in expectation under the sample and model, thereby revealing the difference
between the two. When a Stein operator is applied to the IPM function
class, the expectation under the model is zero, leaving only the expectation
under the sample. A Stein-modified $W^{2,\infty}$ Sobolev ball was
used as the witness function class in \citep{GorMac2015,gorham2016measuring}.
Subsequent work in \citep{ChwStrGre2016,LiuLeeJor2016,GorMac2017}
used as the witness function class a Stein-transformed reproducing
kernel Hilbert ball, as introduced by \citet{OatGirCho2017}: the
resulting goodness-of-fit statistic is known as the kernel Stein discrepancy
(KSD). Conditions for using the KSD in convergence detection were
obtained by \citet{GorMac2017}. While the foregoing work applies
in continuous domains, the approach may also be used for models on
a finite domain, where Stein operators \citep{RanTraAltBle16,YanLiuRaoNev18,BreNag19,ReiRos19,Hodgkinson2020,Shi2022}
and associated goodness-of-fit tests \citep{YanLiuRaoNev18} have
been established. Note that it is also possible to use Stein operators
to construct feature dictionaries for comparing models, rather than
using an IPM: examples include a test based on Stein features constructed
in the sample space so as to maximize test power \citep{JitXuSzaFukGre2017,JitKanPatHayetal2018}
and a test based on Stein-transformed random features \citep{HugMac2018}.
While the aforementioned tests address simple hypotheses, composite
tests that use Stein characterizations have been proposed for specific
parametric families including gamma \citep{HenzeMeintanisEbner2012Goodness,BetschEbner2019new}
and normal distributions \citep{BetschEbner2019Testing,HenzeVisagie2019Testing},
and general univariate parametric families \citep{BetschEbner2019Fixed}
(note that these tests are not based on IPMs). 

While testing goodness of fit alone may be desirable for models of
simple phenomena, it will often be the case that in complex domains,
no model will fit the data perfectly. In this setting, it is more
constructive to ask which model fits better, either within a class
of models or in comparing different model classes. A likelihood-ratio
test would be an ideal choice for this task, since it is uniformly
most powerful \citep{LehRom05}. If the models contain latent variables,
however, a likelihood-ratio test requires evaluating marginal densities
of the models, which are typically intractable. A number of Monte
Carlo techniques have been developed to estimate marginal densities
or log-density ratios \citep[see, e.g.,][for a review]{Friel_2012}.
Constructing a test with such techniques is challenging, however;
e.g., estimating each marginal density induces a bias in the likelihood
ratio, which is difficult to characterize when designing a calibrated
threshold (see Section \ref{subsec:Challenges-with-likelihood-ratio}
for a detailed discussion). Addressing the intractability of the likelihood,
\citet{BouBelBlaAntGre2016} proposed a purely sample-based relative
goodness of fit test, which compares maximum mean discrepancies between
the samples from two rival models with a reference real-world sample.
A second relative test was proposed by \citet{JitKanPatHayetal2018},
generalizing \citet{JitXuSzaFukGre2017} and learning the Stein features
for which each model outperforms the other; this test requires marginal
densities up to normalizing constants, and does not apply to latent
variable models. 

In the present work, we introduce a novel relative goodness-of-fit
test for latent variable models (LVMs), which compares models by computing
approximate kernel Stein discrepancies. Our contribution is to provide
a frequentist test of relative goodness of fit, with an approximate
U-statistic of the kernel Stein discrepancy difference as our test
statistic. The statistic is expressed by a posterior expectation of
the latent given an observation, and is amenable to standard Markov
Chain Monte Carlo techniques: in particular, it does not suffer from
the challenges in characterizing bias observed in likelihood-ratio
estimates for LVMs. Note that our approach differs from Bayesian model
selection  \citep{Jeffreys1961Theory,Schwarz1978Estimatinga,KassRaftery1995Bayes,Watanabe2013widely},
which reports posterior odds (or Bayes factors) and does not concern
controlling frequentist risks such as type-I error rates. To the best
of our knowledge, our test represents the first general-purpose, frequentist,
relative test for latent variable models. 

We recall the Stein operator and kernel Stein discrepancy in Section
\ref{sec:Background}, and the notion of relative tests in Section
\ref{sec:Main-Results}. Our main theoretical contributions, also
in Section \ref{sec:Main-Results}, are two-fold: first, we derive
an appropriate test threshold to account for the randomness in the
test statistic caused by sampling the latent variables. Second, we
provide guarantees that the resulting test has the correct Type-I
level (i.e., that the rate of false positives is properly controlled)
and that the test is consistent under the alternative: the number
of false negatives drops to zero as we observe more data. Finally,
in Section \ref{sec:Experiments}, we demonstrate our relative test
of goodness-of-fit on a variety of latent variable models. Our main
point of comparison is the relative MMD test \citep{BouBelBlaAntGre2016},
where we sample from each model. We demonstrate that the relative
Stein test outperforms the relative MMD test in the particular case
where the low dimensional structure of the latent variables can be
exploited.

\section{The kernel Stein discrepancy and latent variable models \label{sec:Background}}

In this section, we recall the definition of the Stein operator as
used in goodness-of-fit testing, as well as the kernel Stein discrepancy,
a measure of goodness-of-fit based on this operator. We will then
introduce latent variable models, which will bring us to the setting
of relative goodness of fit with competing models in Section \ref{sec:Main-Results}. 

Before proceeding, we call attention to our setting: in this article,
we treat both continuous- and discrete-valued observations, as formally
defined at the outset of Section \ref{subsec:background-ksd}. It
is our intention to study these two data modalities as they admit
the same treatment. The subsequent definitions and analysis of our
test are independent of whether a continuous or discrete Stein operator
is used, apart from experiments concerning discrete-valued observations.
Thus, the detail about discrete models in Section \ref{subsec:background-ksd}
may be initially skipped if desired.

\subsection{Stein operators and kernel Stein discrepancies \label{subsec:background-ksd}}

Let $\calX$ be the space in which the data takes values; for $D\geq1$,
the space $\mathcal{X}$ is either the Euclidean space $\mathbb{R}^{D}$
or a finite lattice $\{0,\dots,L-1\}^{D}$ for some $L>1$. Depending
on $\mathcal{X}$, we shall assume that the densities below are all
defined with respect to the Lebesgue measure or the counting measure;
i.e., the term \textit{density} includes probability mass functions
(pmfs). 

\paragraph{Continuous-valued observations.}

Suppose that we are given data $\{x_{i}\}_{i=1}^{n}\overset{\text{i.i.d.}}{\sim}\data$
from an unknown distribution $\data$, and we wish to test the goodness
of fit of a model $P$. We first consider the case where the probability
distributions $P,\data$ are defined on $\mathbb{R}^{D}$ and have
respective probability densities $p,r$, where all density functions
considered in this paper are assumed strictly positive and continuously
differentiable. We treat the case of densities defined on bounded
domains in the supplement, Section \ref{subsec:bounded-domain}. For
differentiable density functions, we define the \emph{score} \emph{function},
\[
\score p(x)\in\mathbb{R}^{D}\coloneqq\frac{\nabla p(x)}{p(x)}=\nabla\log p(x),
\]
where the gradient operator is $\nabla:=\left[\frac{\partial}{\partial x^{1}},\hdots,\frac{\partial}{\partial x^{D}}\right]^{\top}.$
The score is independent of the normalizing constant for $p$, making
it computable even when $p$ is known only up to normalization. Using
this score, we define the  \emph{Langevin Stein operator} on a space
$\mathcal{F}$ of differentiable functions from $\mathbb{R}^{D}$
to $\mathbb{R}^{D}$ \citep{GorMac2015,OatGirCho2017},
\[
\left[{\cal A}_{P}f\right](x)=\left\langle \score p(x),f(x)\right\rangle +\left\langle \nabla,f(x)\right\rangle ,\quad f\in\mathcal{F}.
\]

A kernel discrepancy may be defined based on the Stein operator \citep{ChwStrGre2016,LiuLeeJor2016,GorMac2017},
which allows us to measure the departure of a distribution $\data$
from a model $P$. We define $\mathcal{F}$ to be a space comprised
of $D$-dimensional vectors of functions $f=(f_{1},\ldots f_{D})$
where the $d$-th function $f_{d}$ is in a reproducing kernel Hilbert
space (RKHS) \citep[Definition 4.18]{aronszajnTheoryReproducingKernels1950,Steinwart2008}
with a positive definite kernel $k(\cdot,\cdot):\calX\times{\cal X}\to\mathbb{R}$
(we use the same kernel for each dimension). The inner product on
$\mathcal{F}$ is $\left\langle f,g\right\rangle _{\mathcal{F}}:=\sum_{d=1}^{D}\left\langle f_{d},g_{d}\right\rangle _{\mathcal{F}_{k}},$
and $\mathcal{F}_{k}$ denotes an RKHS of real-valued functions with
kernel $k$.

The (Langevin) \emph{kernel Stein discrepancy }(KSD) between $P$
and $\data$ is defined as 
\begin{equation}
\ksd PR=\sup_{\left\Vert f\right\Vert _{\calF}\le1}|\EE_{x\sim\data}{\cal A}_{P}f(x)-\EE_{y\sim P}\mathcal{A}_{P}f(y)|.\label{eq:ksd-ipm-def}
\end{equation}
Under appropriate conditions on the kernel and measure $P$, the expectation
$\mathbb{E}_{y\sim P}{\cal A}_{P}f(y)=0$ for any $f\in\mathcal{F}.$
To ensure this property, we will require that $k\in C^{(1,1)}$, the
set of continuous functions on ${\cal \calX\times\calX}$ with continuous
first derivatives and that $\mathbb{E}_{y\sim P}\left[\left\Vert \mathbf{s}_{p}(y)\right\Vert _{2}\right]<\infty$
with $\lVert\cdot\rVert_{2}$ the Euclidean norm. We further assume
that the following tail condition holds outside a bounded set : $p(x)\sqrt{k(x,x)}\leq C\Verts x_{2}^{\delta}$
for some constants $C>0$ and $\delta>D-1$ \citep[see the clarification by][p.12, on the tail condition for the Stein's identity]{South_2021}.
\textcolor{red}{} With the vanishing expectation $\mathbb{E}_{y\sim P}{\cal A}_{P}f(y)=0,$
the KSD reduces to $\ksd PR=\sup_{\left\Vert f\right\Vert _{\calF}\le1}|\EE_{x\sim\data}{\cal A}_{P}f(x)|$
. The use of an RKHS as the function class yields a closed form expression
of the discrepancy by the kernel trick \citep[Proposition 2]{ChwStrGre2016,GorMac2017},
\begin{align*}
\ksdsq P{\data}=\EE_{x,x'\sim\data\otimes\data}[h_{p}(x,x')],
\end{align*}
if $\EE_{x\sim\data}[h_{p}(x,x)^{1/2}]<\infty.$ Here, the symbol
$\data\otimes\data$ denotes the product measure of two copies of
$\data$ (so $x$ and $x'$ are independent random variables identically
distributed with the law $\data$). The function $h_{p}$ (called
a \emph{Stein kernel}) is expressed in terms of the RKHS kernel $k$
and the score function $\score p$, 
\begin{align*}
h_{p}(x,x') & =\score p(x)^{\top}\score p(x')k(x,x')+\score p(x)^{\top}k_{1}(x',x)+\score p(x')^{\top}k_{1}(x,x')+k_{12}(x,x'),
\end{align*}
where we have defined 
\begin{align*}
k_{1}(a,b) & \coloneqq\nabla_{x}k(x,x')\vert_{x=a,x'=b},\\
k_{12}(a,b) & \coloneqq\nabla_{x}^{\top}\nabla_{x'}k(x,x')\vert_{x=a,x'=b}.
\end{align*}
 For a given i.i.d. sample $\{x_{i}\}_{i=1}^{n}\sim\data$, the discrepancy
has a simple closed-form finite sample estimate,
\begin{equation}
\ksdsq P{\data}\approx\frac{1}{n(n-1)}\sum_{i\neq j}h_{p}(x_{i},x_{j}),\label{eq:plain_ksd_ustat}
\end{equation}
which is a U-statistic \citep{hoeffding_class_1948}. When the kernel
is integrally strictly positive definite (ISPD) \citep[Section 2]{SriFukLan2011},
and $\data$ admits a density $r$ that satisfies $\mathbb{E}_{x\sim\data}\left\Vert \nabla\log\bigl(p(x)/r(x)\bigr)\right\Vert _{2}<\infty$,
we have that $\ksd P{\data}=0$ iff $P=\data$ \citep[Proposition 1]{BarpBriolDuncanEtAl2019Minimum}.
 The earlier results of \citet{ChwStrGre2016} and \citet{LiuLeeJor2016}
require more stringent integrability conditions. \citet[Theorem 7]{GorMac2017}
showed that KSD can distinguish any Borel measure $R$ from $P$ by
assuming conditions such as distant dissipativity (satisfied by finite
Gaussian mixtures) \citep[Section 3]{gorham2016measuring}. However,
such conditions may be difficult to validate for latent variable models.
Thus, hereafter, we assume the former condition on the data distribution
$R.$

\paragraph{Discrete-valued observations.}

We next recall the kernel Stein discrepancy in the discrete setting
where distributions are defined on $\mathcal{X}=\{0,\dots,L-1\}^{D}$
with $L>1$, as introduced by \citet{YanLiuRaoNev18}. In place of
derivatives, we specify $\Delta_{k}$ as the cyclic forward difference
w.r.t. $k$-th coordinate: $\Delta_{k}f(x)=f(x^{1},\dots,\tilde{x}^{k},\dots,x^{D})-f(x^{1},\dots,x^{k},\dots,x^{D})$
where $\tilde{x}^{k}=x^{k}+1\mod L$, with the corresponding vector-valued
operator $\Delta=(\Delta_{1},\dots,\Delta_{D})$. The inverse operator
$\Delta_{k}^{-1}$ is given by the backward difference $\Delta_{k}^{-1}f(x)=f(x^{1},\dots,x^{k},\dots,x^{D})-f(x^{1},\dots,\bar{x}^{k},\dots,x^{D})$,
where $\bar{x}^{k}=x^{k}-1\mod L$, and $\Delta^{-1}=(\Delta_{1}^{-1},\dots,\Delta_{D}^{-1})$.
The score is then $\score p(x)\coloneqq p(x)^{-1}\Delta p(x),$ where
it is assumed that the pmf is strictly positive (i.e., it is never
zero). The difference Stein operator is then defined as ${\cal A}_{P}f(x)=\mathrm{tr}\left[f(x)\score p(x)^{\top}+\Delta^{-1}f(x)\right]$,
where it can be shown that $\mathbb{E}_{x\sim P}[{\cal A}_{P}f(x)]=0$
\citep[Theorem 2]{YanLiuRaoNev18} (note that we include a trace for
consistency with the continuous case--this does not affect the test
statistic \citep[Eq. 10]{YanLiuRaoNev18}). We have defined the Stein
operator and the score function slightly differently from \citet{YanLiuRaoNev18};
the change is only in their signs, but this results in the same discrepancy.
The difference Stein operator is not the only allowable Stein operator
on discrete spaces: other alternatives are given by \citet[Theorem 3]{YanLiuRaoNev18},
\citet{Hodgkinson2020}, and \citet{Shi2022}. Although we focus on
the Stein operator above, in practice, one might want to consider
different Stein operators depending on the application. For instance,
the score function $\score p$ can be numerically unstable, as it
contains the reciprocal $1/p(x);$ this can occur when the support
of the model is severely mismatched to that of the data. In this particular
case, one might choose the Barker-Stein operator proposed by \citet{Shi2022},
an instance of the Zanella-Stein operator of \citet[Example 2]{Hodgkinson2020}.
See Appendix \ref{subsec:stable-discrete-stein} for details. We compare
this operator to the difference operator in an experiment where this
mismatch occurs (Section \ref{subsec:arxiv}). 

As in the continuous case, the KSD can be defined as an IPM, given
a suitable choice of reproducing kernel Hilbert space for the discrete
domain. An example of kernel is the exponentiated Hamming kernel,
$k(x,x')=\exp\left(-d_{H}(x,x')\right)$, where $d_{H}(x,x')=D^{-1}\sum_{d=1}^{D}\mathbb{I}(x^{d}\neq x'^{d})$.
The population KSD is again given by the expectation of the Stein
kernel, $\ksdsq P{\data}=\EE_{(x,x')\sim\data\otimes\data}[h_{p}(x,x')]$,
where $h_{p}$ is defined as 
\[
h_{p}(x,x')=\score p(x)^{\top}\score p(x')k(x,x')+\score p(x)^{\top}k_{1}(x',x)+\score p(x')^{\top}k_{1}(x,x')+k_{12}(x,x'),
\]
  and the kernel gradient is replaced by the inverse difference operator,
e.g., $k_{1}(x,x')=\Delta_{x}^{-1}k(x,x')$, where $\Delta_{x}^{-1}$
indicates that the operator $\Delta^{-1}$ is applied with respect
to the argument $x$. From \citet[Lemma 8]{YanLiuRaoNev18}, we have
that $\ksd P{\data}=0$ iff $P=\data$, under the conditions that
the probability mass functions for $P$ and $\data$ are positive
and that the Gram matrix defined over all the configurations in $\mathcal{X}$
is strictly positive definite (i.e., the kernel is integrally strictly
positive definite). One can define a kernel satisfying the required
condition, for example, by embedding ${\cal X}$ into $\mathbb{R}^{L\times D}$
with one-hot encoding and using a Taylor-type kernel such as the exponentiated
quadratic kernel \citep[Theorem 2.2]{ChristmannSteinwart2010Universal}. 

\subsection{Kernel Stein discrepancies of latent variable models \label{subsec:background-ksd-lvms}}

Our objective is to use the KSD to evaluate latent variable models,
and here we formally specify our target model class. Let $\mathcal{\mathcal{L}}_{{\cal X}|\calZ}=\{p(\cdot|z):z\in\mathcal{Z}\}$
be a family of probability density functions on $\calX$ (we call
these \emph{likelihood} functions), which are indexed by elements
of a set $\mathcal{Z}$. A latent variable model $P$ is specified
by such a family $\mathcal{\mathcal{L}}_{\calX|\calZ}$ and a (prior)
probability measure $P_{Z}$ over $\mathcal{Z}$. The combination
of these defines the marginal density function $p(x)=\int p(x|z)\dd P_{Z}(z)$
and the posterior distribution $P_{Z}(\dd z|x)=\{p(x|z)/p(x)\}P_{Z}(\dd z);$
The distribution $P$ induced by the former acts as a model of the
distribution $R$ underlying the observation, and the latter enables
us to draw an inference over the unobserved variable. 

\begin{remark}
In our notation, the variable $z$ can represent multiple latent variables.
The likelihood $p(x|z)$ often contains parameters, but the dependency
on these is suppressed here. If a prior is defined on a parameter,
we may treat it as a latent variable; this consideration is relevant
to predictive distributions. The likelihood and the prior in a model
may be conditioned on some fixed data (e.g., they can be posterior
predictive distributions), which we require to be independent of the
data used for testing -- in such a case, we omit the dependency on
the held-out data. For examples, we refer the reader to Section \ref{sec:Experiments}. 

\end{remark}

The definition of the KSD remains the same for latent variable models,
but an additional difficulty arises in its estimation. Unfortunately
the U-statistic estimator given in \eqref{eq:plain_ksd_ustat} requires
the score function of the marginal $p,$ which is challenging to obtain
due to the intractability of marginalizing out the latent variable.
We will address this challenge by rewriting the score function in
terms of the posterior distribution of the latent. In the following,
we focus on continuous variable models, but the same conclusion holds
for discrete counterparts by replacing gradient operation with cyclic
differences. 

Under a regularity condition, the score function can be expressed
as 
\begin{align}
\score p(x) & =\EE_{z|x}[\score p(x|z)],\label{eq:score_formula}
\end{align}
where $\score p(x|z)$ is the score function of the conditional $p(x|z)$;
i.e., $\score p(x|z)=p(x|z)^{-1}\nabla_{x}p(x|z)$ for continuous-valued
$x.$ The reasoning is as follows: 
\begin{align*}
\frac{\nabla_{x}p(x)}{p(x)} & =\frac{1}{p(x)}\int\nabla_{x}p(x|z)\dd P_{Z}(z)\\
 & =\int\frac{\nabla_{x}p(x|z)}{p(x|z)}\cdot\frac{p(x|z)\dd P_{Z}(z)}{p(x)}=\EE_{z|x}[\score p(x|z)],
\end{align*}
where we have assumed the exchangeability of differentiation and integration:
$\nabla_{x}p(x)=\int\nabla_{x}p(x|z)\dd P_{Z}(z)$. The identity
\eqref{eq:score_formula} is an analogue of Fisher's identity \citep{Fisher1925Theory,DempsterLairdRubin1977Maximum},
which pertinently formed the basis for Stein control variate methodology
in \citep{FrielMiraOates2016Exploiting}, parameter inference for
doubly-intractable models via score matching \citep{Vertes2016},
and Bayesian model selection with Hyv{\"a}rinen score \citep{DawidMusio2015Bayesian,ShaoJacobDingEtAl2019Bayesian}.
Note that the conditional score $\score p(x|z)$ is typically possible
to evaluate. For example, consider the following simple form of an
exponential family density $p(x|z)\propto\exp(T(x)\eta(z))$ defined
on $\mathbb{R}^{D}$ with $T(x):\mathbb{R}^{D}\to\mathbb{R}$ and
$\eta:\mathcal{Z}\to\mathbb{R}$; for this density, $\score p(x|z)=\eta(z)\nabla_{x}T(x).$
As can be seen in this example, the formula \eqref{eq:score_formula}
does not require the likelihood $p(x|z)$ to be normalized. This feature
eliminates the need for estimating the normalizing constant of $p(x|z)$
for each $z,$ which is required to compute goodness-of-fit measures
based on the marginal density $p(x)$ \citep{Friel_2012}; Section
\ref{subsec:boundedPPCA} in the supplementary
presents a use case with a truncated model.

With this identity, the KSD is rewritten as follows.

\begin{lemma}\label{lem:ksdnewform}Let 
\begin{align}
\begin{aligned}H_{p}[(x,z),(x',z')] & =\score p(x|z)^{\top}\score p(x'|z')k(x,x')+\score p(x|z)^{\top}k_{1}(x',x)\\
 & \hphantom{=\ }+k_{1}(x,x')^{\top}\score p(x'|z')+k_{12}(x,x').
\end{aligned}
\label{eq:lsteinkernel}
\end{align}
 Assume $\EE_{(x,z),(x',z')\sim\tilde{\data}\otimes\tilde{\data}}\lvert H_{p}[(x,z),(x',z')]\rvert<\infty$
with the joint distribution $\tilde{R}(\dd(x,z))=P_{Z}(\dd z|x)R(\dd x).$
If the formula \eqref{eq:score_formula} holds, then, 
\begin{align*}
\ksdsq P{\data}= & \EE_{(x,z),(x',z')\sim\tilde{\data}}{}_{\otimes\tilde{\data}}H_{p}[(x,z),(x',z')].
\end{align*}
\end{lemma}

\begin{proof}

Substituting the formula \eqref{eq:score_formula} in the definition
of KSD gives the required equation by the Tonelli-Fubini theorem.\qed

\end{proof}

\begin{remark}\label{rem:integrability}

The integrability assumption holds trivially if the input space $\mathcal{X}$
is finite, while care needs to be taken otherwise. The condition can
be checked by examining the absolute integrability of each term in
\eqref{eq:lsteinkernel}. The integrability assumption on the fourth
term is mild, and is satisfied by common kernels, e.g., the exponentiated
quadratic or the inverse multi-quadratic kernels. The condition on
the other terms needs to be checked on a model-by-model basis. It
can be shown that the example models in Section \ref{sec:Experiments}
satisfy the assumption (please see Section \ref{subsec:Integrability-condition-in}
in the supplementary material for details).

\end{remark}

The new KSD expression is an expectation of a computable symmetric
kernel, and constructing an unbiased estimate is straightforward once
we obtain a sample. In practice, when the model is complex, sampling
from the posterior distribution generally requires simulation, as
the posterior is not available in closed form. Therefore, we propose
to approximate the expectation by Markov Chain Monte Carlo (MCMC)
methods and construct an approximate U-statistic estimator as follows.
Let $\mathbf{z}_{i}^{(t)}=\bigl(z_{i,1}^{(t)},\cdots,z_{i,m}^{(t)}\bigr)\in\calZ^{m}$
be a latent sample of size $m$ drawn by an MCMC method having $P_{Z}(\cdot|x_{i})$
as its invariant measure after $t$ burn-in iterations. Let $\barscore p(x_{i}|\mathbf{z}_{i}^{(t)})=\frac{1}{m}\sum_{j=1}^{m}\score p(x_{i}|z_{i,j}^{(t)})$.
Given a joint sample $\bigl\{\bigl(x_{i},\mathbf{z}_{i}^{(t)}\bigr)\bigr\}_{i=1}^{n},$
we estimate the KSD by 
\begin{align}
U_{n}^{(t)}(P)\coloneqq & \frac{1}{n(n-1)}\sum_{i\neq j}\bar{H}_{p}\bigl[\bigl(x_{i},\mathbf{z}_{i}^{(t)}\bigr),\bigl(x_{j},\mathbf{z}_{j}^{(t)}\bigr)\bigr],\label{eq:mcmc_ksd_ustat}
\end{align}
where 
\begin{align*}
\bar{H}_{p}\bigl[\bigl(x_{i},\mathbf{z}_{i}^{(t)}\bigr),\bigl(x_{j},\mathbf{z}_{j}^{(t)}\bigr)\bigr] & =\barscore p(x_{i}|\mathbf{z}_{i}^{(t)})^{\top}\barscore p(x_{j}|\mathbf{z}_{j}^{(t)})k(x_{i},x_{j})+\barscore p(x_{i}|\mathbf{z}_{i}^{(t)})^{\top}k_{1}(x_{j},x_{i})\\
 & \hphantom{=}\ +k_{1}(x_{i},x_{j})^{\top}\barscore p(x_{j}|\mathbf{z}_{j}^{(t)})+k_{12}(x_{i},x_{j}),
\end{align*}
and the sum is taken over all distinct sample pairs. If $P_{Z}^{(t)}(\dd\mathbf{z}|x)$
denotes the distribution of an MCMC sample $\mathbf{z}^{(t)}=(z_{1}^{(t)},\dots,z_{m}^{(t)}),$
then this estimator is indeed a U-statistic, but its expectation is
that of kernel $\bar{H}_{p}$ with respect to $P_{Z}^{(t)}(\dd\mathbf{z}|x)\data(\dd x)$
instead of $P_{Z}(\dd\mathbf{z}|x)\data(\dd x).$ Thus, the estimator
is biased against the target estimand, the model's KSD, for a finite
burn-in period $t,$ and can therefore be seen an approximation to
the \textit{true} U-statistic $U_{n}^{(\infty)}.$ Designing a statistical
test requires understanding the behavior of the statistic \eqref{eq:mcmc_ksd_ustat},
and we will provide its analysis in the next section. Although we
focus on MCMC for its approximate unbiasedness in our proposed test,
different posterior approximations may be considered in other applications;
for example, with a more computationally efficient approach (e.g.,
variational approximation), the new KSD expression in Lemma \ref{lem:ksdnewform}
might allow us to consider parameter estimation for unnormalized statistical
models with latent variables \citep{BarpBriolDuncanEtAl2019Minimum}. 

\section{A relative goodness-of-fit test \label{sec:Main-Results}}

We now address the setting of statistical testing for model comparison.
We begin this section with our problem settings and notation, and
then define a test by showing the asymptotic normality of approximate
U-statistics. 

\subsection{Problem setup \label{subsec:Problem-setup}}

We consider the case where we have two latent variable models $P$
and $Q$, and we wish to determine which is a closer approximation
of the distribution $R$ generating our data $\{x_{i}\}_{i=1}^{n}.$
The respective density functions of the models are given by the integrals
$p(x)=\int p(x|z)\dd P_{Z}(z)$ and $q(x)=\int q(x|w)\dd Q_{W}(w)$.
As with $P$, the latent variable $w$ is assumed to take values in
a set $\mathcal{W}$ with prior $Q_{W}.$ We assume that $p(x)$ and
$q(x)$ cannot be tractably evaluated, even up to their normalizing
constants. Our goal is to determine the \emph{relative} goodness-of-fit
of the models by comparing each model's discrepancy from the data
distribution. Our problem is formulated as the following hypothesis
test: 
\begin{align}
\begin{aligned}H_{0} & :\ksd P{\data}\leq\ksd Q{\data}\ \text{(null hypothesis),}\\
H_{1} & :\ksd P{\data}>\ksd Q{\data}\ \text{(\text{alternative)}}.
\end{aligned}
\label{eq:hypotheses-1}
\end{align}
In other words, the null hypothesis is that the fit of $P$ to $R$
(in terms of KSD) is as good as $Q,$ or better. Note that the KSD
in \eqref{eq:hypotheses-1} is defined by a particular reproducing
kernel, and thus different kernels yield distinct hypotheses. For
kernel selection, we refer the reader to Section \ref{subsec:Kernel-choice}. 

We next provide an overview of the formal assumptions made throughout
the paper. Let $(\Omega,\mathcal{S},\Pi)$ be a probability space,
where $\Omega$ is a sample space, $\mathcal{S}$ is a $\sigma$-algebra,
and $\Pi$ is a probability measure. All random variables (for example,
data points $x_{i}$ and draws $\mathbf{z}_{i}^{(t)}$ from a Markov
chain sampler) are understood as measurable functions from the sample
space $\text{\ensuremath{\Omega}}.$ The input space $\mathcal{X}$
is equipped with the Borel $\sigma$-algebra generated by its standard
topology. We assume that $\mathcal{Z},\mathcal{W}$ are Polish spaces
with the Borel $\sigma$-algebras defined by their respective topologies,
on which the priors $P_{Z},\ Q_{W}$ are defined. Finally, we require
that the two models are distinct; i.e., their marginal densities disagree
on a set of positive $R$-measure.  

\subsection{Estimating kernel Stein discrepancies of latent variable models}

The hypotheses in \eqref{eq:hypotheses-1}  can be equally stated
in terms of the difference of the (squared) KSDs, $\ksdsq P{\data}-\ksdsq Q{\data},$
which motivates us to design a test statistic by estimating each term.
Let $U_{n}^{(t)}(P,Q)\coloneqq U_{n}^{(t)}(P)-U_{n}^{(t)}(Q)$ be
the difference of KSD estimates, where, $U_{n}^{(t)}(Q)$ is defined
as for $U_{n}^{(t)}(P)$ in \eqref{eq:mcmc_ksd_ustat}. Note that
$U_{n}^{(t)}(P,Q)$ is an \emph{approximate} U-statistic (in the sense
of the final paragraph in Section \ref{subsec:background-ksd-lvms})
defined by the difference kernel 
\[
\bar{H}_{p,q}[(x,\mathbf{z},\mathbf{w}),(x',\mathbf{z}',\mathbf{w}')]\coloneqq\bar{H}_{p}[(x,\mathbf{z}),(x',\mathbf{z}')]-\bar{H}_{q}[(x,\mathbf{w}),(x',\mathbf{w}')]
\]
evaluated on the joint sample $\bigl\{\bigl(x_{i},\mathbf{z}_{i}^{(t)},\mathbf{w}_{i}^{(t)}\bigr)\bigr\}_{i=1}^{n}.$
The statistic takes as input random variables with evolving laws,
and defining a test require us to understand the behavior of such
statistics. This section delivers an analysis in a general setting. 

We first characterize the asymptotic distribution of an approximate
U-statistic. The following theorem shows that such a statistic is
asymptotically normal around the expectation of the true U-statistic
provided its bias vanishes fast.

\begin{restatable}[Asymptotic normality]{theorem}{mcmcustat} \label{thm:mcmcustat}Let $\{\gamma_{t}\}_{t=1}^{\infty}$ be a sequence
of Borel probability measures on a Polish space $\mathcal{Y}$ and
$\gamma$ be another Borel probability measure. Let $\bigl\{ Y_{i}^{(t)}\bigr\}_{i=1}^{n}\overset{\mathrm{i.i.d.}}{\sim}\gamma_{t},$
and for a symmetric function $h:\mathcal{Y}\times\mathcal{Y}\to\mathbb{R},$
define a U-statistic and its mean by
\[
U_{n}^{(t)}=\frac{1}{n(n-1)}\sum_{i\neq j}h\bigl(Y_{i}^{(t)},Y_{j}^{(t)}\bigr),\ \theta_{t}=\EE_{(Y,Y')\sim\gamma_{t}\otimes\gamma_{t}}[h(Y,Y')].
\]
Let $\theta=\EE_{(Y,Y')\sim\gamma\otimes\gamma}[h(Y,Y')].$ Let $\nu_{t}\coloneqq\EE_{(Y,Y')\sim\gamma_{t}\otimes\gamma_{t}}\Bigl[\lvert\tilde{h}_{t}(Y,Y')\rvert^{3}\Bigr]^{1/3}$
with $\tilde{h}_{t}=h-\theta_{t},$ and assume $\limsup_{t\to\infty}\nu_{t}<\infty.$
Assume that $\sigma_{t}^{2}=4\var_{Y'\sim\gamma_{t}}\bigl[\EE_{Y\sim\gamma_{t}}[h(Y,Y')]\bigr]$
converges to a constant $\sigma^{2}.$  Assume that we have $\theta_{t}\to\theta$
as $t\to\infty.$ Then, in the limit of large $n$ and of $t$ growing as
a function of $n$ such that $\sqrt{n}(\theta_{t}-\theta)\to0,$ the
following two statements hold: if $\sigma>0,$ we have
\[
\sqrt{n}\left(U_{n}^{(t)}-\theta\right)\dto\Normal 0{\sigma^{2}}
\]
where $\dto$ denotes convergence in distribution; in the case $\sigma=0,$
$\sqrt{n}(U_{n}^{(t)}-\theta)\to0$ in probability.%
\end{restatable}  
The proof is in Section \ref{sec:Proofs} in the supplement. Note
that in the preceding and following results, the limit of $n$ and
$t$ is taken simultaneously rather than sequentially, such that the
condition $\sqrt{n}(\theta_{t}-\theta)\to0$ holds: see discussion
below and in Section \ref{subsec:Test-procedure}. By letting $Y_{i}^{(t)}=\bigl(x_{i},\mathcal{\mathbf{z}}_{i}^{(t)},\mathcal{\mathbf{w}}_{i}^{(t)}\bigr)$
and $h=\bar{H}_{p,q}$ in the foregoing theorem, we obtain the same
conclusion for the difference estimate $U_{n}^{(t)}(P,Q).$  

The asymptotic normality allows us to define a test procedure. Theorem
\ref{thm:mcmcustat} involves unknown variance $\sigma^{2},$ however;
in order to construct a test, we need to be able to estimate it consistently.
For our test, we propose to use the following jackknife variance estimator
\begin{equation}
v_{n,t}\coloneqq(n-1)\sum_{i=1}^{n}\Bigl(U_{n,-i}^{(t)}-U_{n}^{(t)}\Bigr)^{2}\label{eq:jackknife-variance}
\end{equation}
where $U_{n}^{(t)}$ is defined as in Theorem \ref{thm:mcmcustat},
and $U_{n,-i}^{(t)}$ the U-statistic computed on the sample with
the $i$-th data point removed. We defer the discussion on this choice
until we introduce our test procedure in Section \ref{subsec:Test-procedure}.
Here, we present the required consistency, the proof of which can
be found in Appendix \ref{subsec:var_cov_ustat} (see Lemma \ref{lem:consistency-jackknife}). 

\begin{lemma}[Jackknife consitency]

\label{lem:consistency-jackknife-mainbody} Define symbols as in Theorem
\ref{thm:mcmcustat} and the jackknife variance estimator as in \eqref{eq:jackknife-variance}.
Assume
\[
\limsup_{t\to\infty}\EE_{(Y,Y')\sim\gamma_{t}\otimes\gamma_{t}}[h(Y,Y')^{4}]<\infty.
\]
Let $\sigma^{2}=\lim_{t\to\infty}\sigma_{t}^{2}$ where $\sigma_{t}^{2}=4\zeta_{1,t}=4\mathrm{Var}_{Y\sim\gamma_{t}}\left[\mathbb{E}_{Y'\sim\gamma_{t}}\left[h(Y,Y')\right]\right].$
Then, we have the double limit $\EE\bigl(v_{n,t}^{\mathrm{}}-\sigma^{2}\bigr)^{2}\to0$
as $n,t\to\infty.$ In particular, the limit holds regardless of the
growth rate of $t$ as a function of $n.$

\end{lemma}
We have shown that the jackknife estimator allows consistent estimation
of the asymptotic variance of $U_{n}^{(t)}(P,Q).$ Using the results
obtained in this section, we present our test procedure in the next
section. 

\subsection{Test procedure}\label{subsec:Test-procedure}

We are finally ready to define the test procedure. Recall that our
objective is to compare model discrepancies, which can be accomplished
by estimating the difference $\ksdsq P{\data}-\ksdsq Q{\data}$. The
previous section has established the asymptotic normality of the difference
estimate $U_{n}^{(t)}(P,Q)$ and provides a consistent estimator of
its asymptotic variance. Therefore, we define our test statistic to
be 
\begin{equation}
T_{n,t}=\sqrt{n}\frac{U_{n}^{(t)}(P,Q)}{\sqrt{v_{n,t}}},\label{eq:teststat}
\end{equation}
with $v_{n,t}$ the jackknife variance estimator in \eqref{eq:jackknife-variance}
computed using the joint sample $\bigl\{\bigl(x_{i},\mathbf{z}_{i}^{(t)},\mathbf{w}_{i}^{(t)}\bigr)\bigr\}_{i=1}^{n}$
and kernel $h=\bar{H}_{p,q}.$ 

The following property follows from Theorem \ref{thm:mcmcustat} and
Slutsky's lemma \citep[see e.g.,][p. 13]{van2000} along with the
consistency of $v_{n,t}$ from Lemma \ref{lem:consistency-jackknife-mainbody}.
\begin{corollary}
\label{cor:teststatnormality}Let $\mu_{P,Q}=\ksdsq P{\data}-\ksdsq Q{\data}$.
Let $\mathbf{y}_{1}^{(t)}$ and $\mathbf{y}_{2}^{(t)}$ be i.i.d.
variables; $\mathbf{y}_{1}^{(t)}$ represents a copy of random variables
$(x,\mathbf{z}^{(t)},\mathbf{w}^{(t)})$; the variables $\mathbf{z}^{(t)},\mathbf{w}^{(t)}$
are draws from the respective Markov chains of $P$ and $Q$ conditioned
on $x$ after $t$ burn-in steps, and $x$ obeys $\data$.  Assume
$\limsup_{t\to\infty}\EE[\bar{H}_{p,q}[\mathbf{y}_{1}^{(t)},\mathbf{y}_{2}^{(t)}]^{4}]<\infty.$
If the assumptions in Theorem \ref{thm:mcmcustat} hold for the statistic
$U_{n}^{(t)}(P,Q)$ with asymptotic variance $\sigma_{P,Q}^{2}>0$,
we have $\sqrt{n}(U_{n}^{(t)}(P,Q)-\mu_{P,Q})/\sqrt{v_{n,t}}\dto\Normal 01$
as $n,t\to\infty,$ where the required growth of $t$ as a function
of $n$ is as in Theorem \ref{thm:mcmcustat}. 

\end{corollary}

\begin{remark}
Corollary \ref{cor:teststatnormality} holds for any choice of the
Markov chain sample size $m\geq1.$ However, in practice, a small
value of $m$ leads to large variance of the score estimates $\barscore p,\barscore q,$
and hence the test statistic $T_{n,t},$ which results in a conservative
test. To improve the test's sensitivity, we therefore recommend using
as large an $m$ as possible. 

\end{remark}

Corollary \ref{cor:teststatnormality} leads to the following simple
model comparison test (summarized in Algorithm \ref{alg:testproc}):
for a given significance level $\alpha\in(0,1)$, we compare the test
statistic $T_{n,t}$ against the $(1-\alpha)$-quantile $\tau_{1-\alpha}$
of the standard normal, and reject the null if $T_{n,t}$ exceeds
$\tau_{1-\alpha}$. By this design, under the null hypothesis $H_{0}:\mu_{P,Q}\leq0$,
we have $\lim_{n,t\to\infty}\Pi(T_{n,t}>\tau_{1-\alpha}|H_{0})\leq\alpha$,
and the test is therefore asymptotically level $\alpha$ for each
fixed $R$ satisfying $H_{0}$. On the other hand, under any fixed
alternative $H_{1}:\mu_{P,Q}>0,$ it follows from $\sqrt{n}\mu_{P,Q}\to\infty$
$(n\to\infty)$ that we have $\lim_{n,t\to\infty}\Pi(T_{n,t}>\tau_{1-\alpha}|H_{1})=1$,
indicating that the test is consistent in power. 

\begin{algorithm}[h]
\SetAlgoLined
\SetKwComment{tcc}{/*}{*/}
\KwIn{Data $\{x_i\}_{i=1}^n$, models $P$, $Q$, and significance level $\alpha$}
\KwResult{Test the null $H_0$}
\tcc{ Form a joint sample $\bigl\{\bigl(x_{i},\mathbf{z}_{i}^{(t)},\mathbf{w}_{i}^{(t)}\bigr)\bigr\}_{i=1}^{n}$}
\For{$i\leftarrow 1$ \KwTo $n$}{\label{lst:line:forloopstart}
	Generate $m$ samples $\mathbf{z}_{i}^{(t)}=(z_{i,1}^{(t)},\dots,z_{i,m}^{(t)})$ after $t$ burn-in steps with an MCMC algorithm to simulate $P_{Z}(\dd z|x_{i})$\;
	Generate $m$ samples $\mathbf{w}_{i}^{(t)}=(w_{i,1}^{(t)},\dots,w_{i,m}^{(t)})$ after $t$ burn-in steps with an MCMC algorithm to simulate $Q_{W}(\dd w|x_{i})$\;
}\label{lst:line:forloopend}
$\tau_{1-\alpha}$ $\leftarrow$ $(1-\alpha)$-quantile of $\mathcal{N}(0,1)$\;
\tcc{Compute test statistic $T_{n,t}$ in Equation \eqref{eq:teststat}}
Compute KSD difference estimate $U_{n}^{(t)}(P,Q)$\;
Compute variance estimate $v_{n,t}$\;

\tcc{Direct computation of $T_{n,t}=\sqrt{n}U_n^{(t)}(P,Q)/\sqrt{v_{n,t}}$ can be numerically unstable}
\lIf{$U_{n}^{(t)}(P,Q)>(\sqrt{v_{n,t}}/\sqrt{n})\cdot\tau_{1-\alpha}$}{
	Reject the null $H_0$
}
\caption{Test procedure}\label{alg:testproc}
\end{algorithm}

We remark that the above analysis will not apply in particular extreme
cases, where both models are identical, or both perfectly match the
data distribution. When these occur, then $\ensuremath{\sigma_{P,Q}=0}$
and $\mu_{P,Q}=0$ (note that if $\mu_{p,Q}\neq0,$ the test statistic
diverges as the sample size increases). Applying our procedure as
above to this setting, the normal approximation might fail to correctly
capture the variability of the test statistic, and the type-I error
could exceed the significance level. To detect this failure mode,
we would need to independently check that the two models are not identical,
either by inspection or via two-sample testing. A more systematic
treatment could be performed, e.g., by preventing degeneracy using
a sample splitting technique as proposed by \citet{SchennachWilhelm2017Simple},
and we leave this refinement for future work. 

We empirically found that our choice of the variance estimator acted
as a safeguard against the failure mode mentioned above. The jackknife
estimator is nonnegative, while individually estimating the variances
and covariance of the two U-statistics might yield a negative estimate.
The jackknife is also known to overestimate the variance \citep{EfronStein1981jackknife},
and its use may result in a more conservative test. This estimator
is not the only allowable choice, as the variance estimation in U-statistics
has been extensively studied \citep[for other concrete estimators, see, e.g., ][and references therein]{Maesono1998ASYMPTOTIC}.
In our preliminary analysis, we considered two other estimators, but
the jackknife estimator controlled type-I errors better than these
alternatives in the \emph{near degenerate} case. For details, we refer
the reader to experiments in Sections \ref{subsec:exp-closemodels}
and \ref{subsec:Experiment:-identical-models} in the supplement.

The limiting behaviors of the test are only guaranteed when an appropriate
double limit is taken with respect to the burn-in size $t$ and the
sample size $n.$ Theorem \ref{thm:mcmcustat} suggests that the bias
of the statistic $U_{n}^{(t)}(P,Q)$ should decay faster than $1/\sqrt{n}$
in the limit of $t.$ Our practical recommendation is to take a burn-in
period as long as the computational budget allows; this heuristic
is justified if the bias vanishes as $t\to\infty$. For $\ksdsq P{\data}$
and its estimate, the bias is due to that of the score estimate $\score p^{(t)}(x)=\EE_{\mathbf{z}|x}^{(t)}[\barscore p(x|\mathbf{z})],$
where $\EE_{\mathbf{z}|x}^{(t)}$ denotes the expectation with respect
to $P_{Z}^{(t)}(\dd{\bf z}|x).$ If the score's bias is confirmed
to converge to zero, we can check the bias of the KSD estimate by
examining the convergence of $\EE_{(x,x')\sim R\otimes R}[h_{p,t}(x,x')],$
with $h_{p,t}(x,x')$ a Stein kernel defined by the approximate score
$\score p^{(t)}.$ The convergence of $\score p^{(t)}$ can be established
by assuming appropriate conditions on $\score p(x|z)$ and the sampler;
for instance, for the exponential family likelihood $p(x|z)\propto\exp(T(x)\eta(z)),$
if the natural parameter $\eta$ is a continuous bounded function,
 the weak convergence of the sampler implies the desired convergence
(the score $\score p(x|z)$ is factorized as $\eta(z)\nabla T(x)$).
The quantification of the required growth rate of $t$ relative to
$n$ needs more stringent conditions on the employed MCMC sampler,
which we discuss in the supplement, Section \ref{subsec:conv-assumption-ustat}.
Admittedly, it is often not straightforward to theoretically establish
an explicit relation between the growth rates of $t$ and $n.$ We
therefore experimentally evaluate the finite-sample performance of
our test in Section \ref{sec:Experiments}. 

The overall computational cost of the proposed test is $O\{n^{2}+n(t+m)\},$
assuming that the cost of sampling a latent is constant.  The test
statistic in \eqref{eq:teststat} requires evaluating the U-statistic
kernel $\bar{H}_{p,q}$ on all distinct sample pairs. Note that we
need to perform this computation only once if we memoize the evaluated
values; in particular, the cost of the variance estimate \eqref{lem:consistency-jackknife-mainbody}
can be made $O(n^{2})$ with memoization. Thus, assuming that we have
evaluated and stored the score values $\bigl\{\barscore p\bigl(x_{i}\vert{\bf z}_{i}^{(t)}\bigr),\barscore q\bigl(x_{i}\vert{\bf w}_{i}^{(t)}\bigr)\bigr\}_{i=1}^{n},$
the cost of evaluating the U-statistic kernel is $O(n^{2}),$ but
this operation can be easily parallelized over sample pairs. The additional
$O\{n(t+m)\}$ cost comes from evaluating the approximate score functions,
as it requires running Markov chains for each data point (see the
loop between Lines \ref{lst:line:forloopstart}-\ref{lst:line:forloopend}
in Algorithm \ref{alg:testproc}). We can improve the sample-size
$n$ dependency in score evaluation by parallelization, since MCMC
can be performed independently over sample points $x_{i}$. 

\subsection{Kernel choice \label{subsec:Kernel-choice}}

A discrepancy measure such as KSD encodes a particular sense of how
two distributions differ. In the case of KSD, the magnitude of this
discrepancy is affected not only by evaluated models but also the
choice of a reproducing kernel. Ideally, we should choose a kernel
that makes the KSD reflect the discrepancy of features relevant to
the problem at hand. We provide general guidance on kernel selection
as follows: 

\begin{description}
\item [{Continuous~observations:}] As mentioned in Section \ref{sec:Background},
ISPD kernels enable the KSD to distinguish any two distributions satisfying
certain regularity conditions. Of ISPD kernels, in the light of practical
performances reported in prior work in goodness-of-fit testing \citep{GorMac2017}
and distribution approximation \citep{Chen2019,Riabiz_2022}, we advocate
for the use of the preconditioned IMQ kernel \citep{Chen2019}
\begin{equation}
k(x,x')=\Bigl(c^{2}+\Verts{\Lambda^{-1/2}(x-x')}_{2}^{2}\Bigr)^{-\beta}\label{eq:IMQ-definition}
\end{equation}
with $\Lambda$ a strictly positive definite matrix, and scalars $c>0$
and $0<\beta<1$; as a default choice, we recommend to take $\beta=1/2$
and $c=1.$ Following the kernel method literature, we recommend to
choose the pre-conditioner $\Lambda$ in a data-dependent way so that
the KSD can capture relevant features of the data. We suggest two
default options: the median heuristic, where $\Lambda=\lambda^{2}I$
with $\lambda=\mathrm{median}\bigl\{\Verts{x_{i}-x_{j}}_{2}:1\leq i<j\leq n\}$
and $I$ the identity matrix; the sample covariance $\Lambda=\sum_{i=1}^{n}(x_{i}-\bar{x})(x_{i}-\bar{x})^{\top}/(n-1)$
with $\bar{x}=\sum_{i=1}^{n}x_{i}/n$, which should be suitably regularized.
Each of these choices has its own merits, as we illustrate in a simple
example with Gaussian distributions in the supplement (Section \ref{subsec:kernel-Gauss-data}). 
Moreover, in general, the KSD is not invariant to a change of coordinates
representing the data. The above choices partially address this issue,
as they ensure that the KSD is invariant to rotation and displacement
\citep[see Section \ref{subsec:Coordinate-choice-independence} in the supplement;][Section 5.1]{Matsubara_2022} .
For continuous observations, we additionally need to examine the integrability
of the Stein kernel to use the KSD expression of Lemma \ref{lem:ksdnewform}.
To this end, one might want to make an assumption about the tail decay
of the data distribution. The integrability condition can be alternatively
enforced by reweighting the reproducing kernel so that the Stein kernel
is uniformly bounded; i.e., for a kernel $k,$ define a new kernel
$k_{w}$ by $k_{w}(x,x')=k(x,x')w(x)w(x')$ where $w:\calX\to(0,\infty)$
is some decreasing function dominating the growth of the score function.
We discuss how to choose such $w$ in the supplement, Section \ref{subsec:Integrability-condition-in}.
This reweighting might reduce the sensitivity of the KSD and break
the aforementioned property of coordinate-choice independence, however. 
\end{description}
\begin{description}
\item [{Discrete~observations:}] We have given a condition for a kernel
to be ISPD at the end of Section \ref{sec:Background}; e.g., the
exponentiated quadratic kernel on one-hot encoding, which can be efficiently
implemented in a sparse tensor format. In general, however, it is
challenging to compute such an ISPD kernel for discrete objects in
high-dimensions. Note that ISPD-ness is only required to distinguish
\emph{any} two distributions. In practice, we only require the KSD
to capture aspects relevant to model evaluation, and might therefore
choose a kernel insensitive to some differences, as long as they represent
computationally affordable alternatives suited to the given problem.
An instructive example is testing on distributions over graphs. Graphs
of $V$ nodes can be represented as adjacency matrices that are elements
of $\{0,1\}^{V\times V}.$ The Dirac delta kernel that examines if
two graphs are identical is an ISPD kernel but computationally intractable
(no polynomial algorithm is known). This notion of graph identification
is in practice too restrictive, and therefore one typically uses kernels
that convey other relevant graph properties \citep[see e.g.,][ for more details]{Borgwardt_2020}.
We also  demonstrate this trade-off in our experiment with latent
Dirichlet allocation models in Section \ref{subsec:expLDA}, where
we can ignore the sequential structure of the data. 
\end{description}
\begin{description}
\item [{Use~of~multiple~kernel~functions:}] As we have seen, there
are often multiple choices of the kernel function, and they might
represent distinct features. Our recommendation is to test the hypotheses
corresponding to the kernel choices simultaneously, as it makes evaluation
more rigorous. However, one has to correct for multiple comparisons
such as controlling the family-wise error rate. It should be noted
that a correction typically makes the test more conservative as the
number of kernels grows. The user thus needs to control the number
of kernels to be used (e.g., using a handful of values of scale parameter
$\lambda$ for the above IMQ kernel with $\Lambda=\diag(\lambda,\dots,\lambda)$). 
\end{description}
Finally, we note that specific kernels can be employed that encode
domain-specific expertise in particular problem settings: for instance,
kernels have been defined on groups \citep{Fukumizu2008} and graphs
\citep{Borgwardt_2020}. KSDs and associated statistical tests can
likewise be defined for certain of these cases \citep[e.g.,][]{pmlr-v108-xu20a}.
That being said, it may sometimes be preferable to favor an MMD with
goal-specific features over an omnibus KSD test. 

\subsection{Challenges with likelihood-ratio tests based on marginal density
estimation\label{subsec:Challenges-with-likelihood-ratio}}

As noted in the introduction, a likelihood-ratio test is an alternative
choice for comparing latent variable models. This section details
challenges associated with likelihood-ratio tests. There are two paths
to designing such tests, depending on how we estimate the (log-)density
ratio. One approach is to estimate the marginal density and take the
ratio, and the other is to directly estimate the (log-)density ratio.
We refer the reader to \citep{Friel_2012} for a review of estimation
techniques. 

Methods such as annealed importance sampling \citep{Neal_2001} belong
to the first category. The problem with this approach is that an estimator
ratio is a (typically heavily) biased estimator (let alone the difference
of log density estimates); thus, deriving a calibrated test threshold
from such an estimator is often challenging. In contrast, in our test,
we can characterize the bias due to MCMC relatively straightforwardly,
by evaluating the bias of the approximate Stein kernel. 

The second category includes techniques such as reversible jump MCMC
\citep{GREEN_1995} and thermodynamical integration. It is anecdotally
known that reversible jump MCMC can be challenging to implement, especially
for complex models. Thermodynamic integration uses formula
\[
\log p(x)=\int_{0}^{1}\mathbb{E}_{z|x,t}[\log p(x|z)]\mathrm{d}t,
\]
where $\mathbb{E}_{z|x,t}$ is the expectation with respect to the
posterior over $z$ defined by the tempered likelihood $p(x|z)^{t}$.
This technique can in principle be used to construct a likelihood-ratio
test, as in our approach. For example, we may uniformly sample $t$
from the unit interval to approximate the outer integral, while conducting
MCMC to estimate the inner expectation. We are not aware of a frequentist
test based on this construction, however. That said, the computational
cost of such an approach would be significant: given data of size
$n,$ we would generate $T$ samples for the temperature $t,$ then
$m$ samples for the inner MCMC, giving a (naive) computational complexity
of $O(nmT)$ (here we compute the sum of $n$ log density evaluations
rather than the evidence conditioned on the data batch).

\section{Experiments\label{sec:Experiments}}

We evaluate the proposed test (LKSD, hereafter) through simulations.
Our goal is to show the utility of the KSD in model comparison. To
this end, we compare our test with the relative MMD test \citep{BouBelBlaAntGre2016},
a kernel-based frequentist test that supports a great variety of latent
variable models. Note that this test and ours address different hypotheses,
as the MMD and LKSD tests use different discrepancy measures; it is
indeed possible that they reach conflicting conclusions (e.g., a model
is better in terms of KSD but worse in MMD). To align the judgement
of both tests, we construct problems using models with controllable
parameters; for a given class, a reference distribution, from which
a sample is drawn, is chosen by fixing the model parameter; two candidates
models are then formed by perturbing the reference's parameter such
that a larger perturbation yields a worse model for both tests. We
show that there are cases where the MMD fails to detect model differences
whereas the KSD succeeds. For completeness, we provide the detail
of our implementation of the MMD test in the supplement (Section \ref{subsec:mmd-description}),
since it requires modification to yield satisfactory performance in
our setting.

Following are details shared by the experiments below. All results
below are based on 300 trials, except for the experiment in Section
\ref{subsec:expDPM} (see the section for details). In the light of
the discussion in Section \ref{sec:Main-Results}, unless specified,
the MMD test draws $n_{\mathrm{model}}=m+t$ samples for each model
so that its cost matches the additional computation afforded to LKSD
test, which is of $O\{n(m+t)\}$ from MCMC. 

\subsection{Probabilistic Principal Component Analysis \label{subsec:PPCA}}

We first consider a simple model in which the score of its marginal
is tractable. This allows us to separately assess the impact of employing
a score function approximation. Probabilistic Principal Component
Analysis (PPCA) models serve this purpose since the marginals are
given by Gaussian distributions. Let $\calX=\mathbb{R}^{D}$ and $\calZ=\mathbb{R}^{D_{z}}$
with $1\leq D_{z}<D$. A PPCA model $\mathrm{PPCA}(A,\psi)$ is defined
by 
\[
p(x|z,A,\psi)=\Normal{Az}{\psi^{2}I_{x}},\ P_{Z}=\Normal{\bfzero}{I_{z}},
\]
where $A\in\mathbb{R}^{D\times D_{z}}$, $I_{x}\in\mathbb{R}^{D\times D},I_{z}\in\mathbb{R}^{D_{z}\times D_{z}}$
are the identity matrices, $\psi$ is a positive scalar, and $\bfzero$
is a vector of zeros. The conditional score function is $\score p(x|z)=-(x-Az)/\psi^{2}.$
In particular, the marginal density is given by $p(x)=\Normal{\bfzero}{AA^{\top}+\psi^{2}I_{x}}.$ 

While the posterior in this model is tractable, it is instructive
to see how KSD estimation is performed by MCMC. By using an MCMC method,
such as the Metropolis Adjusted Langevin Algorithm (MALA) \citep{besag1994,Roberts_1996}
or Hamiltonian Monte Carlo (HMC) \citep{duaneHybridMonteCarlo1987,Neal2011},
we obtain latent samples $\mathbf{z}_{i}\in\mathcal{Z}^{m}$ for
each $x_{i},$ which forms a joint sample $\{(x_{i},\mathbf{z}_{i})\}_{i=1}^{n}$
; samples $\mathbf{z}_{i}$ are used to compute a score estimate at
each point $x_{i},$ 
\[
\barscore p(x_{i}|\mathbf{z}_{i})=-\left\{ x_{i}-A\Bigl(\frac{1}{m}\sum_{j=1}^{m}z_{i,j}\Bigr)\right\} /\psi^{2},
\]
and these approximate score values are used to compute the U-statistic
estimate in \eqref{eq:mcmc_ksd_ustat}. By choosing suitably decaying
kernels (Section \ref{subsec:Integrability-condition-in}), we can
guarantee the integrability condition in Lemma \ref{lem:ksdnewform}.
The vanishing bias assumption in Theorem \ref{thm:mcmcustat} corresponds
to the convergence in mean, which can be measured by the Kantorovich--Rubinstein
distance \citep{Kantorovich_2006} (also known as the $L^{1}$-Wasserstein
distance \citep[see, e.g.,][Chapter 6]{Villani_2009}). Note that
the negative logarithm of the unnormalized posterior density is strongly
convex, and its gradient is Lipschitz; the strong convexity- and Lipschitz
constants are independent of $x.$ Therefore, using HMC for example,
by appropriately choosing a duration parameter and a discretization
step size, we can show that the bias of the above score estimate diminishes
uniformly over $x$ \citep{Bou_Rabee_2020}. 

\subsubsection{Type-I error and test power \label{subsec:ppca-typeI-power}}

We investigate the finite-sample performance of the proposed test
in terms of type-I error and power rates. We generate data from a
PPCA model $R=\mathrm{PPCA}(A,\psi)$. The dimensions of the observable
and the latent are set to $D=100,$ $D_{z}=10,$ respectively. Each
element of the weight matrix $A$ is drawn from a uniform distribution
$U[0,1]$ and fixed. The variance parameter $\psi$ is set to $1.$
As PPCA models have tractable marginals, we also compare our test
with the KSD test using exact score functions (i.e. no MCMC simulation),
which serves as the performance upper-bound. The MCMC sampler we use
is HMC; more precisely, we use the NumPyro \citep{phan2019composable}
implementation of No-U-Turn Sampler (NUTS) \citep{JMLR:v15:hoffman14a};
we take $t=200$ burn-in samples and $m=500$ consecutive draws for
computing a score estimate $\barscore p.$ 

We use two kernel functions: (a) the exponentiated quadratic (EQ)
kernel $k(x,x')=\exp\bigl\{-\lVert x-x'\rVert_{2}^{2}/(2\lambda^{2})\bigr\}$,
and (b) the IMQ kernel \eqref{eq:IMQ-definition} with $\beta=0.5,c=1,$
and $\Lambda=\lambda^{2}I.$ All three tests use the same kernel function,
which allows us to investigate the effect of using the Stein-modified
kernel. The length scale parameter $\lambda$ is set to the median
of the pairwise (Euclidean) distances of holdout samples from $R$
so that the parameter (and thus the hypothesis) is fixed across trials.
 We include the EQ kernel in our comparison, as the population MMD
is possible to compute, allowing us to verify the hypothesis in advance.

We simulate null and alternative cases by perturbing the weight parameter
$A$; we add a positive value $\delta>0$ to the $(1,1)$-entry of
$A.$ Let us denote a perturbed weight by $A_{\delta}$. Note that
the data PPCA model has a Gaussian marginal $\Normal{\bfzero}{AA^{\top}+\psi^{2}I_{x}}.$
Therefore, this perturbation gives a model $\Normal{\bfzero}{A_{\delta}A_{\delta}^{\top}+\psi^{2}I_{x}}$,
where the first row and column of $A_{\delta}A_{\delta}^{\top}$ deviate
from those of $AA^{\top}.$ The perturbation is additive and increasing
in $\delta,$ as each element of $A$ is positive. We create a problem
by specifying perturbation parameters $(\delta_{P},\delta_{Q})$ for
$(P,Q).$ For the EQ-kernel MMD, we numerically confirmed that the
perturbation gives a worse model for a larger perturbation. While
the population KSD is not analytically tractable, this perturbation
affects the score function through the covariance matrix, and the
same behavior is expected for KSD; see Section \ref{sec:MMD-and-KSD-pop}
in the supplement for details. 

\paragraph*{Problem 1 (null)}

We create a null scenario by choosing $(\delta_{P},\delta_{Q})=(1,1+10^{-5})$
($P$ has a smaller covariance perturbation and is closer to $R$
than $Q$). For different null settings, we refer the reader to Section
\ref{subsec:exp-closemodels} in the supplement. We run the tests
with significance levels $\alpha=0.01,0.05.$ Table \ref{tab:ppca_type1}
reports the finite-sample size of the three tests for significance
level $\alpha=0.05.$ The result for $\alpha=0.01$ is omitted as
none of the tests rejected the hypotheses. The size of the proposed
LKSD test is indeed controlled. The extremely small type-I errors
of the KSD tests are caused by the sensitivity of KSD to this perturbation;
the population KSD value is negative and far from zero, and the test
statistics easily fall in the acceptance region. The other two tests
also have their error rates lower than the significance level. Note
that their test thresholds are determined by treating the population
discrepancy differences as zero, resulting in conservative tests. 

\begin{table}[t]
\caption{Type-I errors the MMD test of \citet{BouBelBlaAntGre2016}, the proposed
LKSD test, and the KSD test in PPCA Problem 1. Rejection rates are
computed on 300 trials with significance level $\alpha=0.05.$ The
columns EQ-med and IMQ-med denote EQ and IMQ kernels with the median
bandwidth, respectively.}

\label{tab:ppca_type1}\centering\begin{threeparttable}
\begin{centering}
\begin{tabular}{rrrr|rrr}
\hline 
\headcell  \theadmd{Sample size $n$} & \multicolumn{6}{c}{\headcell  \theadmd{Rejection rates}}\tabularnewline
 & \multicolumn{3}{c|}{EQ-med} & \multicolumn{3}{c}{IMQ-med}\tabularnewline
 & \multicolumn{1}{l}{MMD} & \multicolumn{1}{l}{LKSD} & \multicolumn{1}{l|}{KSD} & \multicolumn{1}{l}{MMD} & \multicolumn{1}{l}{LKSD} & \multicolumn{1}{l}{KSD}\tabularnewline
$100$ & 0.000 & 0.013 & 0.000 & 0.000 & 0.010 & 0.000\tabularnewline
$200$ & 0.000 & 0.000 & 0.000 & 0.000 & 0.000 & 0.000\tabularnewline
$300$ & 0.003 & 0.007 & 0.000 & 0.003 & 0.003 & 0.000\tabularnewline
400 & 0.003 & 0.007 & 0.000 & 0.003 & 0.000 & 0.000\tabularnewline
$500$ & 0.007 & 0.013 & 0.000 & 0.007 & 0.007 & 0.000\tabularnewline
\hline 
\end{tabular}
\par\end{centering}
\end{threeparttable}
\end{table}

\paragraph*{Problem 2 (alternative)}

\begin{figure}[H]

\begin{centering}
\subfloat[\label{fig:ppca_h1_p2_q1_Gauss}: EQ kernel with median scaling.]{\begin{centering}
\includegraphics[width=6.5cm]{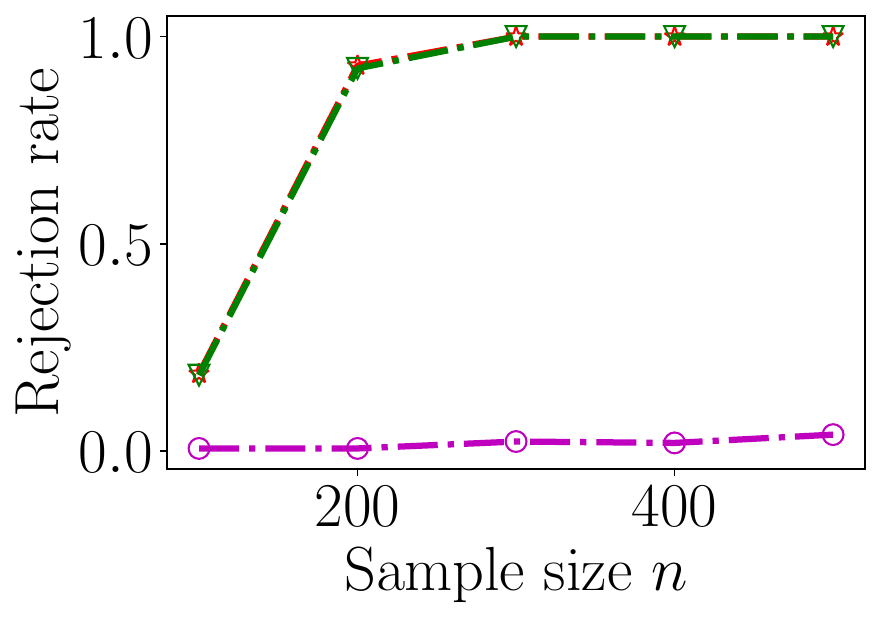}
\par\end{centering}
\centering{}}\hfill{}\subfloat[\label{fig:ppca_h1_p2_q1_IMQ-1} IMQ kernel with median scaling.]{\begin{centering}
\includegraphics[width=6.5cm]{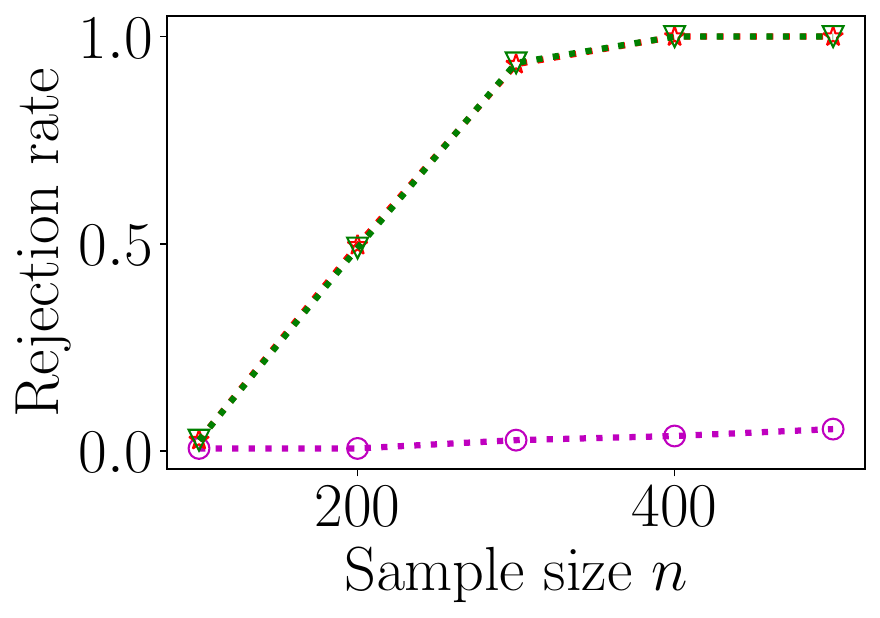}
\par\end{centering}
}
\par\end{centering}
\begin{centering}
\par\end{centering}
\caption{Power curves of the MMD test of \citet{BouBelBlaAntGre2016}, the
proposed LKSD test, and the KSD test with the exact score function
in PPCA Problem 2. The perturbation parameters are set as $(\delta_{P},\delta_{Q}=2,1).$
each result is computed on 300 trials. The significance level $\alpha=0.05.$
Markers: \textifsymbol[ifgeo]{51} (the LKSD test); \FiveStarOpen{}
(the KSD test); \textbigcircle{} (the relative MMD test).}
\label{fig:ppca_power}
\end{figure}

We investigate the power of the proposed test. We set up an alternative
scenario by fixing $\delta_{P}=2$ for $P$ and $\delta_{Q}=1$ for
$Q.$ For comparison with different parameter settings, please see
Appendix \ref{subsec:PPCA-additional}. The significance level $\alpha$
is fixed at $0.05.$ All the other parameters are chosen as in Problem
1. Figure \ref{fig:ppca_power} shows the plot of the test power against
the sample size in each problem. The KSD reaches a near 100 percent
rejection rate relatively quickly, indicating that information from
the score function is helpful for these problems. The effect of the
score approximation is negligible in this experiment, as the power
curve of the LKSD test overlaps with that of KSD. The power of the
MMD test is substantially lower than the other tests, indicating
that the MMD is insensitive to this perturbation to the covariance.

\begin{figure}[H]
\subfloat[$n=100$]{\centering{}\includegraphics[width=6cm]{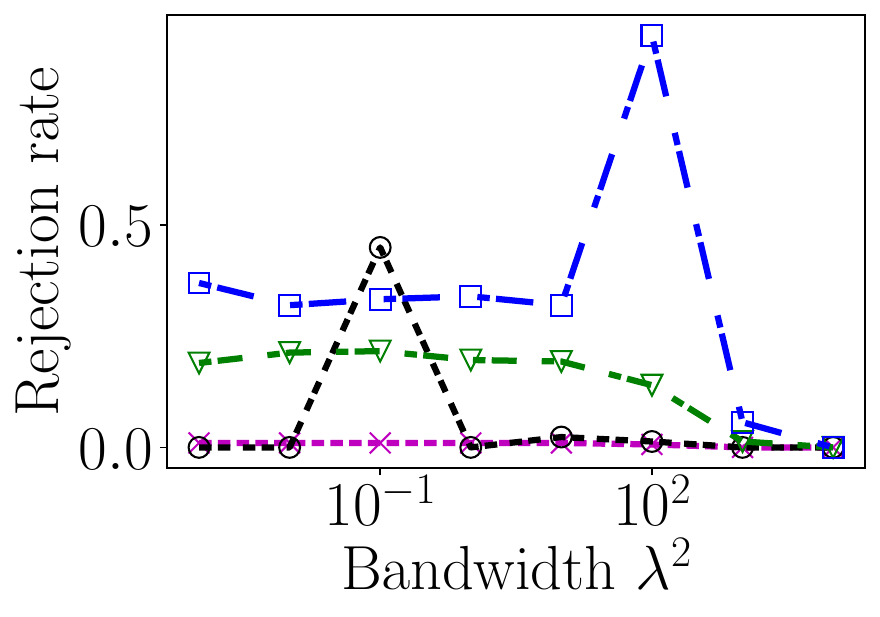}}
\hfill{}\subfloat[$n=300$]{\centering{}\includegraphics[width=6cm]{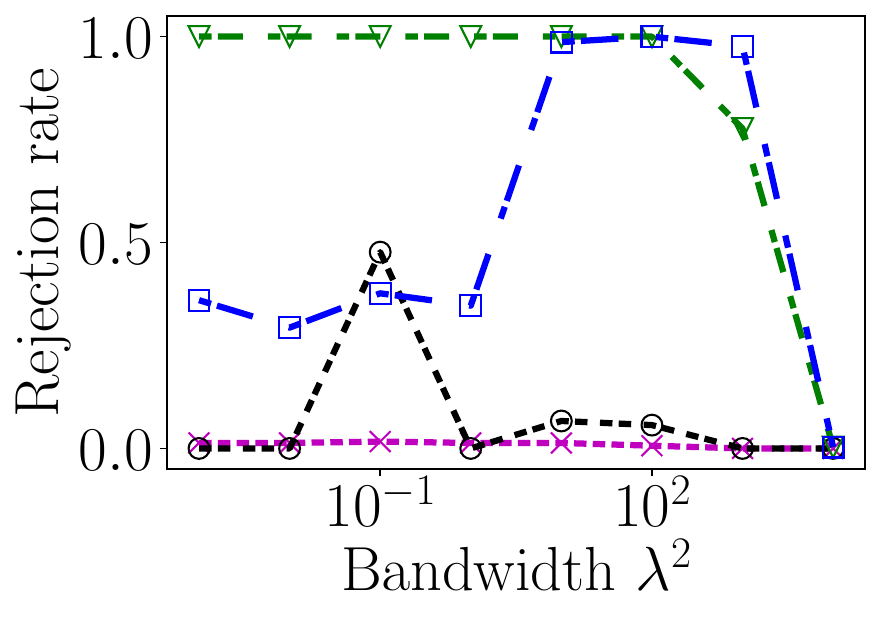}}
\caption{Power curves of the proposed LKSD test and the MMD test in PPCA Problem
2. The perturbation parameters are set as $(\delta_{P},\delta_{Q}=2,1).$
each result is computed on 300 trials. The significance level $\alpha=0.05.$
Markers: \textifsymbol[ifgeo]{51} (LKSD test with IMQ kernel); \textifsymbol[ifgeo]{48}
(LKSD test with EQ kernel); \textbigcircle{} (MMD test with IMQ kernel);
$\times$ (MMD test with EQ kernel).}

\label{fig:kernel_param_power}
\end{figure}

\subsubsection{Effect of kernel parameter choice}

\paragraph{Dependency on scaling parameter. }

Using Problem 2 above, we examine how the test power is affected by
the scaling parameter. We use the EQ and IMQ kernels as above, and
choose their scaling parameter $\lambda^{2}$ from $\{10^{-3},10^{-2},\dots,10^{3}\}.$
For each $n\in\{100,300\}$ we run 300 trials and estimate the test
power of the LKSD and MMD tests. Figure \ref{fig:kernel_param_power}
plots the power curves of the tests. We can see that the high-power
region of the EQ kernel is localized while the IMQ kernel's power
curves are flat, indicating that the IMQ kernel does not depend on
the parameter as much as the EQ. Therefore, for this problem, the
IMQ kernel can be seen as more robust against misspecification of
the scaling parameter. Nonetheless, with the right choice of the scaling
parameter, the EQ kernel yields higher power for both MMD and KSD
tests. It can be considered that the distinction arises because of
the local nature of the difference between the two distributions;
the EQ kernel is more sensitive in choosing features used to compute
the KSD (see Section \ref{subsec:kernel-Gauss-data}). 

\paragraph*{Different parameterization. }

\begin{figure}[t]
\centering{}\includegraphics[width=6cm]{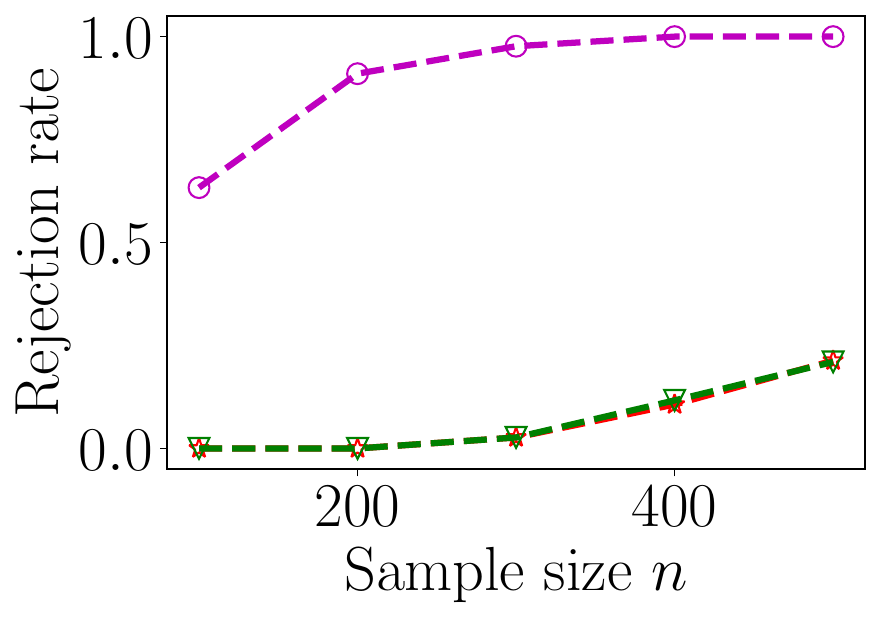}
\caption{Power curves of the MMD test, the proposed LKSD test, and the KSD
test in PPCA Problem 2. All the test use the covariance-preconditioned
IMQ kernel. The perturbation parameters are set as $(\delta_{P},\delta_{Q}=2,1).$
Each result is computed on 300 trials. The significance level $\alpha=0.05.$
Markers: \textifsymbol[ifgeo]{51} (the LKSD test); \FiveStarOpen{}
(the KSD test); \textbigcircle{} (the relative MMD test).}
\label{fig:IMQ-cov-power}
\end{figure}

We also consider a different parameter choice for the preconditioning
matrix. Here, we compare the median-scaled IMQ kernel with the same
kernel having a covariance preconditioning matrix, as suggested in
Section \ref{subsec:Kernel-choice}. Figure \ref{fig:IMQ-cov-power}
shows the power curves of the three tests. Here, the relation between
the MMD and KSD tests is overturned, and the KSD test struggles to
detect the perturbation to the covariance. This result demonstrates
that certain kernel choices can make the testing problem more challenging
than others. Using multiple kernels, rather than relying on a single
choice, could therefore robustify the evaluation, at the expense of
a loss of power due to multiple testing correction. 

\subsubsection{Quality of Markov chain samplers }

The asymptotic property of our test (Corollary \ref{cor:teststatnormality})
hinges on the quality of the Markov chain samplers. This section studies
the effect of these Markov chains on the inference. We vary the burn-in
size $t$ and the score approximation sample size $m,$ which is expected
to affect the type-I error rate and the power of the test. In the
experiments below, we set $\alpha=0.05.$ We choose $t$ from $\{50,100,\dots,600\}$
and $m$ from $\{1,10,100,1000\}.$ 

\begin{figure}[H]
\subfloat[Problem 1 (null $H_{0}$ is true).]{\centering{}\includegraphics[width=6cm]{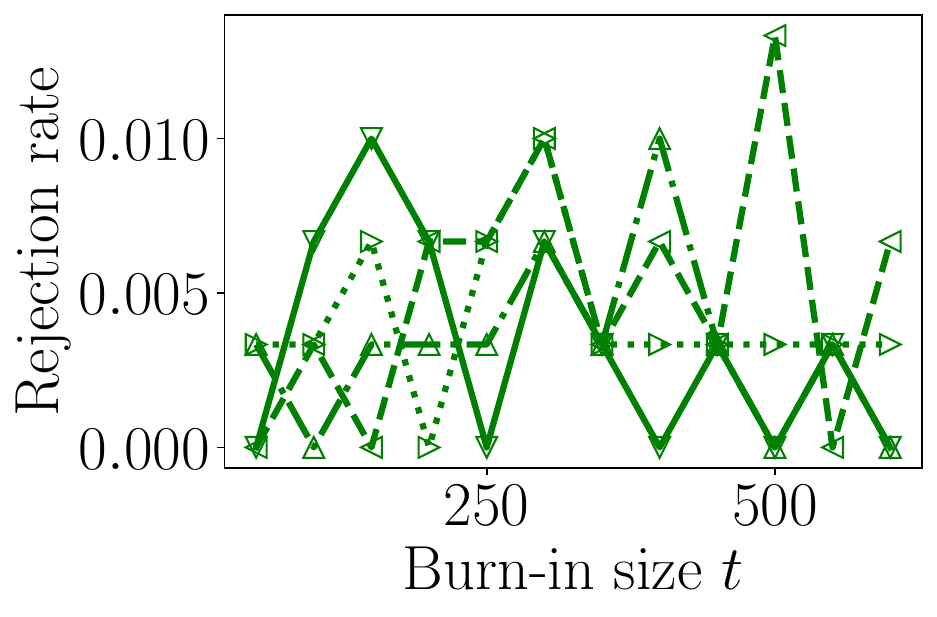}}
\hfill{}
\subfloat[Problem 2 (alternative $H_{1}$ is true).]{\centering{}\includegraphics[width=6cm]{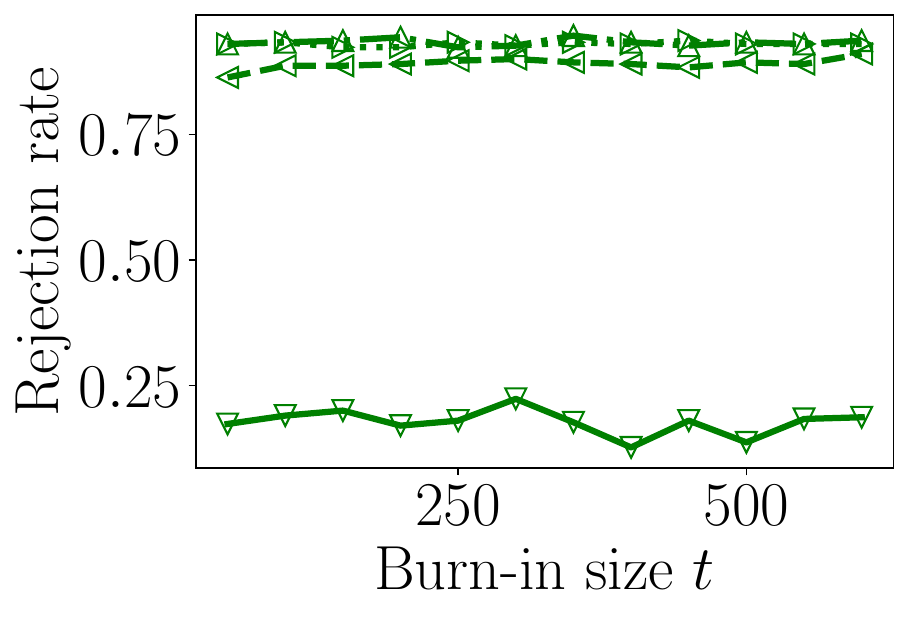}}
\caption{The effect of MCMC quality on the test's performance. Rejection rates
against burn-in size $t$ with varying Markov chain sample size $m.$
PPCA Problems 1 and 2 with $\alpha=0.05.$ Both samplers use NUTS.
Markers: \textifsymbol[ifgeo]{51} ($m=1)$; \textifsymbol[ifgeo]{50}
($m=10$); \textifsymbol[ifgeo]{49} ($m=100$); \textifsymbol[ifgeo]{52}
($m=1000$).}
\label{fig:ppca-reject-burnin-hmc}
\end{figure}

\begin{figure}[H]
\subfloat[$n=100$]{\centering{}\includegraphics[width=6cm]{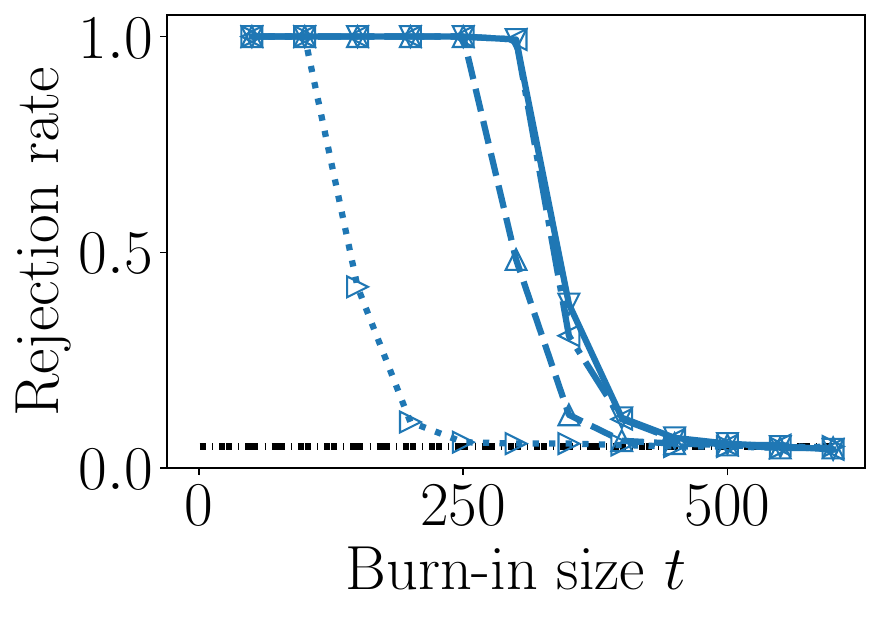}}
\hfill{}
\subfloat[$n=300$]{\centering{}\includegraphics[width=6cm]{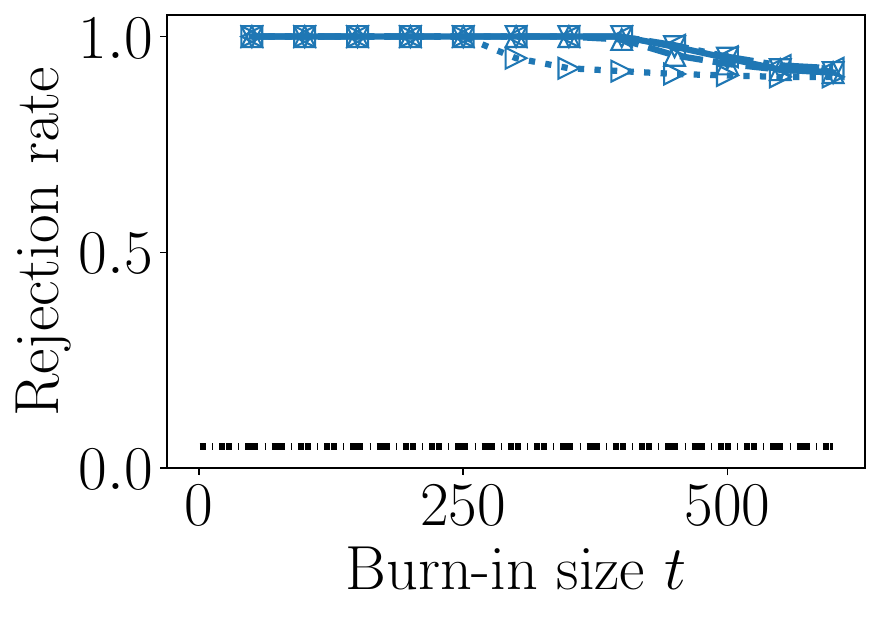}}
\caption{The effect of a poor MCMC sampler on the test. Type-I error rates
against the burn-in size $t$ with varying Markov chain sample size
$m.$ PPCA Problem 1 (the null $H_{0}$ is true). The dark dashed
line indicates the significance level $\alpha=0.05.$ The samplers
for $P$ and $Q$ are respectively MALA and NUTS. Markers: \textifsymbol[ifgeo]{51}
($m=1)$; \textifsymbol[ifgeo]{50} ($m=10$); \textifsymbol[ifgeo]{49}
($m=100$); \textifsymbol[ifgeo]{52} ($m=1000$).}
\label{fig:ppca-burn-in-MALA}
\end{figure}

In our first experiment, as in the previous sections, we use the NUTS
with the same initialization strategy for both models. With $n=300,$
we run the test using Problems 1 and 2 above. Figure \ref{fig:ppca-reject-burnin-hmc}
shows rejection rates of the test for different settings of $t$ and
$m.$ In both cases, the burn-in length $t$ does not affect the test's
performance, indicating the fast convergence of the sampler. The importance
of a larger value of $m$ can be seen when the alternative hypothesis
holds, since the test power improves as $m$ increases. The improved
performance is likely due to reduced variance. 

We next consider a slow-converging sampler for which the burn-in length
$t$ becomes crucial. We consider the null case (Problem 1) and replace
the sampler for the first model $P$ with MALA. We set the step size
for the MALA sampler to make its convergence slow; we use the step
size of $10^{-4}D_{z}^{-1/3}.$ We initialize the two samplers differently
to make sure that the resulting distributions differ when the samplers
have not converged: the MALA sampler for $P$ is initialized with
samples from a Gaussian ${\cal N}\bigl\{(1,\dots,1),I_{z}\bigr\}$
and the NUTS sampler for $Q$ a uniform distribution $U[-2,2]^{D_{z}}.$
Figure \ref{fig:ppca-burn-in-MALA} demonstrates the relation between
type-I error rates and choices of $t$ and $m$. In contrast to the
previous experiment, the burn-in has a clear effect on the type-I
error: insufficient burn-in leads to uncontrolled error rates. The
right panel ($n=300$) shows that the test has substantially higher
type-I error rates than in the left ($n=100$). Comparison between
these cases illustrates that a larger sample size $n$ requires more
intensive burn-in, as the test becomes more confident to reject. A
large value of $m$ improves the test as in the previous experiment.
It can be understood that the contribution of burn-in samples is negligible
in the score approximation. Although our analysis in Corollary \ref{cor:teststatnormality}
requires long burn-in, taking large $m$ appears to be more important
in practice, especially under a computational budget constraint. This
experiment thus confirms the importance of the quality of the sampler.

\subsection{Dirichlet process mixtures \label{subsec:expDPM}}

Our next experiment applies our test to a Dirichlet process mixtures
(DPM) model. Let $\psi(x|z)$ be a probability density function on
$\mathbb{R}^{D}.$ We consider a mixture density
\begin{equation}
\int\psi(x|z)\dd\rho(z)\label{eq:dpm-target-density}
\end{equation}
where $\rho$ is a Borel probability measure on a Polish space $\mathcal{Z}.$
A DPM model \citep{fergusonBAYESIANDENSITYESTIMATION1983} places
a Dirichlet process prior $\mathrm{DP}(a)$ on the mixing distribution
$\rho.$ Thus, a DPM model $\mathrm{DPM}(a)$ assumes the following
generative process:
\[
x_{i}\vert z_{i},\phi,F\indsim\psi(x|z_{i}),\ z_{i}\vert F\iidsim F,\ F\sim\mathrm{DP}(a).
\]
Here, $a$ is a finite Borel measure on $\mathcal{Z}.$ Note that
the marginal density (on a single observation) is given by 
\[
\EE_{F}\left[\int\psi(x|z)\dd F(z)\right].
\]
Although the prior has an infinite-dimensional component, the required
conditional score function is simply $\score{\psi}(x|z,\phi)$; thus
we only need to sample from a finite-dimensional posterior $P_{Z}(\dd z|x)$.
If a model is conditioned on held-out data ${\cal D},$ then the
predictive density $p(x\vert{\cal D})$ is $\EE_{F|\calD}\bigl[\int\psi(x|z)\dd F(z)\bigr],$
which may be used to estimate the density \eqref{eq:dpm-target-density}.
The score function is given by the expectation of $\score{\psi}(x|z)$
with respect to the posterior 
\[
\frac{\psi(x|z,\phi)}{p(x|{\cal D})}\bar{F}_{{\cal D}}(\dd z)
\]
with $\bar{F}_{{\cal D}}$ the mean measure of $P_{F}(\dd F\vert{\cal D}).$
Sampling from the posterior can be performed with a combination of
the Metropolis-Hastings algorithm and Gibbs sampling \citep[see e.g.,][Chapter 5]{ghosalFundamentalsNonparametricBayesian2017a}.
For the score formula and the MCMC procedure, we refer the reader
to Section \ref{subsec:dpm-score-formula} in the supplement. By setting
$\psi$ to an isotropic normal density, for example, we can guarantee
the integrability assumption in Lemma \ref{lem:ksdnewform} (see Section
\ref{subsec:Integrability-condition-in}). 

Our problem below considers comparing the predictive densities defined
by two models with different Dirichlet process priors. Note that since
candidate models $P,Q$ here are point estimates derived from their
respective posterior means of $F,$ we discard some aspects of uncertainty
in estimating the target \eqref{eq:dpm-target-density}; our setting
only concerns evaluating the quality of those point estimates in approximating
the data generating distribution $R.$ 

\begin{figure}[H]
\begin{centering}
\subfloat[$n=50$]{\begin{centering}
\includegraphics[scale=0.28]{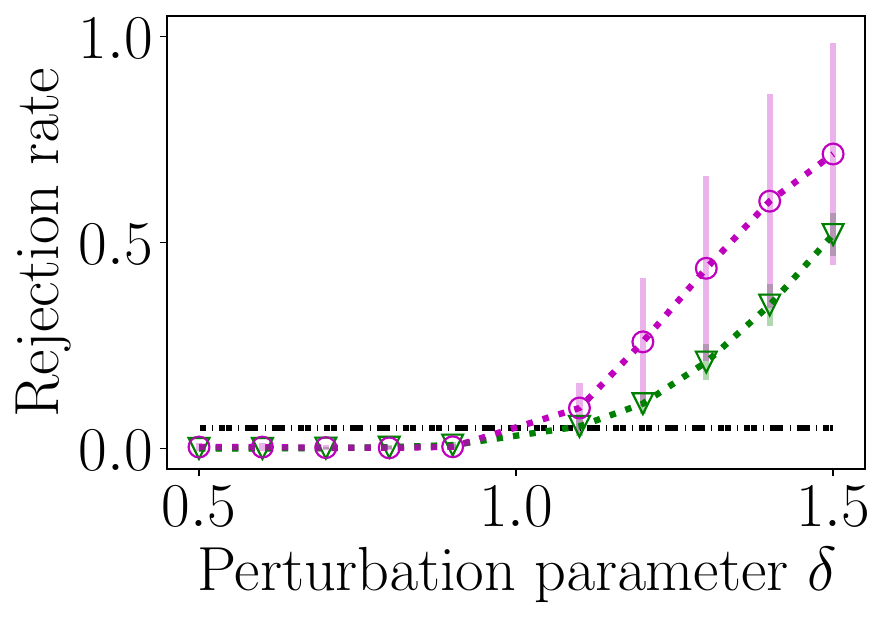}
\par\end{centering}
}\hfill{}\subfloat[$n=100$]{\begin{centering}
\includegraphics[scale=0.28]{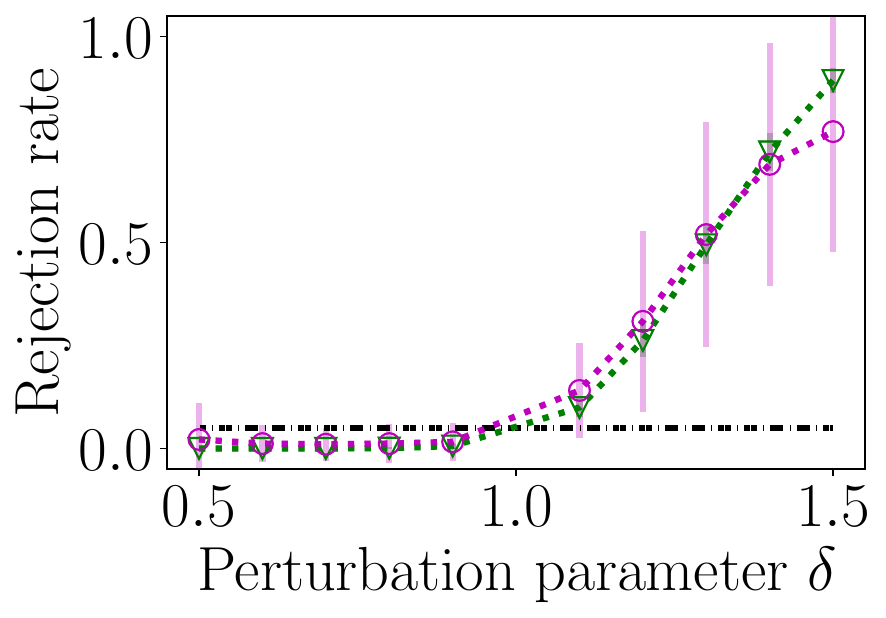}
\par\end{centering}
}\hfill{}\subfloat[$n=200$]{\begin{centering}
\includegraphics[scale=0.28]{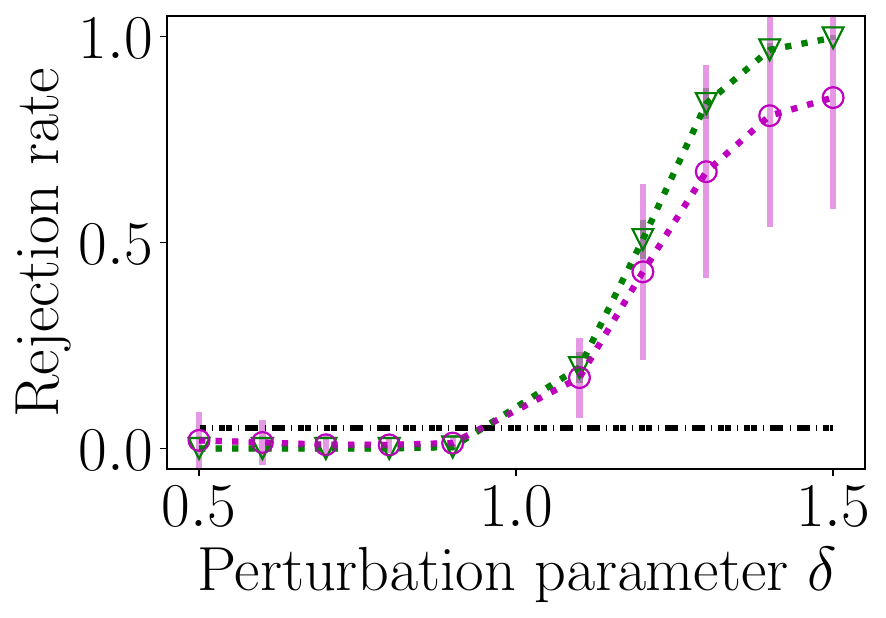}
\par\end{centering}
}
\par\end{centering}
\caption{Comparison in Gaussian Dirichlet mixture models. Rejection rates plotted
against the perturbation parameter $\delta$. The sample size $n$
is chosen from $\{50,100,200\}.$ The rejection rates are averaged
over draws of ${\cal D}_{\mathrm{tr}}.$ The supposed null and alternative
regimes are $\delta<1$ and $\delta>1,$ respectively. Markers: \textifsymbol[ifgeo]{51}
(the LKSD test); \textbigcircle{} (the relative MMD test). The dark
dashed line indicates the significance level $\alpha=0.05.$ The errorbars
indicate the standard deviations of the estimated rejections rates.}
\label{fig:gdpm_pertb}
\end{figure}

\paragraph{Experiment details. }

For the data distribution $R,$ we use the mixture \eqref{eq:dpm-target-density}
defined by $\psi(x|z)=\mathcal{\mathcal{N}}(x;z,2I)$ and $\rho=\mathcal{N}(0,I);$
this choice yields $R=\Normal{{\bf 0}}{3I}.$ We consider the following
simple Gaussian DPM model $\mathrm{GDPM}(\mu),$
\begin{align*}
x_{i}\indsim\Normal{z_{i}}{2I},\ z_{i}\iidsim F,F & \sim\mathrm{DP}(a),\ a=\Normal{\mu}I,
\end{align*}
where $\mu\in\mathbb{R}^{D}.$ Note that without conditioning on observations,
the model's marginal density is simply a Gaussian distribution $\Normal{\mu}{3I},$
which does not require approximation. 

We therefore compare predictive distributions, i.e., we compare two
GDPM models conditioned on \textit{training data} ${\cal D}_{\mathrm{tr}}=\{\tilde{x}_{i}\}_{i=1}^{n_{\mathrm{tr}}}\iidsim R.$
We consider two GDPM models with \emph{wrong} priors, where their
prior means are shifted. Specifically, we take and two models chosen
as $Q=\mathrm{GDPM}(\bar{\bfone})$ and $P=\mathrm{GDPM}(\delta\bar{\bfone})$
with $\bar{\bfone}=\bfone/\sqrt{D}.$ Unlike the preceding experiments,
we condition the two models on the training data, and obtain the predictive
distributions, denoted by $P_{{\cal D}_{\mathrm{tr}}}$ and $Q_{{\cal D}_{\mathrm{tr}}},$
respectively; our problem is thus the comparison between $P_{{\cal D}_{\mathrm{tr}}}$
and $Q_{{\cal D}_{\mathrm{tr}}}.$ The distributions now require simulating
their posterior, and we use a random-scan Gibbs sampler and the Metropolis
algorithm with a burn-in period $t=1,000$ and the size of the latents
$m=500$. For sampling observables from the models, we use a random-scan
Gibbs sampler with a burn-in period $2,000.$ We expect that if the
training sample size $n_{\mathrm{tr}}$ is small, a larger perturbation
would give a worse model as the effect of the prior is still present;
we thus set $n_{\mathrm{tr}}=5.$ Due to the small sample size, the
expected model relation might not hold, depending on the draw of ${\cal D}_{\mathrm{tr}}.$
Therefore, we examine the rejection rates of the LKSD and MMD tests,
averaged over $50$ draws; for each draw of ${\cal D}_{\mathrm{tr}},$
we estimate the rejection rates based on $100$ trials. Our problem
is formed by varying the perturbation scale $\delta$ for $P_{\calD_{\mathrm{tr}}},$
which is chosen from a regular grid $\{0.5,0.6,\cdots,0.9,1.1,\cdots,1.5\}.$
This construction gives a null case when $\delta<1$, the alternative
otherwise. We set the dimension $D$ to $10$ and the significance
level $\alpha$ to $0.05.$ As in Section \ref{subsec:PPCA}, we use
the IMQ kernel with median scaling. 

Figure \ref{fig:gdpm_pertb} reports the rejection rates of the two
tests for each of $n\in\{50,100,200\}.$ Note that the curves in the
graph do not represent type-I errors nor power, as they are rejection
rates \emph{averaged} over draws $\calD_{\mathrm{tr}},$ each of which
forms a different problem. It can be seen that on average, both tests
have correct sizes ($\delta<1$). In the alternative regime ($\delta<1$),
the LKSD test underperforms the MMD with a small sample size ($n=50$);
however, its improvement in power is faster and exceeds the MMD at
$n=200.$ These results imply that the LKSD estimate has a large variance
for a small sample size, whereas its estimand (the population difference)
is also larger, and thus the mean of the test statistic diverges faster.
Thus, it may be understood that the KSD is more sensitive to model
differences in this setting. 

\subsection{Latent Dirichlet Allocation \label{subsec:expLDA}}

Our final experiment studies the behavior of the LKSD test on discrete
data using Latent Dirichlet Allocation (LDA) models. LDA is a mixed-membership
model \citep{Airoldi2014} for grouped discrete data such as text
corpora. We follow \citet{BleNgJor03} and use the terminology of
text data for ease of exposition. Accordingly, the following terms
are defined using our notation. A word is an element in a discrete
set (a vocabulary) $\{0,\dots,L-1\}$ of size $L$. A document $x$
is a sequence of $D$ words, i.e., $x\in\{0,\dots,L-1\}^{D}$ is a
$D$-dimensional discrete vector. A prominent feature of LDA is that
it groups similar words assuming they come from a shared latent \emph{topic},
which serves as a mixture component. An LDA model assumes the following
generative process on a corpus of documents $\{x_{i}\}_{i=1}^{n}$: 
\begin{enumerate}
\item For each document $i\in\{1,\dots,n\}$, generate a distribution over
$K$ topics $\theta_{i}\overset{\mathrm{i.i.d.}}{\sim}\mathrm{Dir}(a)$
(the Dirichlet distribution), where $\theta_{i}$ is a probability
vector of size $K\geq1.$ 
\item For the $j$-th word $x_{i}^{j},j\in\{1,\dots,D\}$ in a document
$i$, 
\begin{enumerate}
\item Choose a topic $z_{i}^{j}\overset{\mathrm{i.i.d.}}{\sim}\mathrm{Cat}(\theta_{i}).$
\item Draw a word from $x_{i}^{j}\overset{\mathrm{i.i.d.}}{\sim}\mathrm{Cat(}b_{k}),$
where $b_{k}$ is the distribution over words for topic $k,$ and
the topic assignment $z_{i}^{j}=k$. 
\end{enumerate}
\end{enumerate}
Here, $a=(a_{1,}\dots,a_{K})$ is a vector of positive real numbers,
and $b=(b_{1},\dots,b_{K})^{\top}\in[0,1]^{K\times L}$ represents
a collection of $K$ distributions over $L$ words. In summary, an
LDA model $P=\mathrm{LDA}(a,b)$ assumes the factorization
\[
\mathrm{}\prod_{i=1}^{n}p(x_{i}|z_{i},\theta_{i};a,b)p(z_{i},\theta_{i};a,b)=\prod_{i=1}^{n}\left\{ \prod_{j=1}^{D}p(x_{i}^{j}|z_{i}^{j},b)p_{z}(z_{i}^{j}|\theta_{i})\right\} p_{\theta}(\theta_{i}|a),
\]
where $z_{i}$ and $\theta_{i}$ act as latent variables.

Because of the independence structure over words, the conditional
score function is simply given as 
\[
\score p(x|z,\theta,a,b)=\score p(x|z,b)=\left(\frac{p(\tilde{x}^{j}|z^{j},b)}{p(x^{j}|z^{j},b)}-1\right)_{j=1,\dots,D},\ \text{where }\tilde{x}^{j}=x^{j}+1\mod L.
\]
Score approximation requires the posterior distribution $p(z|x;a,b)$
with respect to $z.$ Marginalization of $\theta$ renders latent
topics dependent on each other, and thus the posterior is intractable.
A latent topic is conjugate to the corresponding topic distribution
given all other topics. Therefore, an MCMC method such as collapsed
Gibbs sampling allows us to sample from $p(z|x;a,b)$. As the observable
and the latent are supported on finite sets, the use of Lemma \ref{lem:ksdnewform}
is justified; the finite moment assumptions in Corollary \ref{cor:teststatnormality}
are guaranteed; and the consistency of the population mean and variance
of the test statistic follows from the convergence of $\EE_{\mathbf{z}|x}^{(t)}[\barscore p(x|\text{\ensuremath{\mathbf{z})}]}$
and $\EE_{\mathbf{w}|x}^{(t)}[\barscore q(x|\text{\ensuremath{\mathbf{w})}]}$
for each $x\in\calX.$ 

\subsubsection{Synthetic data -- prior sparsity perturbation \label{subsec:lda-prior-ptb}}

In the two problems below, we observe a sample $\{x_{i}\}_{i=1}^{n}$
from an LDA model $R=\mathrm{LDA}(a,b).$ The number of topics is
$K=3.$ The hyper-parameter $a$ is chosen as $a=(a_{0},a_{0},a_{0})$;
for model $R,$ we set $a_{0}=0.1.$ Each of three rows in $b=(b_{1,}b_{2},b_{3})^{\top}\in[0,1]^{3\times L}$
is fixed at a value drawn from the symmetric Dirichlet distribution
with all the concentration parameters one, and the vocabulary size
is $L=10,000.$ Each $x_{i}\in\{0,\dots,L-1\}^{D}$ is a document
consisting of $D=50$ words. 

We design problems by perturbing the sparsity parameter $a_{0}.$
Recall that $\mathrm{Dir}(a)$ is a distribution on the $(K-1)$~-~probability
simplex. A small $a_{0}<1$ makes the prior $p_{\theta}(\theta_{i}|a)=\mathrm{Dir}(a)$
concentrate its mass on the vertices of the simplex; the case $a_{0}=1$
corresponds to the uniform distribution on the simplex; choosing $a_{0}>1$
leads to the prior mass concentrated on the center of the simplex.
The data distribution $R$ (with $a_{0}=0.1$) is thus intended to
draw sparse topic proportions $\theta_{i}$, and a document $x_{i}$
is likely to have words from a particular topic. By increasing $a_{0},$
we can design a departure from this behavior. Therefore, as in the
PPCA experiments, we additively perturb $a_{0}$ with parameters $(\delta_{P},\delta_{Q})$
for respective candidate models $(P,Q).$

\begin{table}[t]
\caption{Rejection rates of the MMD test and the LKSD test in LDA experiments.
Each result is based on 300 trials. }

\centering{}\subfloat[Type-I errors of the KSD and MMD tests in LDA Problem 1; $(\delta_{P},\delta_{Q})=(0.5,0.6).$
The significance level $\alpha=0.05.$]{\label{tab:lda-type1}\begin{threeparttable}

\begin{tabular}{r|rr}
\hline 
\headcell \theadmd{Sample size $n$} & \multicolumn{2}{c}{\headcell \theadmd{Rejection rates}}\tabularnewline
 & \multicolumn{1}{l}{MMD} & \multicolumn{1}{l}{LKSD}\tabularnewline
$100$ & 0.003 & 0.013\tabularnewline
$200$ & 0.010 & 0.007\tabularnewline
$300$ & 0.007 & 0.003\tabularnewline
$400$ & 0.003 & 0.007\tabularnewline
$500$ & 0.007 & 0.010\tabularnewline
\hline 
\end{tabular}

\end{threeparttable}}\hfill{}\subfloat[Power of the KSD and MMD tests in LDA Problem 2; $(\delta_{P},\delta_{Q})=(1.0,0.5).$
The significance level $\alpha$ is chosen from $\{0.01,0.05\}.$]{\label{tab:lda-power}\begin{threeparttable}

\begin{tabular}{rrr|rr}
\hline 
\multicolumn{1}{l}{\headcell \theadmd{Sample size $n$}} & \multicolumn{4}{c}{\headcell \theadmd{Rejection rates}}\tabularnewline
 & \multicolumn{2}{l|}{Level $\alpha=0.01$} & \multicolumn{2}{l}{Level $\alpha=0.05$}\tabularnewline
 & \multicolumn{1}{l}{MMD} & \multicolumn{1}{l|}{LKSD} & \multicolumn{1}{l}{MMD} & \multicolumn{1}{l}{LKSD}\tabularnewline
$100$ & 0.000 & 0.010 & 0.007 & 0.070\tabularnewline
$200$ & 0.003 & 0.030 & 0.010 & 0.183\tabularnewline
$300$ & 0.000 & 0.097 & 0.003 & 0.283\tabularnewline
$400$ & 0.000 & 0.197 & 0.010 & 0.463\tabularnewline
$500$ & 0.000 & 0.280 & 0.007 & 0.570\tabularnewline
\hline 
\end{tabular}

\end{threeparttable}}
\end{table}

As LDA disregards word order, we need a kernel that respects this
structure. We use the Bag-of-Words (BoW) IMQ kernel $k(x,x')=(1+\lVert B(x)-B(x')\rVert_{2}^{2})^{-1/2};$
it is simply the IMQ kernel computed in the BoW representation $B(x)\in\{0,1,2,\dots,D\}^{L}$
whose $\ell$-th entry (counting from $0$ to $L-1$) is the count
of the occurrences of word $\ell\in\{0,\dots,L-1\}$ in a document
$x.$ By Lemma \ref{lem:KSD-perm-invariance} in Section \ref{subsec:Model-invariance},
this choice ensures that arbitrary reordering of text sequences does
not change the KSD value; i.e., the KSD does not assess models by
their ability to generate sequences. We also tested differing input-scaling
values and found that the bandwidth of the IMQ kernel did not have
a significant effect on the test power (Section \ref{subsec:lda-kernel-param}). 

For score estimation in the LKSD test, we use a random scan Gibbs
sampler; we generate $m=1,000$ latent samples after $t=4,000$ burn-in
iterations. 

\paragraph*{Problem 1 (null)}

We create a null situation by having $(\delta_{P},\delta_{Q})=(0.5,0.6).$
In this case, $Q'$s prior on $\theta$ is less sparse than that of
$P.$  Table \ref{tab:lda-type1} shows the size of the different
tests for significance levels $\alpha=0.05;$ the result for $\alpha=0.01$
is omitted as both tests did not reject the hypothesis. It can be
seen that the rejection rates of both tests are bounded by the nominal
level. 

\paragraph*{Problem 2 (alternative)}

We consider an alternative case in which the sparsity parameters are
chosen as $(\delta_{P},\delta_{Q})=(1.0,0.5)$ (we consider other
parameter choices in Appendix \ref{subsec:LDA-additional}). Here,
the model $Q$ is expected to have less mixed topic proportions. Table
\ref{tab:lda-power} demonstrates the power of the MMD and LKSD tests.
The power of the LKSD test improves as the sample size $n$ increases,
whereas the MMD has almost no power in this case. In this problem,
the topics $b$ are not sparse enough for each topic to have a sufficiently
distinctive vocabulary. Thus, the problem is challenging for the MMD,
as it is unable to find distinguishing words, in addition to the high-dimensionality.
By contrast, the KSD is able to distinguish the models by taking advantage
of their underlying structure. 

\subsubsection{Synthetic data -- topic perturbation \label{subsec:lda-topic=002013ptb}}

We provide a negative example to illustrate a failure mode of the
LKSD test for discrete data. The data is generated as in the previous
section, whereas we construct two models differently. We set up a
model by perturbing the topics of the data model $R.$ That is, a
model is given by $\mathrm{LDA}(a,b_{\delta})$ with $b_{\delta}=(1-\delta)b+\delta b_{\mathrm{ptb}}$
with $0<\delta<1.$ We choose $b_{\mathrm{ptb}}$ as we did for $b;$
the value is drawn independently of $b.$ We set the perturbation
parameter for $Q$ as $\delta=0.01$ and vary it for $P,$ where the
value is chosen from $\{0.06,0.11,\dots,0.51\}.$ Thus, $P$ is morphed
from $b$ to $b_{\mathrm{ptb}}$ and therefore expected to underperform
$Q$ as perturbation $\delta$ increases. We run trials with $n=300.$
For score estimation, we take $m=10,000$ and $t=4,000.$ 

\begin{figure}
\centering{}\includegraphics[width=8cm]{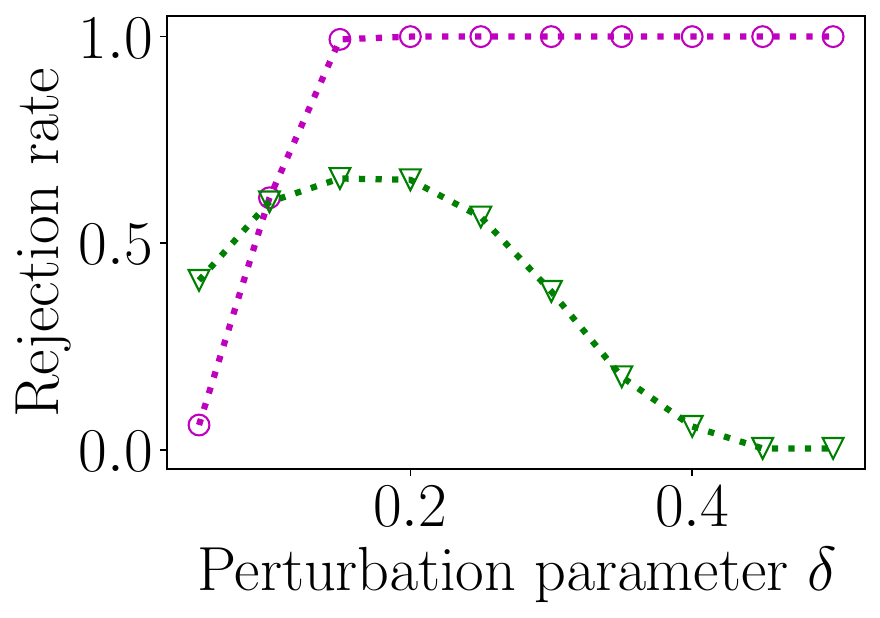}
\caption{Power estimates plotted against perturbation parameters $\delta.$
The significance level $\alpha=0.05;$ the sample size $n=300.$ Markers:
\textifsymbol[ifgeo]{51} (the LKSD test); \textbigcircle{} (the MMD
test).}
\label{fig:lda-topic-ptb}
\end{figure}

Figure \ref{fig:lda-topic-ptb} shows the plot of rejection rates
against perturbation parameters. We see that the power of the LKSD
test degrades as the perturbation increases. As $P$'s topic becomes
close to $b_{\text{ptb}},$ some words in the target's topic $b$
become rare and therefore fall in the low probability region of $P$.
This situation leads to increasing variance of the test statistic
as $\delta$ increases, because the score function contains the reciprocal
$1/p(x).$ The LKSD test can therefore fail when the support of the
model is severely mismatched to that of the data, since the high variance
of the statistic makes is difficult to detect significant departures
from the null. Note that this observation does not apply to the continuous
counterpart as the score can be written as the gradient of the logarithm
of the density, which is typically numerically stable. 

\subsubsection{Comparing topic models for arXiv articles \label{subsec:arxiv}}

Our final experiment investigates the test's performance using the
arXiv dataset \citep{UniCornell2020}. The dataset consists of meta
information of scholarly articles on the e-print service arXiv. We
treat the abstract of an article as a document, and use paper categories
to set up a problem. Specifically, we construct a problem by choosing
three paper categories for model $P,Q$ and the data distribution
$R.$ Unlike the preceding experiments, for a model category, we fit
an LDA model to the dataset of abstracts in the category. As the KSD
requires the number of words to be fixed, then for a given data category,
we extract abstracts of length no less than $D=100$ and subsample
excess words. This process yields a dataset of articles of equal length
$D;$ for each trial, we obtain the data $\{x_{i}\}_{i=1}^{n}$ by
subsampling from the larger set of articles. Thus, our problem is
to compare two LDA models trained on different article sets, and assess
their fit to the dataset. 

In the following experiments, we examine the power of LKSD and MMD
tests. We vary the sample size $n$ from 100 to 500. We fix the dataset
category to stat.TH (statistics theory) and inspect two combinations
of model categories. To train an LDA model $\mathrm{LDA}(a,b)$, we
use the Gensim implementation \citep{Rehurek2011} of the variational
algorithm of \citet{Hoffman2010}. For sparsity parameters $a,$ we
use the parameter returned by this algorithm; we point-estimate topics
$b$ using the mean of the topics under the variational distribution.
The number of topics is set to $100.$ The vocabulary set is comprised
of words that appear in the abstracts of three chosen categories.
As in the previous experiments, we use the IMQ-BoW kernel for both
tests. We fix the significance level $\alpha$ at 0.05. 

As we have seen the numerical instability issue in the previous section,
we also consider an alternative KSD that is stable but computationally
more expensive, as mentioned in \ref{subsec:background-ksd} and the
supplement (Section \ref{subsec:stable-discrete-stein}). For this,
we take a burn-in size $t=500$ and a Markov chain size $m=1,000.$
We denote this method by LKSD-stable.

\paragraph{Probability theory vs Statistical methodology. }

We choose math.PR (mathematics probability theory) for $P$ and stat.ME
(statistics methodology) for $Q.$ In addition to the taxonomic proximity
to stat.TH, the category stat.ME has a larger proportion of articles
shared with the target category: $3,121$ of $18,973$ (stat.ME) vs.
$2,884$ of $46,769$ (math.PR). Thus, we expect $Q$ to outperform
$P.$ This combination results in a vocabulary set of size $L=126,190.$
For score estimation, we set the burn-in length $t$ to $500$ and
the Markov chain sample size $m$ to $5,000.$ Additionally, we run
the LKSD test with $m=15,000$ (labeled LKSD-extra) and the MMD test
with the model sample size $n_{\mathrm{model}}=10,000$ (labeled MMD-extra).
The sample size $n_{\mathrm{model}}$ is thresholded at $10,000$ as the computational
cost exceeds that of the LKSD test (in fact, sampling in this case
makes the MMD by an order of magnitude slower due to the large vocabulary
size). 

Table \ref{tab:mathPRstatME} summarizes the result. The MMD test
underperforms all the KSD-based tests; extra sampling did not lead
to a significant improvement. We can see that increasing the Markov
chain size $m$ boosts the LKSD test, as it reduces the variance of
the score estimator. The low power of the MMD test indicates that
the model difference is too subtle to discern from the word compositions
of generated documents; the LKSD tests offers a different viewpoint
based on the model information. 

\begin{table}[t]
\caption{Rejection rates of the MMD test and the LKSD test in the math.PR vs.
stat.ME experiment. Each result is based on 300 trials.}
\label{tab:mathPRstatME}
\centering{}%
\begin{tabular}{rrrrrr}
\hline 
\multicolumn{1}{l}{\headcell \theadmd{Sample size $n$}} & \multicolumn{5}{c}{\headcell \theadmd{Rejection rates}}\tabularnewline
 & \multicolumn{1}{l}{\raggedleft MMD} & \multicolumn{1}{l}{\raggedleft MMD-extra} & \multicolumn{1}{l}{\raggedleft LKSD} & \raggedleft LKSD-extra & \raggedleft LKSD-stable\tabularnewline
$100$ & 0.150 & 0.157 & 0.333 & 0.673 & 0.437\tabularnewline
$200$ & 0.160 & 0.167 & 0.807 & 0.880 & 0.845\tabularnewline
$300$ & 0.197 & 0.207 & 0.913 & 0.980 & 0.950\tabularnewline
$400$ & 0.180 & 0.187 & 0.950 & 0.986 & 0.970\tabularnewline
$500$ & 0.267 & 0.263 & 0.966 & 0.993 & 0.983\tabularnewline
\hline 
\end{tabular}
\end{table}

\begin{table}[H]
\caption{Rejection rates of the MMD test and the LKSD test in the cs.LG vs.
stat.ME experiment. Each result is based on 300 trials.}
\label{tab:csLGstatME}
\centering{}%
\begin{tabular}{rrrr}
\hline 
\multicolumn{1}{l}{\headcell \theadmd{Sample size $n$}} & \multicolumn{3}{c}{\headcell \theadmd{Rejection rates}}\tabularnewline
 & \multicolumn{1}{l}{\raggedleft MMD} & \multicolumn{1}{l}{\raggedleft LKSD} & \raggedleft LKSD-stable\tabularnewline
$100$ & 1 & 0.000 & 0.287\tabularnewline
$200$ & 1 & 0.007 & 0.643\tabularnewline
$300$ & 1 & 0.013 & 0.833\tabularnewline
$400$ & 1 & 0.013 & 0.873\tabularnewline
$500$ & 1 & 0.113 & 0.923\tabularnewline
\hline 
\end{tabular}
\end{table}

\paragraph{Machine learning vs Statistical Methodology.}

Our second experiment uses cs.LG (computer science machine learning)
for $P$, while $Q$ uses the same category as the previous experiment.
With this combination, the vocabulary size $L$ is $208,671.$ By
the same reasoning as above, the second model $Q$ is expected to
be better than $P.$ We run the same tests as above and compare their
performances. 

Table \ref{tab:csLGstatME} summarizes the result. This experiment
serves as a negative case study for the LKSD test: the MMD tests achieved
power 1 for all sample-size choices (MMD-extra is omitted here), whereas
the power of the LKSD test does not exceed even the significance level
$\alpha$ for most sample size settings (LKSD-extra is omitted as
increasing the Markov chain size did not improve the power). We attribute
this failure to the unmatched support of the model $P$ in the test
distribution. This reasoning is supported by the high power of the
MMD, as the BoW feature easily detects deviation of document patterns
in this case. Thus, as we noted in the synthetic experiment in Section
\ref{subsec:lda-topic=002013ptb}, the LKSD test fails when there
is a severe mismatch in data and model support. The stable LKSD test
approaches the same level as the MMD at $n=500,$ but still underperforms.
While stable, the KSD used for this test can also suffer from the
mismatch of the support, since it depends on the same density ratio
as in the unstable counterpart.

\section{Conclusion}

We have developed a test of relative goodness of fit for latent variable
models based on the kernel Stein discrepancy. The proposed test applies
to a wide range of models, since the requirements of the test are
mild: (a) models have MCMC samplers for inferring their latent variables,
and (b) likelihoods have evaluable score functions.  The proposed
test complements existing model evaluation techniques by providing
a different means of model comparison, which takes advantage of the
known model structure. Our experimental results confirm this view
-- the relative MMD test was unable to detect subtle differences
between models in several of our benchmark experiments. 

Our asymptotic analysis of the test statistic indicates that the test
could suffer from bias if the mixing of the deployed MCMC sampler
is slow. Removing the assumptions on the bias and the moments in Theorem
\ref{thm:mcmcustat} is certainly desirable; we envision that the
recent development of unbiased MCMC \citep{jacobUnbiasedMarkovChain2020}
could be used to construct an alternative unbiased KSD estimator,
and leave this possibility as future work. While we have focused on
comparing two models, extensions to ranking multiple models are possible
as in \citep{LimYamSchJit2019}. Finally, the technique used in this
article can be applied to other Stein discrepancies requiring the
score function \citep{BarpBriolDuncanEtAl2019Minimum,pmlr-v108-xu20a};
one interesting application would be the KSD for directional data
\citep{pmlr-v108-xu20a}, where densities with computable normalizing
constants are scarce.

\section*{Data availability}
Code to reproduce all the results is available at \url{https://github.com/noukoudashisoup/lkgof}.

\section*{Acknowledgements}

We would like to thank the three anonymous referees for their constructive
feedback that greatly improved the manuscript; Chris Oates for pointing
us to the Fisher's identity in the early stage of this project; Jiaxin
Shi for suggesting the Barker-Stein operator; Bharath Sriperumbudur,
Motonobu Kanagawa, and Tamara Fernandez for helpful comments and discussions.
H. Kanagawa and A. Gretton acknowledge support of the Gatsby Charitable
Foundation. 
K. Fukumizu was supported in part by JST CREST Grant Number JPMJCR2015 and JSPS KAKENHI Grant Number 22H05106.

\section*{Author notes}
\emph{Conflicts of interest:} None declared. 

\bibliographystyle{rss}
\bibliography{lstein}

\newpage{}
\input{appendix}

\end{document}

%% file: appendix.tex
\appendix

\section{Proofs \label{sec:Proofs}}

This section provides proofs for the results concerning the asymptotic
normality of our test statistic: (a) Theorem \ref{thm:mcmcustat},
and (b) an estimator of the variance of a U-statistic and its consistency. 

\subsection{Asymptotic normality of approximate U-statistics}

\begin{restatable}[Asymptotic normality]{theorem}{mcmcustat} \label{thm:mcmcustat}Let $\{\gamma_{t}\}_{t=1}^{\infty}$ be a sequence
of Borel probability measures on a Polish space $\mathcal{Y}$ and
$\gamma$ be another Borel probability measure. Let $\bigl\{ Y_{i}^{(t)}\bigr\}_{i=1}^{n}\overset{\mathrm{i.i.d.}}{\sim}\gamma_{t},$
and for a symmetric function $h:\mathcal{Y}\times\mathcal{Y}\to\mathbb{R},$
define a U-statistic and its mean by
\[
U_{n}^{(t)}=\frac{1}{n(n-1)}\sum_{i\neq j}h\bigl(Y_{i}^{(t)},Y_{j}^{(t)}\bigr),\ \theta_{t}=\EE_{(Y,Y')\sim\gamma_{t}\otimes\gamma_{t}}[h(Y,Y')].
\]
Let $\theta=\EE_{(Y,Y')\sim\gamma\otimes\gamma}[h(Y,Y')].$ Let $\nu_{t}\coloneqq\EE_{(Y,Y')\sim\gamma_{t}\otimes\gamma_{t}}\Bigl[\lvert\tilde{h}_{t}(Y,Y')\rvert^{3}\Bigr]^{1/3}$
with $\tilde{h}_{t}=h-\theta_{t},$ and assume $\limsup_{t\to\infty}\nu_{t}<\infty.$
Assume that $\sigma_{t}^{2}=4\var_{Y'\sim\gamma_{t}}\bigl[\EE_{Y\sim\gamma_{t}}[h(Y,Y')]\bigr]$
converges to a constant $\sigma^{2}.$  Assume that we have $\theta_{t}\to\theta$
as $t\to\infty.$ Then, in the limit of large $n$ and of $t$ growing as
a function of $n$ such that $\sqrt{n}(\theta_{t}-\theta)\to0,$ the
following two statements hold: if $\sigma>0,$ we have
\[
\sqrt{n}\left(U_{n}^{(t)}-\theta\right)\dto\Normal 0{\sigma^{2}}
\]
where $\dto$ denotes convergence in distribution; in the case $\sigma=0,$
$\sqrt{n}(U_{n}^{(t)}-\theta)\to0$ in probability.
\end{restatable}  

\begin{proof}

Recall that $(\Omega,\mathcal{S},\Pi)$ is the underlying probability
space, and $U_{n}^{(t)}$ is a random variable on it. We show that
the cumulative distribution function (CDF) of $\sqrt{n}(U_{n}^{(t)}-\theta)$
converges to that of a normal distribution in the limit of $n$ and
$t$ specified in the statement. 

First we consider the case $\sigma>0.$ Note that we can express the
CDF as 
\begin{align*}
\Pi\left[\sqrt{n}\left(\frac{U_{n}^{(t)}-\theta}{\sigma}\right)<\tau\right] & =\Pi\left[\sqrt{n}\left(\frac{U_{n}^{(t)}-\theta_{t}+\theta_{t}-\theta}{\sigma_{t}}\right)<\frac{\sigma}{\sigma_{t}}\tau\right]\\
 & =F_{n,t}\left(\frac{\sigma\tau-\sqrt{n}(\theta_{t}-\theta)}{\sigma_{t}}\right)
\end{align*}
where $F_{n,t}$ denotes the CDF of $\sqrt{n}(U_{n}^{(t)}-\theta_{t})/\sigma_{t}.$
Let $\Phi$ be the CDF of the standard Gaussian distribution. Then,
for $\tau\in\mathbb{R},$
\begin{align*}
\left|\Pi\left[\sqrt{n}\left(\frac{U_{n}^{(t)}-\theta}{\sigma}\right)<\tau\right]-\Phi(\tau)\right| & \leq\underbrace{\left|F_{n,t}\left(\frac{\sigma\tau-\sqrt{n}(\theta_{t}-\theta)}{\sigma_{t}}\right)-\Phi\left(\frac{\sigma\tau-\sqrt{n}(\theta_{t}-\theta)}{\sigma_{t}}\right)\right|}_{(\mathrm{i})}\\
 & \hphantom{\leq}+\underbrace{\left|\Phi\left(\frac{\sigma\tau-\sqrt{n}(\theta_{t}-\theta)}{\sigma_{t}}\right)-\Phi(\tau)\right|}_{\mathrm{(ii)}}.
\end{align*}
We show that both terms on the RHS converge to zero simultaneously.
Let $\nu=\limsup_{t\to\infty}\nu_{t}$ and $\delta$ be a fixed constant
such that $0<\delta<\sigma.$ Note that by the convergence assumptions
on $\nu_{t}$ and $\sigma_{t}$, there exists $t_{\nu,\delta}\geq1$
such that $\nu_{t}<\nu+\delta<\infty$ for any $t\geq t_{\nu,\delta}$;
and there exists $t_{\sigma,\delta}\geq1$ such that $\sigma_{t}>\sigma-\delta>0$
for any $t\geq t_{\sigma,\delta}.$ By the Berry-Esseen bound for
U-statistics \citep{callaert_berry-esseen_1978}, for $t\geq\max(t_{\nu,\delta},t_{\sigma,\delta})$
and any $n\geq2$, the term (i) is bounded as 
\begin{align*}
\mathrm{(i)} & \leq\sup_{\tau'}\left|F_{n,t}(\tau')-\Phi(\tau')\right|\\
 & \leq C\left(\frac{\nu_{t}}{\sigma_{t}}\right)^{3}n^{-\frac{1}{2}}\\
 & <C\left(\frac{\nu+\delta}{\sigma-\delta}\right)^{3}n^{-\frac{1}{2}},
\end{align*}
where $C$ is a universal constant. For the term (ii), by the continuity
of $\Phi$ and our assumptions on $\sqrt{n}(\theta_{t}-\theta)$ and
$\sigma_{t},$ we can make the term (ii) arbitrarily small. Formally,
by assumption, we can specify $t=g(n)$ with an increasing function
$g$ such that $\sqrt{n}(\theta_{g(n)}-\theta)\to0$ as $n\to\infty;$
therefore, for $\varepsilon>0,$ we can take $N_{\varepsilon/2}\geq1$
such that the term (ii) is bounded by $\varepsilon/2$ for any $n\geq N_{\varepsilon/2}.$
Thus, for any $\varepsilon>0$, choosing $n$ and $t$ such that
\[
n\ge\max\left(\left(\frac{2C}{\varepsilon}\cdot\frac{\nu+\delta}{\sigma-\delta}\right)^{2},N_{\varepsilon/2}\right),
\]
and $t\geq\max\bigl(t_{\nu,\delta},t_{\sigma,\delta},g(N_{\varepsilon/2})\bigr),$
we have 
\[
\left|\Pi\left[\sqrt{n}\left(\frac{U_{n}^{(t)}-\theta}{\sigma}\right)<\tau\right]-\Phi(\tau)\right|\leq\mathrm{(i)}+\mathrm{(ii)}\leq\frac{\varepsilon}{2}+\frac{\varepsilon}{2}=\varepsilon.
\]

Next, for the case $\sigma=0,$ consider the squared error $n\EE(U_{n}^{(t)}-\theta)^{2},$
which is decomposed as 
\[
n\EE(U_{n}^{(t)}-\theta)^{2}=n\EE(U_{n}^{(t)}-\theta_{t})^{2}+n(\theta_{t}-\theta)^{2}.
\]
The first term is the variance of the U-statistic $U_{n}^{(t)},$
and so according to \citet[Eq. 5.18]{hoeffding_class_1948}, we have,
for any $n\geq2,$
\begin{align*}
n\EE(U_{n}^{(t)}-\theta_{t})^{2} & =\frac{(n-2)}{(n-1)}\underbrace{4\var_{Y\sim\gamma_{t}}\bigl[\EE_{Y'\sim\gamma_{t}}[h(Y,Y')\bigr]}_{\sigma_{t}^{2}}+\frac{2}{(n-1)}\var_{Y,Y'\sim\gamma_{t}\otimes\gamma_{t}}[h(Y,Y')]\\
 & \leq\sigma_{t}^{2}+\frac{2}{n-1}\var_{Y,Y'\sim\gamma_{t}\otimes\gamma_{t}}[h(Y,Y')].
\end{align*}
We have $w=\limsup_{t\to\infty}\var_{Y,Y'\sim\gamma_{t}\otimes\gamma_{t}}[h(Y,Y')]<\infty$
by the finiteness of (the limit supremum of) the third central moment,
and $\sigma_{t}^{2}\to\sigma^{2}=0$ by assumption. Therefore, for
any $\varepsilon>0,$ we can take $t_{\varepsilon,v}\geq1$ such that
\[
\var_{Y,Y'\sim\gamma_{t}\otimes\gamma_{t}}[h(Y,Y')]<w+1,\text{and }\sigma_{t}^{2}\leq\frac{\varepsilon}{4},
\]
for any $t\geq t_{\varepsilon,v}.$ Choosing $n\geq4(w+1)/\varepsilon+1,$
we have $n\EE(U_{n}^{(t)}-\theta_{t})^{2}\leq\varepsilon/2.$ For
the second term, we can take $N_{\varepsilon/2}\geq1$ such that $n(\theta_{t}-\theta)^{2}\leq\varepsilon/2$
for any $n,t\geq N_{\varepsilon/2}.$ Thus, having 
\[
n,t\geq\max\Bigl(t_{\varepsilon,v},4\frac{w+1}{\varepsilon}+1,N_{\varepsilon/2}\Bigr)
\]
leads to $n\EE(U_{n}^{(t)}-\theta)^{2}\leq\varepsilon.$ We have shown
$n\EE(U_{n}^{(t)}-\theta)^{2}\to0,$ which implies $\sqrt{n}(U_{n}^{(t)}-\theta)\to0$
in probability. \qed

\end{proof}

\subsection{Variance of a U-statistic \label{subsec:var_cov_ustat}}

We first recall known facts about U-statistics. For an i.i.d. sample
$\{y_{i}\}_{i=1}^{n}\sim\data$, let us define a U-statistic 
\[
U_{n}=\frac{1}{n(n-1)}\sum_{i\neq j}h(y_{i},y_{j}),
\]
where $h$ is a symmetric measurable kernel. According to \citet[Eq. 5.18]{hoeffding_class_1948},
the variance of $U_{n}$ is 
\begin{equation}
\var[U_{n}]=\frac{4(n-2)}{n(n-1)}\zeta_{1}+\frac{2}{n(n-1)}\zeta_{2},\label{eq:ustat-var-decomposition}
\end{equation}
where $\zeta_{1}=\mathrm{Var}_{y\sim R}\left[\mathbb{E}_{y'\sim R}\left[h(y,y')\right]\right]$
and $\zeta_{2}=\mathrm{Var}_{y,y'\sim R\otimes R}\left[h(y,y')\right].$
To obtain the asymptotic variance of the $\sqrt{n}(U_{n}-\EE[U_{n}])$,
we only need the first term $\zeta_{1}$ as $n\var[U_{n}]\to4\zeta_{1}$
$(n\to\infty),$ assuming $\EE_{(y,y')\sim R\otimes R}[h(y,y')^{2}]<\infty.$

Recall that our test statistic in Section \ref{subsec:Test-procedure}
requires a consistent estimator of the asymptotic variance in Corollary
\ref{cor:teststatnormality} (see also Theorem \ref{thm:mcmcustat}).
To accommodate the setting of our test, we consider the following
situation: we are given samples $\{y_{i}^{(t)}\}_{i=1}^{n}\sim R_{t}$
where $\{R_{t}\}_{t=1}^{\infty}$ is a sequence of distributions approximating
$R.$ This defines a sequence of U-statistics
\[
U_{n}^{(t)}=\frac{1}{n(n-1)}\sum_{i\neq j}h\bigl(y_{i}^{(t)},y_{j}^{(t)}\bigr),
\]
and the variance $\var[U_{n}^{(t)}]$ is given by the corresponding
parameters $\zeta_{1,t}$ and $\zeta_{2,t}$ (see Eq. \eqref{eq:ustat-var-decomposition}).
In the following, we address the estimation of $\sigma^{2}=\lim_{t\to\infty}\sigma_{t}^{2}$
with $\sigma_{t}^{2}=4\zeta_{1,t},$ assuming that the limit exists.
As Theorem \ref{thm:mcmcustat} suggests, the quantity $\sigma^{2}$
is the asymptotic variance of $\sqrt{n}(U_{n}^{(t)}-\EE[U_{n}]).$

In the main body, we define our test using the jackknife estimator
\begin{equation}
v_{n,t}\coloneqq(n-1)\sum_{i=1}^{n}\bigl(U_{n,-i}^{(t)}-U_{n}^{(t)}\bigr)^{2}\label{eq:sigma-nt-definition}
\end{equation}
(see also the test statistic $T_{n,t}$ in Section \ref{subsec:Test-procedure}),
where $U_{n,-i}^{(t)}$ the U-statistic computed on the sample with
the $i$-th data point removed. In the following, we provide a consistency
proof for this estimator. 

\paragraph*{Jackknife estimator and its consistency. }

We first revisit the definition of the jackknife variance estimator
and its properties. For simplicity, we first drop the dependency on
$t.$ The jackknife estimator of the variance of a (scaled) U-statistic
$\sqrt{n}(U_{n}-\EE[U_{n}])$ is defined as 
\begin{equation}
v_{n}^{J}=(n-1)\sum_{i=1}^{n}(U_{n,-i}-U_{n})^{2},\label{eq:jackknife-without-t}
\end{equation}
where $U_{n,-i}$ is the U-statistic computed with the sample with
the $i$-th data point removed. According to  \citet[][Eq. 25, see also Section \ref{subsec:Jackknife-formula}]{Arvesen1969Jackknifing},
the jackknife estimator has the expansion
\begin{equation}
v_{n}^{J}=\sum_{c=0}^{2}a_{n,c}\hat{U}_{c},\label{eq:jackknife-Ustat-decomp}
\end{equation}
where 
\[
a_{n,c}=\frac{n-1}{n}{n-1 \choose 2}^{-2}\bigl\{ n\mathbb{I}(c>0)-4\bigr\}{n \choose c}{n-c \choose 2-c}{n-2 \choose 2-c}
\]
with $\mathbb{I}(\cdot)$ the indicator function; each term in the
sum is a U-statistic
\begin{align}
\hat{U}_{c} & ={n \choose 4-c}^{-1}\sum_{(\alpha,\beta,\gamma)\in C_{n,4-c}}h_{\mathrm{sym}}(y_{\alpha_{1}},\dots,y_{\alpha_{c}},y_{\beta_{1}},\dots,y_{\beta_{2-c}},y_{\gamma_{1}},\dots,y_{\gamma_{2-c}}),\label{eq:jackknife-inner-Ustat}
\end{align}
where the sum is over all combinations of $(4-c)$ integers chosen
from $\{1,\dots,n\}.$ If indices end with $0$ as in $(\alpha_{1},\dots,\alpha_{0}),$
it should be understood that the corresponding variables are omitted.
The function $h_{\mathrm{sym}}$ in \eqref{eq:jackknife-inner-Ustat}
is defined as a symmetric kernel, 
\begin{align*}
 & h_{\mathrm{sym}}(y_{\alpha_{1}},\dots,y_{\alpha_{c}},y_{\beta_{1}},\dots,y_{\beta_{2-c}},y_{\gamma_{1}},\dots,y_{\gamma_{2-c}})\\
 & =\sum_{\sigma\in\Sigma(\alpha,\beta,\gamma)}\frac{\tilde{h}(y_{\sigma(\alpha_{1})},\dots,y_{\sigma(\alpha_{c})},y_{\sigma(\beta_{1})},\dots,y_{\sigma(\beta_{2-c})})\tilde{h}(y_{\sigma(\alpha_{1})},\dots,y_{\sigma(\alpha_{c})},y_{\sigma(\gamma_{1})},\dots,y_{\sigma(\gamma_{2-c})})}{(4-c)!},
\end{align*}
with $\tilde{h}=h-\EE[U_{n}]$ and $\Sigma(\alpha,\beta,\gamma)$
the set of all permutations of given $4-c$ integers $(\alpha_{1},\dots,\alpha_{c},\beta_{1},\dots,\beta_{2-c},\gamma_{1},\dots,\gamma_{2-c}).$
Note that we have $\EE[\hat{U}_{c}]=\zeta_{c}$ with $\zeta_{0}=0.$
For large $n,$ the coefficient $a_{n,c}$ behaves as 
\begin{align*}
a_{n,c} & \approx4\frac{1}{c!}\left(\frac{1}{(2-c)!}\right)^{2}n^{1-c}\ (c\geq1),\\
a_{n,0} & =O(1).
\end{align*}
Therefore, if $\EE h(y,y')^{2}<\infty,$ the estimator $v_{n}^{J}$
converges to the asymptotic variance $4\zeta_{1}$ of the U-statistic
$\sqrt{n}(U_{n}-\EE[U_{n}]),$ a.s. 

Next, we recover the dependency on $t$ and define a jackknife estimator
\[
v_{n,t}^{J}=(n-1)\sum_{i=1}^{n}\bigl(U_{n,-i}^{(t)}-U_{n}^{(t)}\bigr)^{2},
\]
which is the estimator in \eqref{eq:jackknife-without-t} computed
on the sample $\bigl\{ y_{i}^{(t)}\bigr\}_{i=1}^{n}.$ The following
lemma provides the required consistency. 

\begin{lemma}

\label{lem:consistency-jackknife} Assume
\[
\limsup_{t\to\infty}\EE_{(y_{1},y_{2})\sim R_{t}\otimes R_{t}}[h(y_{1},y_{2})^{4}]<\infty.
\]
Let $\sigma^{2}=\lim_{t\to\infty}\sigma_{t}^{2}$ where $\sigma_{t}^{2}=4\zeta_{1,t}=4\mathrm{Var}_{y\sim R_{t}}\left[\mathbb{E}_{y'\sim R_{t}}\left[h(y,y')\right]\right].$
Then, we have the double limit $\EE\bigl(v_{n,t}^{J}-\sigma^{2}\bigr)^{2}\to0$
as $n,t\to\infty.$ 

\end{lemma}

\begin{proof}

Note that we have the following relation 
\begin{align}
\EE[(v_{n,t}^{J}-\sigma^{2})^{2}] & =\EE[(v_{n,t}^{J}-\EE[v_{n,t}^{J}])^{2}]+(\EE[v_{n,t}^{J}]-\sigma^{2})^{2}\nonumber \\
 & =\underbrace{\EE[(v_{n,t}^{J}-\EE[v_{n,t}^{J}])^{2}]}_{\text{variance}}+\underbrace{\bigl(\EE[v_{n,t}^{J}]-\sigma_{t}^{2}\bigr)^{2}}_{\text{(squared) bias}}-2\bigl(\EE[v_{n,t}^{J}]-\sigma_{t}^{2}\bigr)\bigl(\sigma_{t}^{2}-\sigma^{2}\bigr)+\bigl(\sigma_{t}^{2}-\sigma^{2}\big)^{2}.\label{eq:variance-bias-variance}
\end{align}
The decomposition indicates that as long as the bias and the variance
terms in \eqref{eq:variance-bias-variance} decay as $n,t\to\infty,$
the estimator $v_{n,t}^{J}$ serves as a consistent estimator of $\sigma^{2}$
(note that we have $\sigma_{t}^{2}-\sigma^{2}\to0$ by assumption).
In the following, we show that the assertion holds. 

For the bias term, note that by the decomposition \eqref{eq:jackknife-Ustat-decomp},
we have 
\[
\EE[v_{n,t}^{J}]=\sigma_{t}+O(n^{-1})\zeta_{2,t}.
\]
 By the assumption on the fourth moment, the limit supremum of $\zeta_{2,t}=\var_{R_{t}\otimes R_{t}}[h(y_{1},y_{2})^{2}]$
is finite. Therefore, for any $\varepsilon>0,$ we can take $N_{b,\varepsilon}\geq1$
such that $(\EE[v_{n,t}^{J}]-\sigma_{t})^{2}<\varepsilon$ for $n,t\geq N_{b,\varepsilon}.$

For the variance term, using the decomposition \eqref{eq:jackknife-Ustat-decomp},
we have 
\begin{align*}
\EE[(v_{n,t}^{J}-\EE[v_{n,t}^{J}])^{2}\bigr] & =\sum_{c,c'=0}^{2}a_{c,n}a_{c',n}\cov[\hat{U}_{c}^{(t)},\hat{U}_{c'}^{(t)}]\\
 & \leq\left(\sum_{c=0}^{2}a_{c,n}\var[\hat{U}_{c}^{(t)}]^{1/2}\right)^{2},
\end{align*}
where $\hat{U}_{c}^{(t)}$ is the U-statistic $\hat{U}_{c}$ computed
with the sample $\{y_{i}^{(t)}\}_{i=1}^{n}.$ According to \citet[Section 5.2.1, Lemma A]{Ser2009},
\begin{align*}
\var[\hat{U}_{c}^{(t)}] & \leq\frac{4-c}{n}\var[h_{\mathrm{sym}}(y_{\alpha_{1}},\dots,y_{\alpha_{c}},y_{\beta_{1}},\dots,y_{\beta_{2-c}},y_{\gamma_{1}},\dots,y_{\gamma_{2-c}})],
\end{align*}
where the variance is taken with respect to the product measure $\otimes_{i=1}^{4-c}R_{t}.$
Note that the variance on the RHS is bounded as, for each $c\in\{0,1,2\},$
\begin{align*}
\begin{cases}
\var[h_{\mathrm{sym}}(y_{1}y_{2},y_{3},y_{4})]\leq\EE\left[\tilde{h}_{t}(y_{1},y_{2})^{4}\right] & (c=0),\\
\var[h_{\mathrm{sym}}(y_{1},y_{2},y_{3})]\leq\EE\left[\tilde{h}_{t}(y_{1},y_{2})^{4}\right] & (c=1),\ \text{and }\\
\var[h_{\mathrm{sym}}(y_{1},y_{2})]\leq\EE[\tilde{h}_{t}(y_{1,}y_{2})^{4}] & (c=2),
\end{cases}
\end{align*}
where $\tilde{h}_{t}=h-\EE_{R_{t}\otimes R_{t}}[h(y_{1},y_{2})].$
By the assumption $\limsup_{t\to\infty}\EE_{(y_{1},y_{2})\sim R_{t}\otimes R_{t}}[h(y_{1},y_{2})^{4}]<\infty,$
the above observation implies that for each $c\in\{0,1,2\},$ 
\[
\limsup_{t\to\infty}\var_{\otimes_{i=1}^{4-c}R_{t}.}[h_{\mathrm{sym}}(y_{\alpha_{1}},\dots,y_{\alpha_{c}},y_{\beta_{1}},\dots,y_{\beta_{2-c}},y_{\gamma_{1}},\dots,y_{\gamma_{2-c}})]<\infty.
\]
Therefore, taking appropriate $n,t$ diminishes $\var[\hat{U}_{c}^{(t)}].$
In consequence, as for the bias term, we can take $N_{v,\varepsilon}\geq1$
such that $\EE[(v_{n,t}^{J}-\EE[v_{n,t}^{J}])^{2}\bigr]<\varepsilon$
for $n,t\geq N_{v,\varepsilon}.$ Thus, the bias and variance terms
can be made arbitrarily small by taking $n,t\geq\max(N_{v,\varepsilon},N_{b,\varepsilon})$.
\qed

\end{proof}

\newpage{}

\section{Auxiliary results}

In this section we provide auxiliary results used in the main body. 

\subsection{Kernel Stein discrepancy for bounded domains \label{subsec:bounded-domain}}

We detail the case where data take values in a bounded open set ${\cal X}\subset\mathbb{R}^{D}.$
For a density $p,$ we require that $p\in{\cal C}({\cal \bar{X}})\cap C^{1}({\cal X})$
with $\bar{{\cal X}}$ the closure of ${\cal X}.$ The kernel is assumed
to satisfy the same conditions everywhere; i.e., $k(x,\cdot)\in{\cal C}(\bar{\calX})\cap C^{1}({\cal X}).$
Additionally, we require the following conditions : (a) the domain
${\cal X}$ is assumed to be sufficiently regular -- we assume that
${\cal X}$ is convex or $C^{1};$ (b) for any boundary point $x\in\partial{\cal X}$,
$p(x)k(x,\cdot)=0;$ and (c) $\EE_{x\sim P}\Verts{\score p(x)}_{2}<\infty,$
$\EE_{x\sim P}[k(x,x)^{1/2}]<\infty,$ and $\EE_{x\sim P}[k_{12}(x,x)^{1/2}]<\infty.$
The first assumption allows us to use the divergence theorem, and
the third one ensures the $P$-integrability of $\la\score p,f\ra$
and $\la\nabla,f\ra$ \citep[Corollary 4.36]{Steinwart2008}.  By
the divergence theorem, the second condition on the kernel implies
that any $f=(f_{1},\ldots f_{D})$ of the corresponding RKHS $\mathcal{F}$
satisfies $\EE_{x\sim P}[{\cal A}_{P}f(x)]=0$ \citep[Proposition 1]{GorMac2015}.
We can fulfill the required condition (b) by choosing a kernel $k_{\varphi}$
defined by $k_{\varphi}(x,x')=\varphi(x)\varphi(x')k(x,x')$ where
$\varphi:{\cal X}\to[0,\infty)$ such that $\varphi\in{\cal C}({\cal \bar{X}})\cap C^{1}({\cal X})$
and $\varphi(x)=0$ for $x\in\mathbb{R}^{D}\setminus{\cal X}.$ The
KSD is then defined similarly as in Section \ref{sec:Background}.
We demonstrate an example of this setting in Section \ref{subsec:boundedPPCA}. 

\subsection{Numerically stable Stein operator \label{subsec:stable-discrete-stein}}

As we note in Section \ref{sec:Background}, the Stein operator of
\citet{YanLiuRaoNev18} can induce numerically unstable functions,
as it contains the reciprocal $1/p(x).$ This appendix provides a
more stable alternative using the Zanella-Stein operator proposed
by \citet{Hodgkinson2020}. We focus on the Barker-Stein operator
of \citet{Shi2022} below, but other operators can be considered depending
on the application. For instance, when the size $L$ of the discrete
domain is small (e.g., binary domains such as adjacent matrices of
graphs), we may use the Gibbs-Stein operator \citep{BreNag19,ReiRos19,Shi2022},
as it is manageable to perform the required conditional expectation. 

\subsubsection{Univariate case}

We first consider the univariate case ${\cal X}=\{0,\dots,L-1\}$
with $L>1,$ as the multivariate case builds on the construction here.
Let be a density $p$ on ${\cal X}.$ The Zanella-Stein operator is
defined by 
\[
{\cal {\cal A}}_{p}^{\mathrm{Uni}}f(x)=\sum_{y\in{\cal N}_{x}}a\left(\frac{p(y)}{p(x)}\right)\bigl(f(y)-f(x)\bigr),
\]
where the function $a:[0,\infty)\to[0,\infty)$ is referred to as
a balancing function, which is assumed to satisfy $a(0)=0,$ and $a(r)=ra(1/r)$
for $r>0.$ The symbol ${\cal N}_{x}\subset{\cal X}$ denotes the
neighborhood of $x.$ The operator is derived from the infinitesimal
generator of a Markov jump process with the jump rate given by $a$
as a function of $p(y)/p(x).$ In the following, we use the Barker
balancing function $a(r)=r/(1+r);$ this results in 
\[
a\left(\frac{p(y)}{p(x)}\right)=\frac{\frac{p(y)}{p(x)}}{1+\frac{p(y)}{p(x)}}=\frac{p(y)}{p(x)+p(y)}.
\]
This choice of the balancing function was proposed by \citet{Shi2022}
and is referred to as the Barker-Stein operator. The above ratio takes
values between $0$ and $1$ (it saturates to one as the ratio $p(y)/p(x)$
gets large), and is thus numerically stable even when $p(x)$ is close
to zero. The neighborhood ${\cal N}_{x}$ is chosen such that the
process can transit from any starting point to any other point, and
admits $p$ as the invariant distribution \citep[see Example 2 of][for details]{Hodgkinson2020}.
In the following, ${\cal N}_{x}$ is assumed to be the two adjacent
points of $x$ with respect to the cyclic difference. This construction
is not limited to $D=1,$ but it can be challenging to compute the
sum in a high-dimensional space when applied to the KSD, since the
number of neighbors typically grows with the dimension. 

\subsubsection{Multivariate case }

We next consider the multivariate case ${\cal X}=\{0,\dots,L-1\}^{D}$
with $D>1.$ We follow the product space construction of a Stein operator
\citep[Proposition 2]{Hodgkinson2020}. We define an operator ${\cal A}_{p}$
that acts on a $D$-dimensional vector-valued function $f=(f_{1},\dots,f_{D}):{\cal X\to\mathbb{R}}^{D}$
by 
\[
{\cal A}_{p}f(x)=\sum_{d=1}^{D}{\cal A}_{p_{d}(\cdot|x_{-d})}^{\mathrm{\mathrm{Uni}}}{\cal P}_{d}^{x}f_{d}(x),
\]
where $p_{d}$ is the distribution given by conditioning on all but
the $d$-th coordinate, $x_{-d}=(x_{1},\dots,x_{d-1},x_{d+1},\dots x_{D}),$
and ${\cal P}_{d}^{x}$ is the projection defining a function on the
$d$-th coordinate by freezing all the other coordinates; i.e., ${\cal P}_{d}^{x}f:y\in\{0,\dots,L-1\}\mapsto f(x_{1},\dots x_{d-1},y,x_{d+1},\dots,x_{D}).$
We can define the KSD using a vector-valued RKHS ${\cal {\cal F}}=\prod_{d=1}^{D}{\cal {\cal F}}_{k}$
with ${\cal {\cal F}}_{k}$ the RKHS of a scalar kernel $k:{\cal X\times{\cal X}\to\mathbb{R}}.$
With this choice, we can define the KSD as in \ref{subsec:background-ksd};
in particular, for this operator, we have 
\[
\ksdsq PR=\EE_{x,x'\sim R\otimes R}[h_{p}(x,x')]
\]
where 
\begin{align*}
h_{p}(x,y) & =\sum_{d=1}^{D}{\cal A}_{x,p_{d}(\cdot|x_{-d})}^{\mathrm{\mathrm{Uni}}}\otimes{\cal A}_{y,p_{d}(\cdot|y_{-d})}^{\mathrm{\mathrm{Uni}}}k(x,y)
\end{align*}
for any $x,y\in{\cal X},$ with ${\cal A}_{*,p_{d}(\cdot|x_{-d})}^{\mathrm{\mathrm{\mathrm{Uni}}}}$
acting on $*.$ Specifically, the Stein kernel $h_{p}$ is given by
\begin{align*}
h_{p}(x,y) & =\sum_{d=1}^{D}\sum_{\nu\in{\cal N}_{x,d}}\sum_{\tilde{\nu}\in{\cal N}_{y,d}}a_{\nu}(x)a_{\nu}(y)\{k(\nu(x),\tilde{\nu}(y))+k(x,y)-k(x,\tilde{\nu}(y))-k(\nu(x),y)\}.
\end{align*}
Here, ${\cal N}_{x,d}$ denotes the set of neighborhood points with
respect to the $d$th coordinate; we identify this as a set of functions,
each of which maps $x$ to the corresponding neighboring point. The
weight $a_{\nu}$ is defined by 
\[
a_{\nu}(x)=a\left(\frac{p(\nu(x))}{p(x)}\right).
\]
From this expression, we can conclude that the KSD is possible to
estimate as long as we can evaluate $a_{\nu}.$ Note that a single
evaluation of the Stein kernel $h_{p}$ requires $O(DN)$ with $N$
being the maximum of the sizes of the neighborhood ${\cal N}_{x,d}$
(in our case, $N=2).$ 

\subsubsection{Application to latent variable models}

The KSD defined above admits essentially the same treatment as in
the main body. The Stein operator above requires evaluating the marginal
density $p(x)=\int p(x|z)P_{Z}(\dd z).$ Following the approach in
Section \ref{subsec:background-ksd-lvms}, we can circumvent this
issue as 
\begin{align*}
a_{\nu}(x) & =\frac{p(\nu(x))}{p(x)+p(\nu(x))}\\
 & =\int\frac{p(\nu(x)|z)}{p(x|z)+p(\nu(x)|z)}\frac{p(x|z)+p(\nu(x)|z)}{p(x)+p(\nu(x))}P_{Z}(\dd z).
\end{align*}
The weight $a_{\nu}(x)$ can be estimated by sampling from the \emph{modified}
posterior 
\[
P_{Z,\nu}(\dd z|x)\propto\{p(x|z)+p(\nu(x)|z)\}P_{Z}(\dd z).
\]
Given a sample $\mathbf{z}=\{z_{i}\}_{i=1}^{m}$ for simulating $P_{Z,\nu}(\dd z),$
we can estimate $a_{\nu}(x)$ by 
\[
a_{\nu}(x|\mathbf{z})\coloneqq\frac{1}{m}\sum_{i=1}^{m}\frac{p(\nu(x)|z_{i})}{p(x|z_{i})+p(\nu(x)|z_{i})},
\]
which is possible to evaluate as long as the likelihood $p(x|z)$
is tractable. In contrast to the difference operator KSD, this approach
requires simulating $O(D)$ posterior distributions, as each dimension
generates distinct modified posterior distributions. Thus, while numerically
stable, this approach is computationally more expensive. The user
might want to consider this limitation when they choose between the
operator defined here and the difference Stein operator considered
in the main body. 

\subsection{Integrability condition in Lemma \ref{lem:ksdnewform} \label{subsec:Integrability-condition-in}}

We give sufficient conditions for the integrability condition 
\begin{equation}
\EE_{(x,z),(x,z')\sim\tilde{\data}\otimes\tilde{\data}}\lvert H_{p}[(x,z),(x',z')]\rvert<\infty.\label{eq:integrablity-lemma-ksd}
\end{equation}
By the triangle inequality, we have 
\begin{align*}
 & \EE_{(x,z),(x,z')\sim\tilde{\data}\otimes\tilde{\data}}\lvert H_{p}[(x,z),(x',z')]\rvert\\
 & \leq\EE_{(x,z),(x,z')\sim\tilde{\data}\otimes\tilde{\data}}\left\{ \left|\score p(x|z)^{\top}\score p(x'|z')k(x,x')\right|+\left|\score p(x|z)^{\top}k_{1}(x',x)\right|\right.\\
 & \hphantom{=\EE_{(x,z),(x,z')\sim\tilde{\data}\otimes\tilde{\data}}\vphantom{}}\left.+\left|k_{1}(x,x')^{\top}\score p(x'|z')\right|+\left|k_{12}(x,x')\right|\right\} .\\
 & \leq\EE_{(x,x')\sim\data\otimes\data}\left\{ \EE_{z|x}\norm{\score p(x|z)}_{2}\EE_{z'|x'}\norm{\score p(x'|z')}_{2}k(x,x')+\EE_{z|x}\norm{\score p(x|z)}_{2}\norm{k_{1}(x',x)}_{2}\right.\\
 & \hphantom{=\EE_{(x,z),(x,z')\sim\tilde{\data}\otimes\tilde{\data}}\vphantom{}}\left.+\norm{k_{1}(x,x')}_{2}\EE_{z'|x'}\norm{\score p(x'|z')}_{2}+\left|k_{12}(x,x')\right|\right\} .
\end{align*}
 From this, the integrability condition is satisfied if 
\begin{enumerate}
\item $\EE_{(x,x')\sim\data\otimes\data}\left[\EE_{z|x}\norm{\score p(x|z)}_{2}\EE_{z'|x'}\norm{\score p(x'|z')}_{2}k(x,x')\right]<\infty,$ 
\item $\EE_{(x,x')\sim\data\otimes\data}\left[\EE_{z|x}\norm{\score p(x|z)}_{2}\norm{k_{1}(x',x)}_{2}\right]<\infty,$
and
\item $\EE_{(x,x')\sim\data\otimes\data}\left|k_{12}(x,x')\right|<\infty.$ 
\end{enumerate}
Note that for a finite domain\textbf{ }$\calX=\{0,\cdots,L-1\}^{D},$
these conditions are trivial, as 
\begin{align*}
\EE_{z|x}\norm{\score p(x|z)}_{2} & =\int\frac{\norm{\Delta_{x}p(x,z)}_{2}}{p(x,z)}\frac{p(x,z)}{p(x)}\dd P_{Z}(z)\\
 & \leq\frac{2}{p(x)}\leq\max_{x\in\mathcal{X}}\frac{2}{p(x)}.
\end{align*}
In what follows, we consider continuous-valued $x.$ As mentioned
in the main body, the third condition is mild, and fulfilled by e.g.,
the exponentiated quadratic kernel. Unfortunately we do not have a
handy test for the other requirements, and therefore deal with specific
scenarios below. To this end, we clarify the growth of $\EE_{z|x}\norm{\score p(x|z)}_{2}$
as a function of $x$ so that the user can check the required conditions
above.

\paragraph{Exponential families with bounded natural parameters. }

Let us first consider an exponential family likelihood $p(x|z)\propto\exp\bigl(\sum_{s=1}^{S}T_{s}(x)\eta_{s}(z)\bigr),$
$T_{s}:\mathbb{R}^{D}\to\mathbb{R},$ $\eta_{s}:\calZ\to\mathbb{R}$
for $1\leq s\leq S$. For this likelihood, we have 
\begin{align*}
\EE_{z|x}\norm{\score p(x|z)}_{2} & =\EE_{z|x}\norm{\sum_{s=1}^{S}\eta_{s}(z)\nabla_{x}T_{s}(x)}_{2}\\
 & \leq\sum_{s=1}^{S}\norm{\nabla_{x}T_{s}(x)}_{2}\EE_{z|x}|\eta_{s}(z)|.
\end{align*}
The conditions concerning the score function are satisfied provided
that we have
\begin{enumerate}
\item $\EE_{(x,x')\sim\data\otimes\data}\left[\norm{\nabla_{x}T_{s}(x)}_{2}\EE_{z|x}|\eta_{s}(z)|\norm{\nabla_{x}T_{s'}(x')}\EE_{z'|x'}|\eta_{s'}(z)|k(x,x')\right]<\infty$
\\
for any $s,s'\in\{1,...,S\}.$
\item $\EE_{(x,x')\sim\data\otimes\data}\left[\norm{\nabla_{x}T_{s}(x)}_{2}\EE_{z|x}|\eta_{s}(z)|\norm{k_{1}(x',x)}_{2}\right]<\infty$
for any $s\in\{1,...,S\}.$
\end{enumerate}
Let $a(x)\coloneqq\norm{\nabla_{x}T_{s}(x)}_{2}\EE_{z|x}|\eta_{s}(z)|.$
These conditions can be verified if both the kernel and its derivative
decay faster than $a(x)$. This can be challenging in practice as
we have a posterior expectation in $a(x)$ whose dependency on $x$
may not be easily analyzed. If we restrict the likelihood to have
bounded parameters (i.e., $\eta_{s}(z)$ is bounded), then the posterior
expectation is bounded, so we only need to choose a kernel such as 

\[
k(x,x')=\frac{1}{\sqrt{1+\sum_{s=1}^{S}\norm{\nabla_{x}T_{s}(x)}^{2}}}\frac{1}{\sqrt{1+\sum_{s=1}^{S}\norm{\nabla_{x}T_{s}(x')}_{2}^{2}}}\kappa(x,x')
\]
for a given kernel $\kappa.$ We summarize this observation in the
following lemma: 

\begin{lemma}

Consider a latent variable model with likelihood $p(x|z)\propto\exp\bigl(\sum_{s=1}^{S}T_{s}(x)\eta_{s}(z)\bigr),$
$T_{s}:\mathbb{R}^{D}\to\mathbb{R},$ $\eta_{s}:\calZ\to\mathbb{R}$
for $1\leq s\leq S$ and an arbitrary prior $P_{Z}.$ If $\sup_{z\in{\cal Z}}n_{s}(z)<\infty,$
then 
\[
\EE_{z|x}\norm{\score p(x|z)}_{2}\leq\max_{s}\sup_{z\in{\cal Z}}n_{s}(z)\cdot\sum_{s=1}^{S}\norm{\nabla_{x}T_{s}(x)}_{2}.
\]
Furthermore, for a given bounded kernel $\kappa,$ reweighting it
by 
\[
k(x,x')=\Bigl(1+\sum_{s=1}^{S}\norm{\nabla_{x}T_{s}(x)}^{2}\Bigr)^{-\delta}\Bigl(1+\sum_{s=1}^{S}\norm{\nabla_{x}T_{s}(x')}^{2}\Bigr)^{-\delta}\kappa(x,x')
\]
with some $\delta\geq1/2$ ensures that the condition \eqref{eq:integrablity-lemma-ksd}
holds. 

\end{lemma}

The boundedness assumption on the natural parameter $\eta_{s}(z)$
may not be satisfied for some models. For instance, if we consider
a normal mixture model with prior on the mean of the mixture component,
the support of the prior could be unbounded. 

\paragraph{Location-scale mixtures.}

Alternatively, we consider a location-scale family given by a radial-basis
function $\psi:[0,\infty)\to(0,\infty),$
\[
p\bigl(x|z=(\mu,\sigma^{2})\bigr)\propto\frac{1}{\sigma^{D}}\psi\left(\frac{\norm{x-\mu}_{2}^{2}}{\sigma^{2}}\right),
\]
 with prior $P_{\mu,\sigma}$ placed on the parameters. Here, we make
the following assumptions:

\begin{assumption}\label{assu:mixkernel}

The function $\psi$ is monotonically decreasing. The derivative-to-function
ratio $\lvert\psi'/\psi\rvert$ is uniformly upper-bounded by a constant
$M_{\psi}>0.$ \end{assumption}

\begin{assumption}\label{assu:mean-var-integrability}

The prior satisfies $\int\lVert\mu\rVert_{2}^{2}/\sigma^{4}\dd P_{\mu,\sigma}(\mu,\sigma)<\infty$
and $\int\sigma^{-4}\dd P_{\sigma}(\sigma)<\infty.$ Furthermore,
it satisfies
\[
\int\left(\frac{\psi(\lVert x-\mu\rVert_{2}^{2}/\sigma^{2})}{\sigma^{D}}\right)^{2}\dd P_{\mu,\sigma}(\mu,\sigma)<\infty.
\]

\end{assumption}

Isotropic normal- and Student's t-densities satisfy Assumption \ref{assu:mixkernel}.
As $\psi$ is monotonically decreasing, the third condition in Assumption
\ref{assu:mean-var-integrability} can be verified alternatively by
\[
\int\frac{1}{\sigma^{D}}\dd P_{\sigma}(\sigma)<\infty.
\]
These assumptions effectively prevent the density from being \emph{peaky}
and thus control the growth of the score function. 

Under these assumptions, we can quantify the growth of the score function
as follows. 

\begin{lemma}

Consider a latent variable model having likelihood
\[
p(x|z=(\mu,\sigma^{2}))\propto\frac{1}{\sigma^{D}}\psi\left(\frac{\norm{x-\mu}_{2}^{2}}{\sigma^{2}}\right)
\]
with radial-basis function $\psi:[0,\infty)\to(0,\infty)$ and prior
$P_{\mu,\delta}.$ Under Assumptions \ref{assu:mixkernel}, \ref{assu:mean-var-integrability},
in the limit of $\Verts x_{2}\to\infty,$ we have 
\[
\EE_{z|x}\norm{\score p(x|z)}_{2}=O(\Verts x_{2}).
\]
Furthermore, for a given bounded kernel $\kappa,$ reweighting it
by 
\[
k(x,x')=\bigl(1+\Verts x_{2}^{2}\bigr)^{-\delta}\bigl(1+\Verts{x'}_{2}^{2}\bigr)^{-\delta}\kappa(x,x')
\]
with some $\delta\geq1/2$ ensures that the condition \eqref{eq:integrablity-lemma-ksd}
holds. 

\end{lemma}

\begin{proof}
First, note that we have 

\begin{align*}
 & \EE_{z|x}\norm{\score p(x|z)}_{2}\\
 & =\int\frac{2}{\sigma^{D+2}}\lVert x-\mu\rVert_{2}\left|\frac{\psi'(\lVert x-\mu\rVert_{2}^{2}/\sigma^{2})}{\psi(\lVert x-\mu\rVert_{2}^{2}/\sigma^{2})}\right|\frac{\psi(\lVert x-\mu\rVert_{2}^{2}/\sigma^{2})}{\int\frac{1}{\sigma^{D}}\psi\left(\lVert x-\mu\rVert_{2}^{2}/\sigma^{2}\right)\dd P_{\mu,\sigma}(\mu,\sigma)}\dd P_{\mu,\sigma}(\mu,\sigma)\\
 & \leq2M_{\psi}\underbrace{\frac{\int\frac{1}{\sigma^{D+2}}\lVert x-\mu\rVert_{2}\psi(\lVert x-\mu\rVert_{2}^{2}/\sigma^{2})\dd P_{\mu,\sigma}(\mu,\sigma)}{\int\frac{1}{\sigma^{D}}\psi\left(\lVert x-\mu\rVert_{2}^{2}/\sigma^{2}\right)\dd P_{\mu,\sigma}(\mu,\sigma)}}_{g(x)}.
\end{align*}
We therefore show that the growth of the function $g$ is of the order
of $\norm x_{2}$ by examining the limit
\[
\lim_{\norm x_{2}\to\infty}\frac{g(x)}{\norm x_{2}}.
\]
Let $\tilde{p}_{x}(\mu,\sigma)=\psi(\lVert x-\mu\rVert_{2}^{2}/\sigma^{2})/\sigma^{D}.$
For the numerator of $g,$ note that we have
\begin{align*}
\int\frac{\lVert x-\mu\rVert_{2}}{\sigma^{2}}\frac{\psi(\lVert x-\mu\rVert_{2}^{2}/\sigma^{2})}{\sigma^{D}}\dd P_{\mu,\sigma}(\mu,\sigma) & \leq\sqrt{\int\bigl(\lVert x-\mu\rVert_{2}/\sigma^{2}\bigr)^{2}\dd P_{\mu,\sigma}(\mu,\sigma)}\cdot\lVert\tilde{p}_{x}\rVert_{L_{2}(P_{\mu,\sigma})}\\
 & \overset{\mathrm{Asm.}\ref{assu:mean-var-integrability}}{<}\infty,
\end{align*}
where the first inequality is given by the Cauchy-Schwarz inequality.
Thus,
\begin{align}
\frac{g(x)}{\norm x_{2}} & \leq\frac{\sqrt{\int\bigl(\lVert x-\mu\rVert_{2}/\sigma^{2}\bigr)^{2}\dd P_{\mu,\sigma}(\mu,\sigma)}\cdot\lVert\tilde{p}_{x}\rVert_{L_{2}(P_{\mu,\sigma})}}{\lVert x\rVert_{2}\lVert\tilde{p}_{x}\rVert_{L_{1}(P_{\mu,\sigma})}}\nonumber \\
 & \leq\sqrt{\frac{\int\bigl(\lVert x-\mu\rVert_{2}/\sigma^{2}\bigr)^{2}\dd P_{\mu,\sigma}(\mu,\sigma)}{\lVert x\rVert_{2}^{2}}}\nonumber \\
 & \leq\sqrt{\int\frac{(\lVert x\rVert_{2}+\lVert\mu\rVert_{2})^{2}}{\lVert x\rVert_{2}^{2}}\frac{1}{\sigma^{4}}\dd P_{\mu,\sigma}(\mu,\sigma)}.\label{eq:gbound-rhs}
\end{align}
The second line is obtained with the relation between the $L_{1}$-
and $L_{2}$-norms: $\lVert\tilde{p}_{x}\rVert_{L_{1}(P_{\mu,\sigma})}\leq\lVert\tilde{p}_{x}\rVert_{L_{2}(P_{\mu,\sigma})}.$
The function inside the integral in \eqref{eq:gbound-rhs} is monotonically
decreasing at each $(\mu,\sigma)$ as 
\[
\lim_{\lVert x\rVert_{2}\to\infty}\left(1+2\frac{\lVert\mu\rVert_{2}}{\lVert x\rVert_{2}}+\frac{\lVert\mu\rVert_{2}^{2}}{\lVert x\rVert_{2}^{2}}\right)\frac{1}{\sigma^{4}}\searrow\frac{1}{\sigma^{4}},
\]
and by Assumption \ref{assu:mean-var-integrability}, it is integrable
when $\lVert x\rVert_{2}=1.$ Thus, by the monotone convergence theorem,
we have 
\[
\lim_{\lVert x\rVert_{2}\to\infty}\frac{g(x)}{\norm x_{2}}\leq\sqrt{\int\frac{1}{\sigma^{4}}\dd P_{\sigma}(\sigma)},
\]
where the upper-bound is finite by Assumption \ref{assu:mean-var-integrability},
indicating that $g(x)=O(\lVert x\rVert_{2}).$ Therefore, for the
location-scale family, we have that the score part grows at the speed
at most of $\norm x_{2}$. Therefore, modifying kernel $\kappa$ by
\[
k(x,x')=\bigl(1+\Verts x_{2}^{2}\bigr)^{-\delta}\bigl(1+\Verts{x'}_{2}^{2}\bigr)^{-\delta}\kappa(x,x')\ \text{with }\delta\geq\frac{1}{2}
\]
ensures that the decay of the kernel is as fast as the score part.
\qed

\end{proof}

\subsection{Convergence assumption in Theorem \ref{thm:mcmcustat} \label{subsec:conv-assumption-ustat}}

To apply Theorem \ref{thm:mcmcustat} to the KSD estimate, we need
to verify that the bias 
\begin{equation}
\left|\EE_{y,y'\sim\tilde{R}_{t}\otimes\tilde{R}_{t}}\bar{H}_{p}(y,y')-\EE_{y,y'\sim\tilde{R}\otimes\tilde{R}}H_{p}(y,y')\right|\label{eq:ksd-bias}
\end{equation}
decays at some rate. Here, $\tilde{R}_{t}(\dd(x,\mathbf{z}))=P_{Z}^{(t)}(\dd\mathbf{z}|x)\data(\dd x)$
(see the paragraph before Theorem \ref{thm:mcmcustat} for the notation). 

In what follows, we assume that the MCMC sampler satisfies 
\begin{equation}
\left|\EE_{\mathbf{z}^{(t)}|x}\barscore{p,d}(x|\mathbf{z}^{(t)})-\score{p,d}(x)\right|\leq r(t,x)\coloneqq M(x)r(t),\ \text{for }d=1,\dots,D,\label{eq:scoremcmcbound}
\end{equation}
for some function $r(t):\mathbb{N}\to(0,1]$ decreasing to $0$ in
$t$ and a positive function $M(x),$ where $\barscore{p,d}$ denotes
the $d$-th element of the conditional score $\barscore p$ (the same
rule applies to $\score p$). There is usually dependency in $M(x)$
on the initial state of the chain, but this is suppressed here for
simplicity. This assumption can be understood as the convergence of
the $t$-step transition law of the Markov chain $P_{Z}^{(t)}(\dd z|x)$
to the target $P_{Z}(\dd z|x)$ in terms of the test functions $\score{p,d}(x|\cdot).$
The convergence can then be checked with the assumption that the test
functions belong to a certain class, and the upper-bound of \eqref{eq:scoremcmcbound}
can be stated as a worst-case convergence rate in that class. For
example, the convergence in total variation distance corresponds to
the class of nonnegative measurable function bounded uniformly by
1. 

A specification of the decay rate $r$ and the function class relates
the condition \eqref{eq:scoremcmcbound} to standard notions of ergodicity
studied in the Markov chain literature \citep{Roberts_2004,Meyn_2009}.
For instance, suppose that we have the geometric rate $r(t)=\rho^{t}$
for some $0<\rho<1,$ and that the function class is such that all
members are measurable and bounded uniformly by a function $V:{\cal Z}\to[1,\infty).$
Then, the corresponding convergence is known as $V$-uniform convergence
(if $V\equiv1,$ this corresponds to the uniform ergodicity) \citep[Chapter 16]{Meyn_2009}. 

We can reduce the convergence of the bias \eqref{eq:ksd-bias} to
that of the score estimate \eqref{eq:scoremcmcbound}. To see this
assertion, we first note that 
\begin{align*}
 & \left|\text{The first term of }\left(\EE_{y,y'\sim\tilde{R}_{t}\otimes\tilde{R}_{t}}\bar{H}_{p}(y,y')-\EE_{y,y'\sim\tilde{R}\otimes\tilde{R}}\bar{H}_{p}(y,y')\right)\right|\\
 & \leq\EE_{x,x'}\left|k(x,x')\sum_{d=1}^{D}\left\{ \EE_{\mathbf{z}^{(t)}|x}\barscore{p,d}(x|\mathbf{z}^{(t)})\EE_{\mathbf{z}'^{(t)}|x'}\barscore{p,d}(x'|\mathbf{z}'^{(t)})-\score{p,d}(x)\score{p,d}(x')\right\} \right|\\
 & \leq\sum_{d=1}^{D}\EE_{x,x'}\left[k(x,x')\left\{ \left|\EE_{\mathbf{z}^{(t)}|x}\barscore{p,d}(x|\mathbf{z}^{(t)})\right|r(t,x')+r(t,x)\left|\score{p,d}(x')\right|\right\} \right]\\
 & \leq\sum_{d=1}^{D}\EE_{x,x'}\left[k(x,x')\left\{ r(t,x)r(t,x')+r(t,x')\left|\score{p,d}(x)\right|+r(t,x)\left|\score{p,d}(x')\right|\right\} \right]\\
 & \leq Dr(t)^{2}\EE_{x,x'}[k(x,x')M(x)M(x')]+2r(t)\sum_{d=1}^{D}\EE_{x.x'}\left[k(x,x')M(x')\left|\score{p,d}(x')\right|\right].
\end{align*}
That is, if $\EE_{x,x'}[k(x,x')M(x)M(x')]<\infty$ and $\sum_{d=1}^{D}\EE_{x.x'}\left[k(x,x')M(x')\left|\score{p,d}(x')\right|\right]<\infty,$
the difference in the first terms can be bounded by $r(t).$ A similar
argument can be applied to the second and third terms. The constant
$M(x)$ in the bound in \eqref{eq:scoremcmcbound} often depends on
certain properties of the target $P_{Z}(\dd z|x).$ If those properties
hold uniformly over $x$ (i.e., $M(x)$ can be treated as a constant),
then the validation of these conditions is straightforward. A concrete
example of such properties is strong log-convexity and having a Lipschitz
continuous gradient of the target (assuming the target is given by
a density $p(z|x)$) \citep{Dal17,DwivediChenWainwrightEtAl2019Log,Bou_Rabee_2020}.
In this situation, the bias \eqref{eq:ksd-bias} is determined by
the rate $r(t)$ at which the score function converges. Hence, $t$
has to grow as a function of $n$ such that $t(n)=O\{r^{-1}(n^{-s})\}$
with $s>1/2$ to apply Theorem \ref{thm:mcmcustat}. 

\subsection{The maximum mean discrepancy relative goodness-of-fit test \label{subsec:mmd-description}}

We provide the detail of the MMD relative goodness-of-fit test proposed
by \citet{BouBelBlaAntGre2016} and describe our modification to correct
for underestimation of the variances required in the test (see Appendix
\ref{subsec:exp-closemodels}). Recall that the MMD between two probability
distributions $P$ and $R$ is defined as the following IPM: 
\[
\mmd PR\coloneqq\sup_{\Verts f_{{\cal F}_{k}}\leq1}\left|\int f\dd P-\int f\dd R\right|,
\]
where ${\cal F}_{k}$ is the RKHS of (scalar-valued) kernel $k:\calX\times{\cal X\to\mathbb{R}}.$
The MMD can be written in terms of the kernel as 
\[
\mmdsq PR=\EE_{x,x'\sim P\otimes P}[k(x,x')]-2\EE_{x,z\sim P\otimes R}[k(x,z)]+\EE_{z,z'\sim R\otimes R}[k(z,z')]
\]
if $\EE_{x\sim P}[k(x,x)^{1/2}]<\infty$ and $\EE_{z\sim R}[k(z,z)^{1/2}]<\infty.$
Given mutually independent samples $\{x_{i}\}_{i=1}^{n_{P}}\iidsim P$
and $\{z_{i}\}_{i=1}^{n_{R}}\iidsim R,$ we can estimate the squared
MMD with a two-sample U-statistic \citep[p. 131]{Kowalski_2007}
\[
\mmdsqhat PR=\frac{1}{{n_{P} \choose 2}}\frac{1}{{n_{R} \choose 2}}\sum_{j_{1}<j_{2}}\sum_{i_{1}<i_{2}}\ell(x_{i_{1}},x_{i_{1}};z_{j_{1}},z_{j_{2}})
\]
with 
\[
\ell(x,x';z,z')=k(x,x')+k(z,z')-\frac{1}{2}(k(x,z)+k(x,z')+k(x',z)+k(x',z')).
\]
Note that this statistic is equal to the unbiased estimator of \citet[Eq. 3]{GreBorRasSchSmo2012}. 

Given a sample $\{z_{i}\}_{i=1}^{n_{R}}\iidsim R$ and two competing
models $P,Q,$ the relative MMD test is defined as follows: 
\[
\begin{aligned}H_{0} & :\mmd PR\leq\mmd QR\ \text{(null hypothesis),}\\
H_{1} & :\mmd PR>\mmd QR\ \text{(\text{alternative)}}.
\end{aligned}
\]
Note that here we use a different symbol for the data in the main
body (there, the data variable is $x$). Their procedure does not
consider the case where the sample size $n_{R}$ does not match the
sizes $n_{P},$ $n_{Q}$ of respective samples $\{x_{i}\}_{i=1}^{n_{P}},$
$\{y_{i}\}_{i=1}^{n_{Q}}$ from $P$ and $Q.$ Therefore, we provide
the test procedure accommodating this case. 

The test statistic is defined by the difference between estimates
of the squared MMDs 

\[
\mmdsqhat PR-\mmdsqhat QR=\frac{1}{{n_{P} \choose 2}}\frac{1}{{n_{Q} \choose 2}}\frac{1}{{n_{R} \choose 2}}\sum_{i_{x1}<i_{x2}}\sum_{i_{y1}<i_{y2}}\sum_{i_{z1}<i_{z2}}\ell_{\mathrm{diff}}(x_{i_{x1}},x_{i_{x2}};y_{i_{y1}},y_{i_{y2}};z_{i_{z1}},z_{i_{z2}}),
\]
where 
\[
\ell_{\mathrm{diff}}(x,x';y,y';z,z')=\ell(x,x';z,z')-\ell(y,y';z,z').
\]
This statistic is a three-sample U-statistic; its asymptotic distribution
is normal under the same assumptions on the relations between the
models and data distribution as in the main body (the second paragraph
of Section \ref{subsec:Problem-setup}). Let $n_{\mathrm{sum}}=n_{P}+n_{Q}+n_{R}.$
Assume the following growth condition on the sample sizes 
\[
\frac{n_{\mathrm{sum}}}{n_{P}}\to\rho_{P}^{2},\ \frac{n_{\mathrm{sum}}}{n_{Q}}\to\rho_{Q}^{2},\ \text{and }\frac{n_{\mathrm{sum}}}{n_{R}}\to\rho_{R}^{2}
\]
with finite constants $\rho_{P},$ $\rho_{Q},$ and $\rho_{R}.$ Assume
$\EE[\ell_{\mathrm{diff}}(x,x';y,y';z,z')^{2}]<\infty.$ Then, according
to \citet[Theorem 3, p.168]{Kowalski_2007}, the limit of $(n_{P},n_{Q},n_{R})$
gives 
\[
\sqrt{n_{\mathrm{sum}}}\left[\Bigl\{\mmdsqhat PR-\mmdsqhat QR\Bigr\}-\{\mmdsq PR-\mmdsq QR\}\right]\dto\Normal 0{\sigma_{P,Q,R}^{2}},
\]
where 
\[
\sigma_{P,Q,R}^{2}=4\left(\rho_{P}^{2}\var_{x}\left[\EE[\ell_{\mathrm{diff}}\vert x]\right]+\rho_{Q}^{2}\var_{y}\left[\EE[\ell_{\mathrm{diff}}\vert y]\right]+\rho_{R}^{2}\var_{z}\left[\EE[\ell_{\mathrm{diff}}\vert z]\right]\right),
\]
with 
\[
\EE[\ell_{\mathrm{diff}}\vert x]=\EE[\ell_{\mathrm{diff}}(x,x';y,y';z,z')\vert x],
\]
and the same notation applies to the conditional expectations of $\ell_{\mathrm{diff}}$
of $y$ and $z.$ 

With a consistent estimator $\hat{\sigma}_{P,Q,R},$ \citet{BouBelBlaAntGre2016}
proposed an asymptotically level-$\alpha$ test that rejects the null
hypothesis if $\mmdsqhat PR-\mmdsqhat QR\geq(\hat{\sigma}_{P,Q,R}/\sqrt{n_{\mathrm{sum}}})\cdot\tau_{1-\alpha}$
with $\tau_{1-\alpha}$ the $(1-\alpha)$-quantile of the standard
normal distribution. We found that the estimator given by \citet{BouBelBlaAntGre2016}
tends to underestimate the target variance, and that their test exceeded
the nominal level in some problems where two models are close to each
other. We therefore consider another estimator for our experiments,
as described below. 

\paragraph{Variance estimators}

Following are the variances required for $\sigma_{P,Q,R}^{2}:$ 

\begin{align*}
\var_{x}\left[\EE[\ell_{\mathrm{diff}}\vert x]\right] & =\var_{x}\left[\EE[\ell(x,x';z,z')\vert x]\right]\\
 & =\var_{x}\left[\EE_{x',z}[k(x,x')-k(z,x)\vert x]\right],\\
\var_{y}\left[\EE[\ell_{\mathrm{diff}}\vert y]\right] & =\var_{y}\left[\EE[\ell(y,y';z,z')\vert y]\right]\\
 & =\var_{y}\left[\EE_{z,y'}[k(y,y')-k(z,y)\vert y]\right],\ \text{and }\\
\var_{z}\left[\EE[\ell_{\mathrm{diff}}\vert z]\right] & =\var_{z}\left[\EE[\ell(x,x';z,z')-\ell(y,y';z,z')\vert z]\right]\\
 & =\var_{z}\left[\EE_{x,y}[k(z,x)-k(z,y)\vert z]\right].
\end{align*}
The first two quantities are symmetric in terms of $x$ and $y,$
and therefore we only need to consider one of them. An estimator for
the first variance is given by 
\begin{align}
\var_{x}\left[\underbrace{\EE_{x',z}[k(x,x')-k(z,x)\vert x]}_{f(x)}\right] & \approx\frac{1}{n_{P}(n_{P}-1)}\sum_{i\neq j}\frac{\left(\hat{f}_{i}(x_{i})-\hat{f}_{j}(x_{j})\right)^{2}}{2},\label{eq:mmd-variance-x}
\end{align}
where $\hat{f}_{i}(x_{i})$ is an approximation to $f(x_{i})$ defined
by 
\[
\hat{f}_{i}(x_{i})=\frac{1}{(n_{P}-1)}\sum_{l\neq i}k(x_{l},x_{i})-\frac{1}{n_{R}}\sum_{l=1}^{n_{R}}k(z_{l},x_{i}).
\]
We similarly estimate the third variance using 
\begin{align}
\var_{z}\left[\underbrace{\EE_{x,y}[k(z,x)-k(z,y)\vert z]}_{g(z)}\right] & \approx\frac{1}{n_{R}(n_{R}-1)}\sum_{i\neq j}\frac{\left(\hat{g}_{i}(z_{i})-\hat{g}_{j}(z_{j})\right)^{2}}{2},\label{eq:mmd-variance-z}
\end{align}
where 
\[
\hat{g}_{i}(z_{i})=\frac{1}{n_{P}}\sum_{l=1}^{n_{P}}k(x_{l},z_{i})-\frac{1}{n_{Q}}\sum_{l=1}^{n_{Q}}k(y_{l},z_{i})\left(\approx g(z_{i})\right).
\]
The estimators \eqref{eq:mmd-variance-x} and \eqref{eq:mmd-variance-z}
are simple to compute and always nonnegative. The consistency of these
estimators can be checked by expanding the expressions. The derivation
is tedious, and therefore we only prove it for \eqref{eq:mmd-variance-x}. 

\begin{lemma}

Assume $\EE_{x,x'\sim P\otimes P}[k(x,x')^{2}]<\infty$ and $\EE_{x,x'\sim P\otimes R}[k(x,z)^{2}]<\infty.$
Then, Eq. \eqref{eq:mmd-variance-x} estimates 
\[
\var_{x}\left[\EE_{x',z}[k(x,x')-k(z,x)\vert x]\right]
\]
consistently in the limit of $(n_{P},n_{R})$ such that the ratio
$n_{P}/n_{R}$ converges to a finite constant. 

\end{lemma}

\begin{lemma}

Assume $\EE_{x,z\sim Q\otimes R}[k(x,z)^{2}]<\infty,$ $\EE_{y,z\sim P\otimes R}[k(y,z)^{2}]<\infty,$
and $\EE_{z,z'\sim R\otimes R}[k(z,z')^{2}]<\infty.$ Then, Eq. \eqref{eq:mmd-variance-z}
estimates 
\[
\var_{z}\left[\EE_{x,y}[k(z,x)-k(z,y)\vert z]\right]
\]
consistently in the limit of $(n_{P},n_{Q},n_{R})$ such that the
ratios $n_{P}/n_{R}$ and $n_{Q}/n_{R}$ converge to finite constants. 

\end{lemma}

\begin{proof}

Note that the estimator \eqref{eq:mmd-variance-x} is the (approximate)
sample variance
\[
\frac{1}{n_{P}-1}\left\{ \sum_{i=1}^{n_{P}}\hat{f}_{i}(x_{i})^{2}-n_{P}\left(\frac{1}{n_{P}}\sum_{i=1}^{n_{P}}\hat{f}_{i}(x_{i})\right)^{2}\right\} .
\]
Showing the consistency is equivalent to proving the following limits
(the symbol $\pto$ denotes convergence in probability): 
\[
\frac{1}{n_{P}}\sum_{i=1}^{n_{P}}\hat{f}_{i}(x_{i})\pto\EE_{x}[f(x)]\ \text{and\ }\frac{1}{n_{P}}\sum_{i=1}^{n_{P}}\hat{f}_{i}(x_{i})^{2}\pto\EE_{x}[f(x)^{2}].
\]
The first limit is immediate as 
\[
\frac{1}{n_{P}}\sum_{i=1}^{n_{P}}\hat{f}_{i}(x_{i})=\frac{1}{n_{P}(n_{P}-1)}\sum_{l\neq i}k(x_{l},x_{i})-\frac{1}{n_{P}}\frac{1}{n_{R}}\sum_{l=1}^{n_{R}}k(z_{l},x_{i}),
\]
which is a U-statistic of $\EE_{x}[f(x)].$ 

For the second convergence claim, we expand the expressions as follows:
\begin{align*}
\hat{f}_{i}(x_{i})^{2} & =\frac{1}{(n_{P}-1)^{2}}\sum_{j\neq i}\sum_{l\neq i}k(x_{j},x_{i})k(x_{l},x_{i})+\frac{1}{n_{R}^{2}}\sum_{j,j'}k(z_{j},x_{i})k(z_{j'},x_{i})-2\frac{1}{(n_{P}-1)n_{R}}\sum_{j\neq i}\sum_{l=1}^{n_{R}}k(x_{j},x_{i})k(z_{l},x_{i})\\
 & =\frac{(n_{P}-2)}{(n_{P}-1)}\frac{1}{(n_{P}-1)(n_{P}-2)}\sum_{j\neq i}\sum_{l\neq i,j}k(x_{j},x_{i})k(x_{l},x_{i})-\underbrace{\frac{1}{(n_{P}-1)^{2}}\sum_{j\neq i}k(x_{j},x_{i})^{2}}_{\mathrm{A}}+\frac{1}{n_{R}^{2}}\sum_{j\neq j'}k(z_{j},x_{i})k(z_{j'},x_{i})\\
 & \hphantom{=}+\underbrace{\frac{1}{n_{R}^{2}}\sum_{j}k(z_{j},x_{i})^{2}}_{\mathrm{B}}-2\frac{1}{(n_{P}-1)n_{R}}\sum_{j\neq i}\sum_{l=1}^{n_{R}}k(x_{j},x_{i})k(z_{l},x_{i}),\ \text{and}\\
\EE[f(x_{i})^{2}] & =\EE_{x_{i}}\left[\EE_{x'}[k(x_{i},x')\vert x_{i}]^{2}+\EE_{z}[k(z,x_{i})\vert x_{i}]^{2}-2\EE_{x'}[k(x_{i},x')\vert x_{i}]\EE_{z}[k(z,x_{i})\vert x_{i}]\right].
\end{align*}
Note that the terms $n_{P}^{-1}\sum_{i=1}^{n_{P}}\hat{f}_{i}(x_{i})^{2}$
corresponding to A and B above vanish in the limit, since by the law
of large numbers for U-statistics, 
\begin{align*}
\frac{1}{n_{P}(n_{P}-1)}\sum_{i=1}^{n_{P}}\sum_{j\neq i}k(x_{j},x_{i})^{2} & \pto\EE_{x}[k(x,x_{i})^{2}]<\infty\ \text{and }\frac{1}{n_{P}n_{R}}\sum_{i,j}k(z_{j},x_{i})^{2}\pto\EE[k(x,z)^{2}]<\infty.
\end{align*}
The other three terms are U-statistics (up to scaling negligible in
the limit) of their counterparts in $\EE_{x}[f(x)^{2}].$ Thus, by
the same reasoning, the second limit holds. \qed

\end{proof}

\subsection{MMD and KSD for Gaussian distributions\label{sec:MMD-and-KSD-pop}}

We provide explicit forms of MMD and KSD measured for Gaussian distributions.
These results are used in constructing the PPCA experiment in Section
\ref{sec:Experiments} in the main body. In the process, we also obtain
an understanding of the role of the reproducing kernel in the KSD. 

\subsubsection{MMD }

This section provides an explicit formula for MMD between two normal
distributions, defined by the exponentiated quadratic kernel 
\[
k(x,x')=\exp\left(\frac{-\norm{x-x'}_{2}^{2}}{2\lambda^{2}}\right),\text{where }\lambda>0.
\]
The MMD expression in this setting has been shown \citep[see e.g.,][Example 3]{Sriperumbudur_2012}.
We use this formula to compute the difference of MMDs so that we can
construct a problem as in Section \ref{subsec:PPCA}. 

\begin{lemma}

Let $k(x,x')=\exp\Big\{-\norm{x-x'}_{2}^{2}/(2\lambda^{2})\Big\}$
with $\lambda>0.$ For three $D$-dimensional Gaussian distributions
$P=\Normal{\bfzero}{\Sigma_{p}},$ $Q=\Normal{\bfzero}{\Sigma_{q}},$
and $R=\Normal{\bfzero}{\Sigma_{r}},$ we have

\begin{align*}
\mmdsq PR-\mmdsq QR & =\lambda^{D}\left\{ \frac{1}{\sqrt{\bigl|2\Sigma_{p}+\lambda^{2}I\bigr|}}-\frac{1}{\sqrt{2\Sigma_{q}+\lambda^{2}I\bigr|}}\right.\\
 & \hphantom{=\lambda^{D}\quad}-2\left.\left(\frac{1}{\sqrt{\bigl|\Sigma_{p}+\Sigma_{r}+\lambda^{2}I)\bigr|}}-\frac{1}{\sqrt{\bigl|\Sigma_{q}+\Sigma_{r}+\lambda^{2}I\bigr|}}\right)\right\} ,
\end{align*}
where $\lvert\cdot\rvert$ denotes the determinant. 

\end{lemma}

Note that we can numerically evaluate this expression given covariance
matrices. For completeness, we provide a proof below. 

\begin{proof}

Recall that the MMD between two distributions $P,R$ is given by 
\[
\mmdsq PR=\EE_{(x,x')\sim P\otimes P}[k(x,x')]-2\EE_{x\sim P,x'\sim R}[k(x,x')]+\EE_{(x,x')\sim R\otimes R}[k(x,x')].
\]
Let $p(x)=\Normal{x;\mu_{p}}{\Sigma_{p}},$ $r(x)=\Normal{x;\mu_{r}}{\Sigma_{r}}.$
Note that when properly scaled, the exponentiated quadratic kernel
can be also seen as a Gaussian density function. Therefore, by convolution,
we have 
\begin{align*}
\EE_{x'\sim P}[k(x,x')] & =(2\pi\lambda^{2})^{D/2}\Normal{x;\mu_{p}}{\Sigma_{p}+\lambda^{2}I}.
\end{align*}
Then, the first term is 
\begin{align*}
 & \EE_{x,x'\sim P\otimes P}[k(x,x')]=\int\EE_{x'\sim P}[k(x,x')]\Normal{x;\mu_{p}}{\Sigma_{p}}\dd x\\
 & =(2\pi\lambda^{2})^{D/2}\int\Normal{x;\mu_{p}}{\Sigma_{p}+\lambda^{2}}\Normal{x;\mu_{p}}{\Sigma_{p}}\dd x\\
 & =\lambda{}^{D}\sqrt{\frac{\bigl|\tilde{\Sigma}_{p}\bigr|}{\bigl|(\Sigma_{p}+\lambda^{2}I)\bigr|\bigl|\Sigma_{p}\bigr|}}\exp\left(\frac{1}{2}\left\{ \tilde{\mu}_{p}^{\top}\tilde{\Sigma}_{p}^{-1}\tilde{\mu}_{p}-\mu_{p}^{\top}(\Sigma_{p}+\lambda^{2}I)^{-1}\mu_{p}-\mu_{p}^{\top}\Sigma_{p}^{-1}\mu_{p}\right\} \right),
\end{align*}
where $\lvert\cdot\rvert$denotes the determinant, and
\[
\tilde{\Sigma}_{p}=\bigl((\Sigma_{p}+\lambda^{2}I)^{-1}+\Sigma_{p}^{-1}\bigr)^{-1},\quad\tilde{\mu}_{p}=\tilde{\Sigma}_{p}\bigl((\Sigma_{p}+\lambda^{2}I)^{-1}\mu_{p}+\Sigma_{p}^{-1}\mu_{p}\bigr).
\]
The second term is 
\begin{align*}
 & \EE_{x\sim P,x'\sim R}[k(x,x')]\\
 & =(2\pi\lambda^{2})^{D/2}\int\Normal{x;\mu_{p}}{\Sigma_{p}+\lambda^{2}I}\Normal{x;\mu_{r}}{\Sigma_{r}}\dd x\\
 & =\lambda{}^{D}\sqrt{\frac{\bigl|\Sigma_{p,r}\bigr|}{\bigl|(\Sigma_{p}+\lambda^{2}I)\bigr|\bigl|\Sigma_{r}\bigr|}}\exp\left(\frac{1}{2}\left\{ \mu_{p,r}\Sigma_{p,r}^{-1}\mu_{p,r}-\mu_{p}^{\top}(\Sigma_{p}+\lambda^{2}I)^{-1}\mu_{p}-\mu_{r}^{\top}\Sigma_{r}^{-1}\mu_{r}\right\} \right),
\end{align*}
where 
\[
\Sigma_{p,r}=\bigl(\Sigma_{r}^{-1}+(\Sigma_{p}+\lambda^{2}I)^{-1}\bigr)^{-1},\quad\mu_{p,r}=\Sigma_{p,r}\left\{ (\Sigma_{p}+\lambda^{2}I)^{-1}\mu_{p}+\Sigma_{r}^{-1}\mu_{r}\right\} .
\]
The third term is similarly obtained, but its form is not necessary
for comparing models. We then impose the condition $\mu_{p}=\mu_{r}=\bfzero.$
In this case, we have
\begin{align*}
\mmdsq PR & =\lambda{}^{D}\left(\sqrt{\frac{\bigl|\tilde{\Sigma}_{p}\bigr|}{\bigl|(\Sigma_{p}+\lambda^{2}I)\bigr|\bigl|\Sigma_{p}\bigr|}}-2\sqrt{\frac{\bigl|\Sigma_{p,r}\bigr|}{\bigl|(\Sigma_{p}+\lambda^{2}I)\bigr|\bigl|\Sigma_{r}\bigr|}}\right)+\EE_{(x,x)'\sim R\otimes R}[k(x,x')]\\
 & =\lambda{}^{D}\left(\frac{1}{\sqrt{\bigl|2\Sigma_{p}+\lambda^{2}I\bigr|}}-2\frac{1}{\sqrt{\bigl|(\Sigma_{p}+\Sigma_{r}+\lambda^{2}I\rvert}}\right)+\EE_{(x,x)'\sim R\otimes R}[k(x,x')]
\end{align*}
Thus, substituting three Gaussian distributions $P=\Normal{\bfzero}{\Sigma_{p}},$
$Q=\Normal{\bfzero}{\Sigma_{q}},$ and $R=\Normal{\bfzero}{\Sigma_{r}},$
we obtain the desired equality. \qed

\end{proof}

\subsubsection{KSD \label{subsec:KSD-Gaussian}}

The KSD can be equivalently written in terms of the difference of
score functions \citep[Definition 3.2]{LiuLeeJor2016}: 
\[
\ksdsq PR=\EE_{x_{1},x_{2}\sim R\otimes R}\left[(\score p(x_{1})-\score r(x_{1}))^{\top}(\score p(x_{2})-\score r(x_{2}))k(x_{1},x_{2})\right].
\]
For two Gaussian densities $p(x)=\Normal{x;\bfzero}{\Sigma_{p}},$
$r(x)=\Normal{x;\bfzero}{\Sigma_{r}},$ the difference between their
score functions is

\begin{align*}
\score p(x)-\score r(x) & =-(\Sigma_{p}^{-1}-\Sigma_{r}^{-1})x.
\end{align*}
Therefore,

\begin{align*}
\ksdsq PR & =\EE_{x_{1},x_{2}\sim R\otimes R}\left[(\score p(x_{1})-\score r(x_{1}))^{\top}(\score p(x_{2})-\score r(x_{2}))k(x_{1},x_{2})\right]\\
 & =\EE_{x_{1},x_{2}\sim R\otimes R}\left[\inner{\Bigl(\Sigma_{p}^{-1}-\Sigma_{r}^{-1}\Bigr)^{2}}{k(x_{1},x_{2})x_{1}x_{2}^{\top}}\right]\\
 & =\inner{\Bigl(\Sigma_{p}^{-1}-\Sigma_{r}^{-1}\Bigr)^{2}}{\underbrace{\EE_{x_{1},x_{2}\sim R\otimes R}[k(x_{1},x_{2})x_{1}x_{2}{}^{\top}]}_{M_{k,R}}},
\end{align*}
where $\la\cdot,\cdot\ra$ denotes the matrix inner product, and $M_{k,R}$
is a matrix depending on the kernel $k$ and the data distribution
$R.$  Therefore, it can be informally understood that the KSD depends
on the difference between the covariances $\Sigma_{p}$ and $\Sigma_{r}$;
if $\Sigma_{p}$ is given by additive perturbation as $\Sigma_{r}+E$,
the difference depends on the perturbation matrix $E.$ Note that
in the PPCA experiments, the perturbation matrix is an increasing
function of $\delta,$ element-wise. 

\subsection{Kernel choice and KSD: Gaussian models and data}\label{subsec:kernel-Gauss-data}

We illustrate how kernel choice affects the sensitivity of the KSD.
We consider the following setting: 
\[
P\sim\Normal 0{\diag(1,,1,\dots,1}),R\sim{\cal N}(0,\diag(\sigma_{1}^{2},\dots,1))
\]
for some positive $\sigma_{1}\neq1.$ The model $P$ misestimates
the variance of the first coordinate of the data. Let us consider
the effect of a parameter choice for the IMQ kernel. We specifically
compare the following kernels: 
\[
k_{\mathrm{IMQ}}(x,y)=\frac{1}{\sqrt{1+\Verts{x-y}_{2}^{2}}},\ k_{\mathrm{IMQ}}^{\mathrm{scale}}(x,y)=\frac{1}{\sqrt{1+\sum_{i>1}(x^{i}-y^{i})^{2}+\sigma_{1}^{-2}(x^{1}-y^{1})^{2}}},
\]
where $x^{1}$ denotes the first coordinate of $x.$ The latter kernel
can be considered as the (preconditioned) IMQ kernel with dimension-wise
scaling $\Lambda=\diag(\sigma_{1}^{2},1,\dots,1)$ where the scale
is determined by the dimension-wise variance of the data. The KSDs
corresponding to these kernel choices are given as follows:

\begin{align*}
 & \ksd P{R;k_{\mathrm{IMQ}}}^{2}=\left(\sigma_{1}-1\right)^{2}\underbrace{\EE_{X,Y\sim{\cal N}(0,I)}\left[\frac{X^{1}Y^{1}}{\sqrt{1+\sigma_{1}^{2}(X^{1}-Y^{1})^{2}+\sum_{i>1}(X^{i}-Y^{i})^{2}}}\right]}_{\EE\bigl[k_{\mathrm{IMQ}}(X,Y)X^{1}Y^{1}\bigr]},\\
 & \ksd P{R;k_{\mathrm{IMQ}}^{\mathrm{scale}}(x,y)}^{2}=(\sigma_{1}-1)^{2}\underbrace{\EE_{X,Y\sim{\cal N}(0,I)}\left[\frac{X^{1}Y^{1}}{\sqrt{1+(X^{1}-Y^{1})^{2}+\sum_{i>1}(X^{i}-Y^{i})^{2}}}\right]}_{\EE\bigl[k_{\mathrm{IMQ}}^{\mathrm{scale}}(X,Y)X^{1}Y^{1}\bigr]},
\end{align*}
where the expectations are taken with respect to independent standard
Gaussian random variables $X,Y.$ For the KSD to be sensitive to this
deviation in variance, the expectations on the RHS have to be large.
In this regard, the key difference between these expressions is that
the variance $\sigma_{1}^{2}$ appears in the coefficient of $(X^{1}-Y^{1})^{2}.$
When $\sigma_{1}\gg1,$ the non-scaled IMQ kernel $k_{\mathrm{IMQ}}$
pays more attention to the first coordinate than the scaled counterpart
$k_{\mathrm{IMQ}}^{\mathrm{scale}},$ and we can therefore expect
a higher expectation value for $k_{\mathrm{IMQ}}$. On the other hand,
when $\sigma_{1}\ll1,$ the relation flips as $\sigma_{1}$ reduces
the contribution of the first coordinate. Note that this relation
holds more starkly in high dimensions, as the rest of the coordinates
have greater influence on the kernel output. These considerations
show that the ability to choose an effective kernel depends on the
problem (not surprisingly). In particular, it shows that dimension-wise
scaling (or covariance preconditioning) could hurt the performance
in some problems.

Finally, if we instead use the exponentiated quadratic (EQ) kernel
$k_{\mathrm{EQ}}(x,y)=\exp(-\Verts{x-y}_{2}^{2}),$ then we have
\[
\ksd PR^{2}=\left(\sigma_{1}-1\right)^{2}\underbrace{\EE_{X,Y\sim{\cal N}(0,I)}\left[\exp(-\sum_{i>1}(X^{i}-Y^{i})^{2})\exp(-\sigma_{1}^{2}(X^{1}-Y^{1})^{2})X^{1}Y^{1}\right]}_{\EE[k_{\mathrm{EQ}}(X,Y)X^{1}Y{}^{1}]}.
\]
When $\sigma_{1}\gg1,$ the EQ has higher selectivity in the first
coordinate than the IMQ because of its exponential decay; the KSD
with the EQ kernel could be more useful in revealing this perturbation.
However, in practice, we do not know the discrepancy of our models
a priori. At least in terms of local sensitivity such as the example
above, the IMQ kernel could be considered more robust against poor
specification of input scaling than the EQ, as the effect of input
scaling is less significant. 

\subsection{The score formula for Dirichlet process models\label{subsec:dpm-score-formula}}

We provide an explicit formula for the score formula \eqref{eq:score_formula}
for Dirichlet process mixture models, mentioned in \ref{sec:Experiments}.
We first note that the density $p(x|\calD)$ is given by 
\begin{align*}
p(x|\calD) & =\EE_{F}\left[\left.\int\psi(x|z)\dd F(z)\right|\calD\right]\\
 & =\int\int\psi(x|z)\dd\bar{F}_{\calD}(z),
\end{align*}
where $\bar{F}_{\calD}$ is the mean measure of the posterior distribution
of $F$ given $\calD.$ Note that $\bar{F}_{\calD}$ is the mean of
a mixture of Dirichlet processes with the mixing distribution given
by the distribution of the latents of the training data $\{\tilde{z}_{i}\}_{i=1}^{n_{\mathrm{tr}}}$
conditioned on ${\cal D}$ \citep[see][Remark 5.4]{ghosalFundamentalsNonparametricBayesian2017a}.
We can interchange the inside integral and the expectation of $F,$
which results in the second line. This expression immediately gives
\begin{align*}
\score p(x) & =\frac{\int\int\nabla_{x}\psi(x|z)\dd\bar{F}_{\calD}(z)}{p(x|\calD)}\\
 & =\int\int\score{\psi}(x|z,\phi)\frac{\psi(x|z)}{p(x|\calD)}\dd\bar{F}_{\calD}(z).
\end{align*}

We discuss how to evaluate the expectation with MCMC. Our target distribution
is 
\[
\frac{\psi(x|z)}{p(x|\calD)}\bar{F}_{\calD}(\dd z).
\]
Note that this distribution is a mixture of two distributions written
as 

\begin{align*}
\frac{\psi(x|z)}{p(x|\calD)}\bar{F}_{\calD}(\dd z) & =\frac{1}{p(x|\calD)}\left\{ \frac{C_{a}}{n+1}\frac{\psi(x|z)}{C_{a}}\dd a(z)+\frac{nC_{b}}{n+1}\frac{\psi(x|z)}{C_{b}}\underbrace{\EE\left[\left.\frac{1}{n}\sum_{i}\delta_{\tilde{Z}_{i}}(\dd z)\right|\calD\right]}_{b(\dd z)}\right\} \\
 & =\pi_{a}\frac{\psi(x|z)}{C_{a}}\dd a(z)+\pi_{b}\frac{\psi(x|z)}{C_{b}}\dd b(z),
\end{align*}
where $C_{\alpha}=\int\psi(x|z)\dd a(z),$ $C_{b}=\int\psi(x|z)\dd b(z)$,
$\pi_{a}=C_{a}/(C_{a}+nC_{b}),$ and $\pi_{b}=1-\pi_{a}.$ For the
Gaussian DPM model in \ref{subsec:expDPM}, we can sample from the
posterior $\psi(x|z,\phi)/C_{a}\dd a(z),$ and it can be used for
initializing the Markov chain. The distribution in the second term
is not given in closed form as the mean measure $b$ is unknown, but
we can sample from $b$ (and so from $\bar{F}_{{\cal D}})$ by Gibbs
sampling \citep[Theorem 5.3]{ghosalFundamentalsNonparametricBayesian2017a}.
Assuming that we can generate samples from $\bar{F}_{{\cal D}},$
we can use the random walk Metropolis algorithm where the acceptance
probability of the transition from $z$ to $z'$ is given by 
\[
\min\left(1,\frac{\psi(x|z')}{\psi(x|z)}\right)
\]
with the proposal distribution $\bar{F}_{{\cal D}}.$ However, sampling
from $\bar{F}_{D}$ cannot be performed exactly, and therefore we
use Gibbs sampling after sufficient burn-in. Consequently, the use
of Gibbs and Metropolis samplers allows us to sample from $\psi(x|z)/p(x|\calD)\bar{F}_{\calD}$
approximately. 

\subsection{Invariance properties of kernel Stein discrepancy }

\subsubsection{Model invariance \label{subsec:Model-invariance}}

In some applications, models are designed to be invariant to certain
transformations. When comparing a class of models invariant under
a transformation, model comparison should be made so that the ranks
of the models remain unaffected under the transformation of the data.
We show that essentially for rotational transformations, we can make
the KSD invariant by choosing a rotationally invariant kernel. In
the following, for a map $T:\calX\to\calX$ and a distribution $P,$
we denote $T$-push-forward of $P$ by $T_{\#}P;$ i.e., $T_{\#}P$
is defined as the distribution of a random variable $Tx$ with $x\sim P.$ 

\begin{lemma}

Assume that for an orthogonal matrix $O,$ the following conditions
hold: (a) $P$ has a density $p$ such that $p(Ox)=p(x)$ for any
$x\in\mathbb{R}^{D},$ and (b) kernel $k$ satisfies $k(Ox,Oy)=k(x,y)$
for any $x,y\in{\cal \mathbb{R}}^{D}.$ Then, we have $\ksd P{O_{\#}R}=\ksd PR.$ 

\end{lemma}

\begin{proof}

When we push forward the data distribution by $O,$ the KSD becomes 

\begin{align*}
\ksd P{O_{\#}R}^{2} & =\EE_{x,x'\sim R\otimes R}\bigl[h_{p}(Ox,Ox')\bigr].
\end{align*}
The assumption $p(Ox)=p(x)$ implies that 
\begin{align*}
\score p(Ox) & =\frac{1}{p(x)}\Bigl(\partial_{h}p(x+hO^{-1}e_{d})\rvert_{h=0}\Bigr)_{d=1}^{D}\\
 & =(O^{-1})^{\top}\frac{\nabla p(x)}{p(x)}=O\score p(x),
\end{align*}
where $\{e_{1},\dots,e_{D}\}$ denotes the standard basis of $\mathbb{R}^{D}.$
For the kernel derivatives $k_{1}$ and $k_{12},$ we obtain 
\begin{align*}
k_{1}(Ox,Ox') & =Ok_{1}(x,x')\ \text{and}\ k_{12}(Ox,Ox')=k_{12}(x,x').
\end{align*}
Thus, 
\begin{align*}
h_{p}(Ox,Ox') & =\score p(Ox)^{\top}\score p(Ox')k(Ox,Ox')+\score p(Ox)^{\top}k_{1}(Ox',Ox)\\
 & \hphantom{=\ }+k_{1}(Ox,Ox')^{\top}\score p(Ox')+k_{12}(Ox,Ox')\\
 & =\score p(x)^{\top}\score p(x')k(x,x')+\score p(x)^{\top}k_{1}(x',x)\\
 & \hphantom{=\ }+k_{1}(x,x')^{\top}\score p(x')+k_{12}(x,x')=h_{p}(x,x'),
\end{align*}
and therefore $\ksd P{O_{\#}R}=\ksd PR.$ \qed

\end{proof}

An analogous result holds for the KSD for discrete observations.

\begin{lemma}\label{lem:KSD-perm-invariance}

Let $\sigma:\{1,\dots,D\}\to\{1,\dots,D\}$ be a permutation represented
by a permutation matrix $O_{\sigma}.$ Assume that $P$ is invariant
to $O_{\sigma}$; i.e., $(O_{\sigma})_{\#}P=P.$ Assume that kernel
$k$ satisfies $k(O_{\sigma}x,O_{\sigma}y)=k(x,y)$ for any $x,y\in\{0,\dots,L-1\}^{D}.$
Then, $\ksd P{(O_{\sigma})_{\#}P}=\ksd PR.$ 

\end{lemma}

\begin{proof}

The proof proceeds as in the previous lemma. Note that taking the
cyclic forward difference with respect to the $i$-th coordinate gives
\begin{align*}
\Delta_{i}p(O_{\sigma}x) & =p(x^{\sigma(1)},\dots,\tilde{x}^{i},\dots x^{\sigma(D)})-p(x^{\sigma(1)},\dots,x^{\sigma(i)},\dots x^{\sigma(D)})\\
 & =p(x^{1},\dots,\tilde{x}^{\sigma^{-1}(i)},\dots x^{D})-p(x)\\
 & =\Delta_{\sigma^{-1}(i)}p(x),
\end{align*}
where $\tilde{x}=x+1\mod L.$ Thereby, 
\begin{align*}
\score p(O_{\sigma}x) & =O_{\sigma}^{-1}\score p(x)=O_{\sigma}^{\top}\score p(x).
\end{align*}
Similarly, for the kernel \emph{derivatives} $k_{1}$ and $k_{12},$
we have 
\begin{align*}
k_{1}(O_{\sigma}x,O_{\sigma}x') & =O_{\sigma}^{^{\top}}k_{1}(x,x')\\
k_{12}(O_{\sigma}x,O_{\sigma}x') & =k_{12}(x,x').
\end{align*}
Thus, 
\begin{align*}
h_{p}(O_{\sigma}x,O_{\sigma}x') & =\score p(O_{\sigma}x)^{\top}\score p(O_{\sigma}x')k(O_{\sigma}x,O_{\sigma}x')+\score p(O_{\sigma}x)^{\top}k_{1}(O_{\sigma}x',O_{\sigma}x)\\
 & \hphantom{=\ }+k_{1}(O_{\sigma}x,O_{\sigma}x')^{\top}\score p(O_{\sigma}x')+k_{12}(O_{\sigma}x,O_{\sigma}x')\\
 & =\score p(x)^{\top}\score p(x')k(x,x')+\score p(x)^{\top}k_{1}(x',x)\\
 & \hphantom{=\ }+k_{1}(x,x')^{\top}\score p(x')+k_{12}(x,x')=h_{p}(x,x'),
\end{align*}
and therefore $\ksd P{(O_{\sigma})_{\#}R}=\ksd PR.$ \qed

\end{proof}

\subsubsection{Coordinate-choice independence \label{subsec:Coordinate-choice-independence}}

In general, the KSD is not invariant to a change of coordinates, and
the KSD may be affected if we transform both the model and the data
distribution. Precisely, for some one-to-one map $T:\calX\to\calX,$
we might have $\ksd{T_{\#}P}{T_{\#}R}\neq\ksd PR.$ The following
result is essentially the same as the previous lemmas except that
here we do not have the invariance assumptions for the model; it shows
that the KSD can be made invariant at least under rotation and translation.

\begin{lemma}

Let $T$ be an affine transform such that $Tx=Ox+b$ where $O$ is
an orthogonal matrix and $b$ is a vector. Let $P,R$ be probability
distributions. Let $k=k_{R}$ be a \emph{data-dependent} kernel $k_{R}$
such that $k_{T_{\#}R}(x,x')=k_{R}(T^{-1}x,T^{-1}x')$ for any $x,x'\in\mathbb{R}^{D}.$
Then the KSD between $P$ and $R$ is invariant under $T;$ that is,
$\ksd{T_{\#}P}{T_{\#}R}=\ksd PR.$

\end{lemma}

\begin{proof}

Let us denote the density of $T_{\#}P$ by $p_{T}(x)=p(T^{-1}x).$
Then, its score function satisfies
\begin{align*}
\score{p_{T}}(x) & =O\score p(T^{-1}x).
\end{align*}
Similarly, for the kernel derivatives $k_{1}$ and $k_{12},$ we have
\begin{align*}
k_{T_{\#}R,1}(x,y) & =\nabla_{\tilde{x}}k_{R}(T^{-1}\tilde{x},T^{-1}y)\vert_{\tilde{x}=x}\\
 & =Ok_{R,1}(T^{-1}x,T^{-1}y),\ \text{and}\\
k_{T_{\#}R,12}(x,y) & =\nabla_{\tilde{x}}^{\top}\nabla_{\tilde{y}}k_{R}(T^{-1}\tilde{x},T^{-1}y)\vert_{\tilde{x}=x,\tilde{y}=y}\\
 & =(\nabla_{\tilde{x}}^{\top}O^{-1}O^{\top}\nabla_{\tilde{y}})k_{R}(\tilde{x},\tilde{y})\vert_{\tilde{x}=x,\tilde{y}=y}=k_{R,12}(x,y).
\end{align*}
These relations imply that the Stein kernel satisfies 
\begin{align*}
h_{p_{T}}(Tx,Tx') & =\score{p_{T}}(Tx)^{\top}\score{p_{T}}(Tx')k_{T_{\#}R}(Tx,Tx')+\score{p_{T}}(x)^{\top}k_{T_{\#}R,1}(Tx',Tx)\\
 & \hphantom{=\ }+k_{T_{\#}R,1}(Tx,Tx')^{\top}\score{p_{T}}(x')+k_{T_{\#}R,12}(Tx,Tx')\\
 & =\score p(x)O^{^{\top}}O\score p(x')k_{R}(x,x')+\score p(x)^{\top}O^{\top}Ok_{R,1}(x',x)\\
 & \hphantom{=\ }+k_{R,1}(x,x')^{\top}O^{-1}O\score p(x')+(\nabla_{\tilde{x}}^{\top}O^{-1}O\nabla_{\tilde{y}})k_{R,12}(T^{-1}\tilde{x},T^{-1}\tilde{y})\vert_{\tilde{x}=Tx,\tilde{y}=Tx'}\\
 & =\score p(x)^{\top}\score p(x')k_{R}(x,x')+\score p(x)^{\top}k_{R,1}(x',x)\\
 & \hphantom{=\ }+k_{R,1}(x,x')^{\top}\score p(x')+k_{R,12}(x,x')=h_{p}(x,x').
\end{align*}
Thus, 
\[
\ksd{T_{\#}P}{T_{\#}R}^{2}=\EE_{x,x'\sim R\otimes R}\Bigl[h_{p_{T}}(Tx,Tx')\Bigr]=\EE_{x,x'\sim R\otimes R}\Bigl[h_{p}(x,x')\Bigr]=\ksd PR^{2}.\qed
\]

\end{proof}

An example of the data-dependent kernel is a covariance-preconditioned
kernel 
\[
k_{R}^{\mathrm{precond}}(x,y)=\phi\bigl((x-y)^{\top}\hat{\Sigma}_{R}^{-1}(x-y)\bigr)
\]
where $\hat{\Sigma}_{R}$ is the sample covariance matrix of $R$
and $\phi$ is some positive-definite function. Another example is
the median-scaled kernel 
\[
k_{R}^{\mathrm{med}}(x,y)=\phi\bigl(\Verts{x-y}_{2}^{2}/\sigma_{R,\mathrm{med}}^{2}\bigr),
\]
where $\sigma_{R,\mathrm{med}}$ is the sample median: $\mathrm{median}\{\Verts{x_{i}-x_{j}}_{2}|1<i<j<n\}.$
In fact, radial basis kernels with data-independent scaling also satisfy
the required condition since $k_{T_{\#}R}(x,y)=k_{R}(x,y)=k_{R}(T^{-1}x,T^{-1}y).$

\subsection{\citet{Arvesen1969Jackknifing}'s formula for the jackknife variance
estimator \label{subsec:Jackknife-formula}}

The formula in \citep[Eq. 25]{Arvesen1969Jackknifing} has minor errors.
For completeness, we provide a proof for the decomposition \eqref{eq:jackknife-Ustat-decomp}.

\begin{lemma}

For an i.i.d. sample $\calD_{n}=\{y_{i}\}_{i=1}^{n}$ from some distribution,
define a U-statistic with symmetric kernel $f:{\cal Y}^{s}\to\mathbb{R},$
\[
U_{n}={n \choose s}^{-1}\sum_{C_{n,s}}f(y_{i_{1}},\dots,y_{i_{s}}),
\]
where $C_{n.s}$ denotes the set of $s$ combinations of integers
chosen from $\{1,\dots,n\}$ with $n\geq s\geq1.$ Then, we have
\begin{align*}
v_{n}^{J} & \coloneqq(n-1)\sum_{i=1}^{n}(U_{n,-i}-U_{n})^{2}\\
 & =\sum_{c=0}^{s}a_{n,c}\hat{U}_{c},
\end{align*}
with $U_{n,-i}$ the U-statistic computed with $\calD_{n}\setminus\{y_{i}\},$
\[
a_{n,c}=\frac{n-1}{n}{n-1 \choose s}^{-2}\{n\mathbb{I}(c>0)-s^{2}\}{n \choose c}{n-c \choose s-c}{n-s \choose s-c},
\]
and $\hat{U}_{c}$ is a U-statistic given in Eq. \eqref{eq:jackknife-Uc}
in the proof. 

\end{lemma}

\begin{proof}

For a combination in $C_{n,s},$ we fix a order of integers (say,
sorted in ascending order) and evaluate $f.$ We may assume that the
statistic is centered so that $\EE[U_{n}]=0.$ 

The jackknife estimator is expressed as 
\begin{equation}
\begin{aligned}v_{n}^{J} & =(n-1)\sum_{i=1}^{n}(U_{n,-i}-U_{n})^{2}\\
 & =(n-1)\Bigl\{\sum_{i=1}^{n}(U_{n,-i})^{2}-n(U_{n})^{2}\Bigr\}\\
 & =(n-1){n-1 \choose s}^{-2}\left[\sum_{i=1}^{n}\sum_{\alpha\in C_{n,s-1}^{i}}f(y_{\alpha_{1}},\dots,f_{\alpha_{s}})\sum_{\beta\in C_{n,s-1}^{i}}f(y_{\beta_{1}},\dots,f_{\beta_{s}})\right.\\
 & \hphantom{=(n-1){n-1 \choose m}^{-2}\ }\left.\ -\frac{(n-s)^{2}}{n}\sum_{\alpha\in C_{n,s}}f(y_{\alpha_{1}},\dots,f_{\alpha_{s}})\sum_{\beta\in C_{n,s}}f(y_{\beta_{1}},\dots,f_{\beta_{s}})\right],
\end{aligned}
\label{eq:jackknife-expansion-mess}
\end{equation}
where $C_{n,s-1}^{i}$ denotes the set of all $s$ combinations of
integers chosen from $\{1,\dots,i-1,i+1,\dots,n\}$, and $C_{n,s}$
denotes the set of all $s$ combinations of integers chosen from $\{1,\dots,n\}.$
The first sum can be expressed as
\begin{align*}
 & \sum_{i=1}^{n}\sum_{\alpha\in C_{n,s-1}^{i}}f(y_{\alpha_{1}},\dots,f_{\alpha_{s}})\sum_{\beta\in C_{n,s-1}^{i}}f(y_{\beta_{1}},\dots,f_{\beta_{s}})\\
 & =\sum_{i=1}^{n}\left\{ \sum_{\alpha\in C_{n,s}}f(y_{\alpha_{1}}\dots,f_{\alpha_{s}})-\sum_{\alpha\in C_{n,s-1}^{i}}f(y_{i},y_{\alpha_{1}}\dots,f_{\alpha_{s-1}})\right\} \left\{ \sum_{\beta\in C_{n,s}}f(y_{\beta_{1}}\dots,f_{\beta_{s}})-\sum_{\beta\in C_{n,s-1}^{i}}f(y_{i},y_{\beta_{1}}\dots,f_{\beta_{s-1}})\right\} .
\end{align*}
Note that we have 
\begin{align*}
\sum_{i=1}^{n}\sum_{\alpha\in C_{n,s-1}^{i}}f(y_{i},y_{\alpha_{1}}\dots,f_{\alpha_{s-1}}) & =\frac{1}{(s-1)!}\sum_{i=1}^{n}\sum_{\alpha\in P_{n,s-1}^{i}}f(y_{i},y_{\alpha_{1}}\dots,f_{\alpha_{s-1}})\\
 & =\frac{1}{(s-1)!}\sum_{\alpha\in P_{n,s}}f(y_{\alpha_{1}}\dots,f_{\alpha_{s}})\\
 & =\frac{s!}{(s-1)!}\sum_{\alpha\in C_{n,s}}f(y_{\alpha_{1}}\dots,f_{\alpha_{s}})\\
 & =s\sum_{\alpha\in C_{n,s}}f(y_{\alpha_{1}}\dots,f_{\alpha_{s}}),
\end{align*}
where $P_{n,s-1}^{i}$ is the set of all ordered $(s-1)$-tuples of
integers chosen from $\{1,\dots i-1,i+1,\dots,n\},$ and $P_{n,s}$
denotes the set of all ordered $s$-tuple of integers chosen from
$\{1,\dots,n\}.$ The first and third lines are due to the symmetry
of $f.$ The second lines holds as the indices on the RHS on the first
line runs all over $P_{n,s}.$ Moreover, we have
\begin{align*}
 & \sum_{i=1}^{n}\sum_{\alpha\in C_{n,s-1}^{i}}f(y_{i},y_{\alpha_{1}}\dots,f_{\alpha_{s-1}})\sum_{\beta\in C_{n,s-1}^{i}}f(y_{i},y_{\beta_{1}}\dots,f_{\beta_{s-1}})\\
 & =\sum_{\substack{\alpha,\beta\in C_{n,s}\\
\lvert\alpha\cap\beta\rvert\geq1
}
}f(y_{\alpha_{1}},\dots,f_{\alpha_{s}})f(y_{\beta_{1}}\dots,f_{\beta_{s}}),
\end{align*}
where $\lvert\alpha\cap\beta\rvert$ expresses the number of common
elements between two sets of integers $\alpha,\beta.$ Thus, we have
\begin{align*}
 & \sum_{i=1}^{n}\sum_{\alpha\in C_{n,s-1}^{i}}f(y_{\alpha_{1}^{i}},\dots,f_{\alpha_{s}^{i}})\sum_{\beta\in C_{n,s-1}^{i}}f(y_{\beta_{1}^{i}},\dots,f_{\beta_{s}^{i}})\\
 & =n\sum_{\alpha\in C_{n,s}}f(y_{\alpha_{1}}\dots,f_{\alpha_{s}})\sum_{\beta\in C_{n,s}}f(y_{\beta_{1}}\dots,f_{\beta_{s}})-2s\sum_{\alpha\in C_{n,s}}f(y_{\alpha_{1}}\dots,f_{\alpha_{s}})\\
 & \hphantom{=}+\sum_{\substack{\alpha,\beta\in C_{n,s}\\
\lvert\alpha\cap\beta\rvert\geq1
}
}f(y_{\alpha_{1}},\dots,f_{\alpha_{s}})f(y_{\beta_{1}}\dots,f_{\beta_{s}}).
\end{align*}

Now, the expression \eqref{eq:jackknife-expansion-mess} can be summarized
as 
\begin{align}
\eqref{eq:jackknife-expansion-mess}= & \frac{n-1}{n}{n-1 \choose s}^{-2}\left[\sum_{c=0}^{s}\{n^{2}-2sn+n\mathbb{I}(c>0)\}\sum_{\substack{\alpha,\beta\in C_{n,s}\\
\lvert\alpha\cap\beta\rvert=c
}
}f(y_{\alpha_{1}},\dots,f_{\alpha_{s}})f(y_{\beta_{1}},\dots,f_{\beta_{s}})\right.\nonumber \\
 & \hphantom{\frac{n-1}{n}{n-1 \choose m}^{-2}\ }\left.\ -(n-s)^{2}\sum_{c=0}^{s}\sum_{\substack{\alpha,\beta\in C_{n,s}\\
\lvert\alpha\cap\beta\rvert=c
}
}f(y_{\alpha_{1}},\dots,f_{\alpha_{s}})f(y_{\beta_{1}},\dots,f_{\beta_{s}})\right]\nonumber \\
 & =\frac{n-1}{n}{n-1 \choose s}^{-2}\sum_{c=0}^{s}\{n\mathbb{I}(c>0)-s^{2}\}\sum_{\substack{\alpha,\beta\in C_{n,s}\\
\lvert\alpha\cap\beta\rvert=c
}
}f(y_{\alpha_{1}},\dots,f_{\alpha_{s}})f(y_{\beta_{1}},\dots,f_{\beta_{s}}).\label{eq:jackknife-summary-sum}
\end{align}

Finally, we show that each term of the RHS \eqref{eq:jackknife-summary-sum}
is written by a U-statistic $\hat{U}_{c}$ with kernel 
\begin{align*}
 & f_{\mathrm{sym}}(y_{\alpha_{1}},\dots,y_{\alpha_{c}},y_{\beta_{1}},\dots,y_{\beta_{s-c}},y_{\gamma_{1}},\dots,y_{\gamma_{s-c}})\\
 & \coloneqq\frac{1}{(2s-c)!}\sum_{\sigma\in\Sigma(\alpha,\beta,\gamma)}f(y_{\sigma(\alpha_{1})},\dots,y_{\sigma(\alpha_{c})},y_{\sigma(\beta_{1})},\dots,y_{\sigma(\beta_{s-c})})f(y_{\sigma(\alpha_{1})},\dots,y_{\sigma(\alpha_{c})},y_{\sigma(\gamma_{1})},\dots,y_{\sigma(\gamma_{s-c})}),
\end{align*}
where $\Sigma(\alpha,\beta,\gamma)$ is the set of all permutations
of given integers $(\alpha_{1},\dots,\alpha_{c},\beta_{1},\dots,\beta_{s-c},\gamma_{1},\dots,\gamma_{s-c}).$
The reasoning is as follows:

\begin{align}
 & \hat{U}_{c}\nonumber \\
 & \coloneqq{n \choose 2s-c}^{-1}\sum_{C_{n,2s-c}}f_{\mathrm{sym}}(y_{\alpha_{1}},\dots,y_{\alpha_{c}},y_{\beta_{1}},\dots,y_{\beta_{s-c}},y_{\gamma_{1}},\dots,y_{\gamma_{s-c}})\nonumber \\
 & =\frac{(n-2s+c)!}{n!}\sum_{P_{n,2s-c}}f_{\mathrm{sym}}(y_{\alpha_{1}},\dots,y_{\alpha_{c}},y_{\beta_{1}},\dots,y_{\beta_{s-c}},y_{\gamma_{1}},\dots,y_{\gamma_{s-c}})\nonumber \\
 & =\frac{(n-2s+c)!}{n!}\frac{1}{(2s-c)!}\nonumber \\
 & \hphantom{=}\cdot\sum_{P_{n,2s-c}}\sum_{\sigma\in\Sigma(\alpha,\beta,\gamma)}f(y_{\sigma(\alpha_{1})},\dots,y_{\sigma(\alpha_{c})},y_{\sigma(\beta_{1})},\dots,y_{\sigma(\beta_{s-c})})f(y_{\sigma(\alpha_{1}),}\dots,y_{\sigma(\alpha_{c})},y_{\sigma(\gamma_{1})},\dots,y_{\sigma(\gamma_{s-c})})\nonumber \\
 & =\frac{(n-2s+c)!}{n!}\frac{1}{(2s-c)!}\nonumber \\
 & \hphantom{=}\cdot\sum_{\sigma}\sum_{\sigma(\alpha)\in P_{n,c}}\sum_{\substack{\sigma(\beta)\in P_{n,s-c}\\
\sigma(\alpha)\cap\sigma(\beta)=\emptyset
}
}\sum_{\substack{\sigma(\gamma)\in P_{n,s-c}\\
\sigma(\gamma)\cap\{\sigma(\alpha)\cup\sigma(\beta)\}=\emptyset
}
}\big(f(y_{\sigma(\alpha_{1})},\dots,y_{\sigma(\alpha_{c})},y_{\sigma(\beta_{1})},\dots,y_{\sigma(\beta_{s-c})})\nonumber \\
 & \hphantom{=\cdot\sum_{\sigma}\sum_{\sigma(\alpha)\in P_{n,c}}\sum_{\substack{\sigma(\beta)\in P_{n,s-c}\\
\sigma(\alpha)\cap\sigma(\beta)=\emptyset
}
}\sum_{\substack{\sigma(\gamma)\in P_{n,s-c}\\
\sigma(\gamma)\cap\{\sigma(\alpha)\cup\sigma(\beta)\}=\emptyset
}
}\quad}\cdot f(y_{\sigma(\alpha_{1}),}\dots,y_{\sigma(\alpha_{c})},y_{\sigma(\gamma_{1})},\dots,y_{\sigma(\gamma_{s-c})})\bigr)\label{eq:jackknife-intermediate-Uc}
\end{align}
The second line follows from the permutation invariance of $f_{\mathrm{sym}}.$
The third line is obtained by inserting the definition. The fourth
line is a result of exchanging the sums. Continuing to manipulate
the expression, we obtain 
\begin{align}
\begin{aligned}\hat{U}_{c} & =\frac{(n-2s+c)!}{n!}(s-c)!(s-c)!c!\\
 & \hphantom{=}\cdot\sum_{\alpha\in C_{n,c}}\sum_{\substack{\beta\in C_{n,s-c}\\
\alpha\cap\beta=\emptyset
}
}\sum_{\substack{\gamma\in C_{n,s-c}\\
\gamma\cap\{\alpha\cup\beta\}=\emptyset
}
}f(y_{\alpha_{1}},\dots,y_{\alpha_{c}},y_{\beta_{1}},\dots,y_{\beta_{s-c}})f(y_{\alpha_{1},}\dots,y_{\alpha_{c}},y_{\gamma_{1}},\dots,y_{\gamma_{s-c}}).\\
 & =\frac{(n-2s+c)!(s-c)!}{(n-s)!}\frac{(n-s)!(s-c)!}{(n-c)!}\frac{c!(n-c)!}{n!}\\
 & \hphantom{=}\cdot\sum_{\alpha\in C_{n,c}}\sum_{\substack{\beta\in C_{n,s-c}\\
\alpha\cap\beta=\emptyset
}
}\sum_{\substack{\gamma\in C_{n,s-c}\\
\gamma\cap\{\alpha\cup\beta\}=\emptyset
}
}f(y_{\alpha_{1}},\dots,y_{\alpha_{c}},y_{\beta_{1}},\dots,y_{\beta_{s-c}})f(y_{\alpha_{1},}\dots,y_{\alpha_{c}},y_{\gamma_{1}},\dots,y_{\gamma_{s-c}})\\
 & ={n-s \choose s-c}^{-1}{n-c \choose s-c}^{-1}{n \choose c}^{-1}\\
 & \hphantom{=}\cdot\sum_{\alpha\in C_{n,c}}\sum_{\substack{\beta\in C_{n,s-c}\\
\alpha\cap\beta=\emptyset
}
}\sum_{\substack{\gamma\in C_{n,s-c}\\
\gamma\cap\{\alpha\cup\beta\}=\emptyset
}
}f(y_{\alpha_{1}},\dots,y_{\alpha_{c}},y_{\beta_{1}},\dots,y_{\beta_{s-c}})f(y_{\alpha_{1},}\dots,y_{\alpha_{c}},y_{\gamma_{1}},\dots,y_{\gamma_{-c}}).
\end{aligned}
\label{eq:jackknife-Uc}
\end{align}
The first line holds because the inner sum \eqref{eq:jackknife-intermediate-Uc}
in $\Sigma_{\sigma}$ has the same value for each $\sigma$ and because
of the permutation invariance of $f.$ Note that the sum in the final
expression is 
\[
\sum_{\substack{\alpha,\beta\in C_{n,s}\\
\lvert\alpha\cap\beta\rvert=c
}
}f(y_{\alpha_{1}},\dots,f_{\alpha_{s}})f(y_{\beta_{1}},\dots,f_{\beta_{s}}).
\]
Therefore, 
\begin{align*}
v_{n}^{J} & =(n-1)\sum_{i=1}^{n}(U_{n,-i}-U_{n})^{2}\\
 & =\sum_{c=0}^{s}a_{n,c}\hat{U}_{c},
\end{align*}
with 
\[
a_{n,c}=\frac{n-1}{n}{n-1 \choose s}^{-2}\{n\mathbb{I}(c>0)-s^{2}\}{n \choose c}{n-c \choose s-c}{n-s \choose s-c}.\qed
\]

\end{proof}

\newpage{}

\section{Additional experiments }

\subsection{PPCA: type-I errors and test power \label{subsec:PPCA-additional}}

We provide results supplementary to the results in Section \ref{subsec:lda-prior-ptb}. 

\paragraph*{Problem 1 (null)}

Table \ref{tab:ppca-null-complete} summarizes the result for the
experiment with ($\delta_{P},\delta_{Q})=(1,1+10^{-5})$ in \eqref{tab:ppca-null-complete},
where all the tests use the IMQ kernel with covariance preconditioning.
The result for $\alpha=0.01$ is omitted as none of the examined tests
rejected the hypotheses. 

\begin{table}[H]
\caption{Type-I errors the MMD test of \citet{BouBelBlaAntGre2016}, the proposed
LKSD test, and the KSD test with the covariance preconditioner in
PPCA Problem 1. Rejection rates are computed on 300 trials for significance
level $\alpha=0.05.$ }

\begin{centering}
\label{tab:ppca-null-complete}
\par\end{centering}
\centering{}%
\begin{tabular}{rrrr}
\hline 
\headcell  \theadmd{Sample size $n$} & \multicolumn{3}{c}{\headcell  \theadmd{Rejection rates}}\tabularnewline
\hline 
 & MMD & KSD & LKSD\tabularnewline
100 & 0.000 & 0.017 & 0.010\tabularnewline
200 & 0.013 & 0.000 & 0.003\tabularnewline
300 & 0.020 & 0.000 & 0.007\tabularnewline
400 & 0.017 & 0.000 & 0.007\tabularnewline
500 & 0.013 & 0.000 & 0.007\tabularnewline
\hline 
\end{tabular}
\end{table}

\paragraph*{Problem 2 (alternative)}

\begin{figure}[H]
\subfloat[\label{fig:ppca_h1_p2_q1_Gauss-alpha001}(a): PPCA $\delta_{P}=2$,
$\delta_{Q}=1.$ EQ kernel with median scaling.]{\begin{centering}
\includegraphics[scale=0.28]{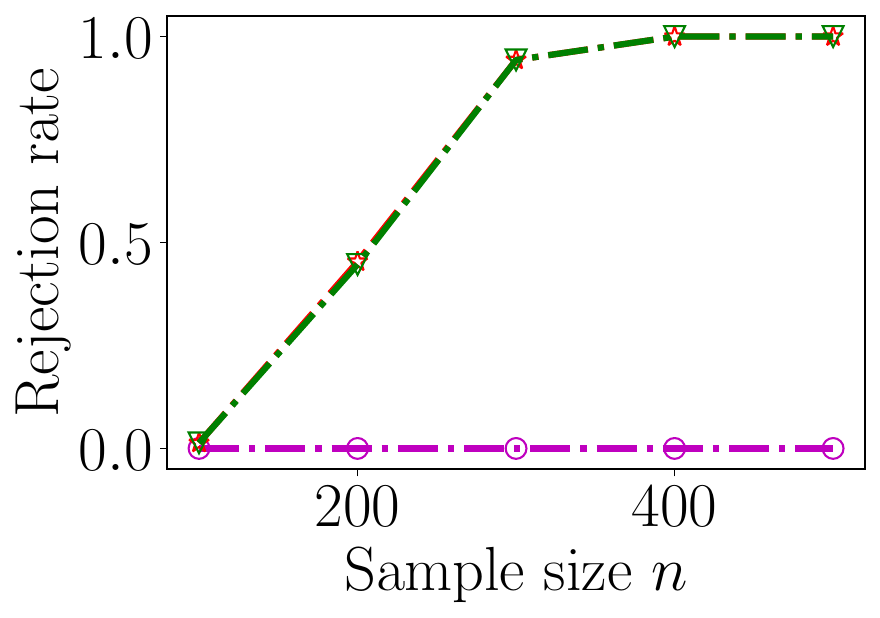}
\par\end{centering}
\centering{}}\hfill{}\subfloat[\label{fig:ppca_h1_p2_q1_IMQ-alpha001}PPCA $\delta_{P}=2$, $\delta_{Q}=1.$
IMQ kernel with median scaling.]{\begin{centering}
\includegraphics[scale=0.28]{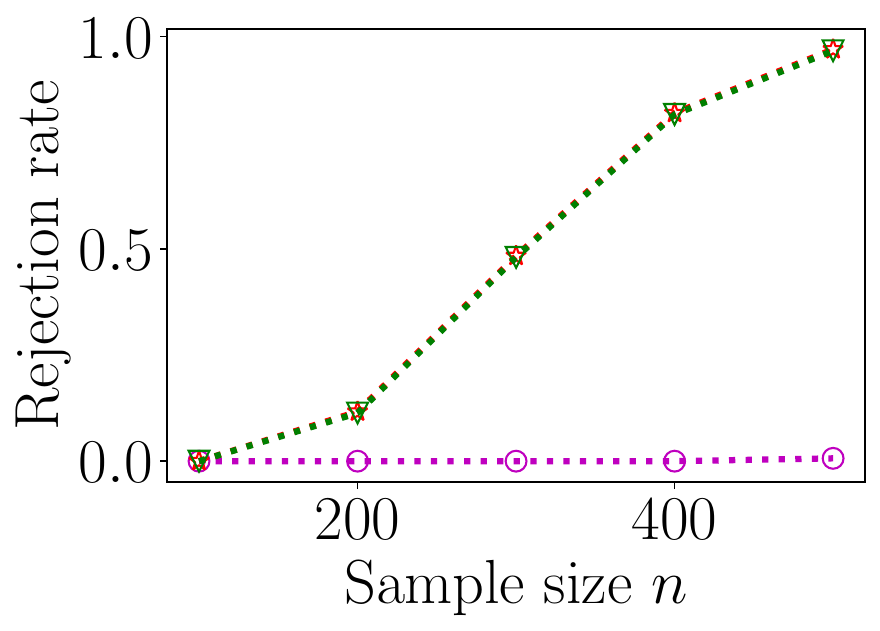}
\par\end{centering}
}\hfill{}\subfloat[\label{fig:ppca_h1_p2_q1_IMQ-cov-alpha001}PPCA $\delta_{P}=2$, $\delta_{Q}=1.$
IMQ kernel with covariance preconditioning.]{\begin{centering}
\includegraphics[scale=0.28]{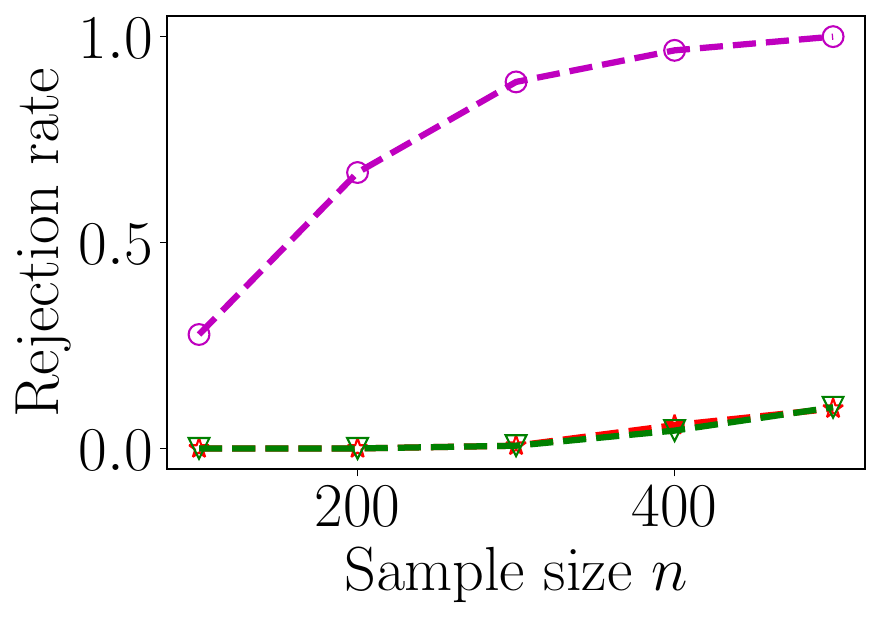}
\par\end{centering}
}\caption{Power curves of the MMD test of \citet{BouBelBlaAntGre2016}, the
proposed LKSD test, and the KSD test with the exact score function
in PPCA Problem 2. The perturbation parameters are set as $(\delta_{P},\delta_{Q}=2,1).$
Each result is computed on 300 trials. The significance level $\alpha=0.01.$
Markers: \textifsymbol[ifgeo]{51} (the LKSD test); \FiveStarOpen{}
(the KSD test); \textbigcircle{} (the relative MMD test).}
\label{fig:ppca_h1_p2_q1-alpha001}
\end{figure}

We present the result of the same experiment with $\alpha=0.01$ (Figure
\ref{fig:ppca_h1_p2_q1-alpha001}) to show power decay due to the
conservatism. 

\begin{figure}[H]
\subfloat[\label{fig:ppca_h1_p3_q1_Gauss}(a): PPCA $\delta_{P}=3$, $\delta_{Q}=1.$
EQ kernel with median scaling.]{\begin{centering}
\includegraphics[scale=0.28]{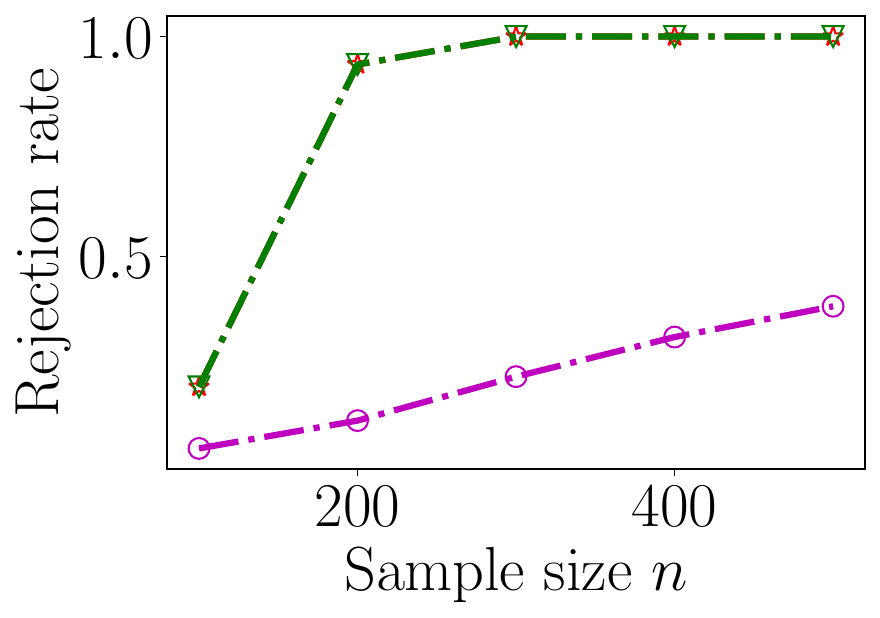}
\par\end{centering}
\centering{}}\hfill{}\subfloat[\label{fig:ppca_h1_p3_q1_IMQ}PPCA $\delta_{P}=3$, $\delta_{Q}=1.$
IMQ kernel with median scaling.]{\begin{centering}
\includegraphics[scale=0.28]{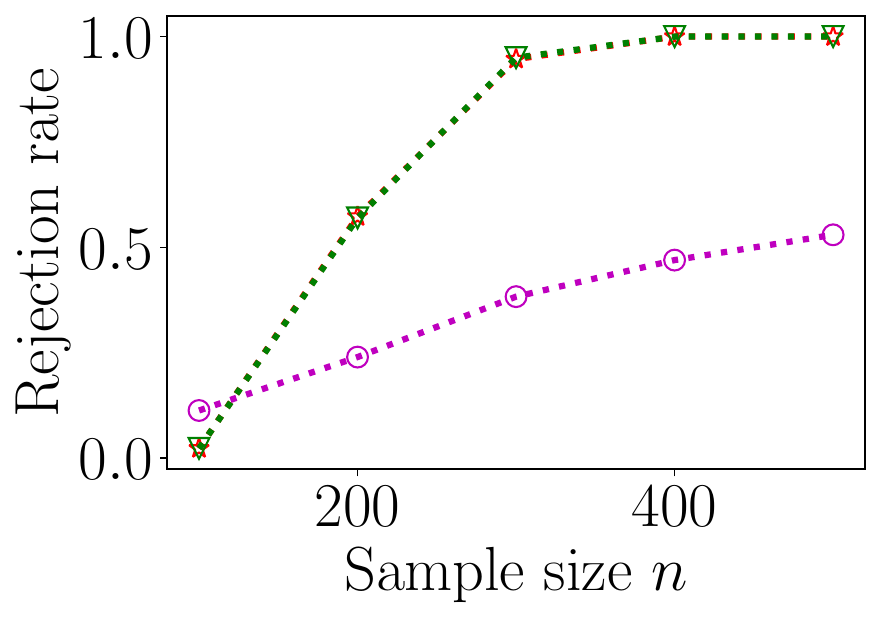}
\par\end{centering}
}\hfill{}\subfloat[\label{fig:ppca_h1_p3_q1_IMQ-cov}PPCA $\delta_{P}=3$, $\delta_{Q}=1.$
IMQ kernel with covariance preconditioning.]{\begin{centering}
\includegraphics[scale=0.28]{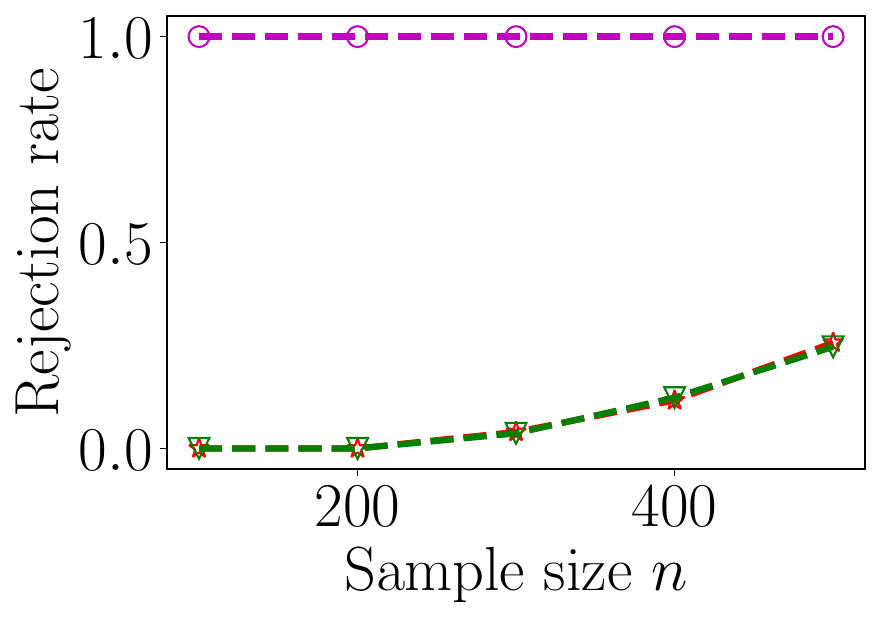}
\par\end{centering}
}\caption{Power curves of the MMD test of \citet{BouBelBlaAntGre2016}, the
proposed LKSD test, and the KSD test with the exact score function
in PPCA Problem 2. The perturbation parameters are set as $(\delta_{P},\delta_{Q}=3,1).$
Each result is computed on 300 trials. The significance level $\alpha=0.05.$
Markers: \textifsymbol[ifgeo]{51} (the LKSD test); \FiveStarOpen{}
(the KSD test); \textbigcircle{} (the relative MMD test).}
\label{fig:ppca_h1_p3_q1}
\end{figure}

Next, we provide the result for the same power experiment with a different
choice of perturbation parameters $(\delta_{P},\delta_{Q})=(3,1).$
Figure \ref{fig:ppca_h1_p3_q1} shows the power curves of the tests.
The MMD test with the covariance pre-conditioner achieves power $1$
at $n=100.$ 

\begin{figure}[H]
\centering{}\subfloat[\label{fig:ppca_h1_p3_q1_Gauss-1}(a): PPCA $\delta_{P}=3$, $\delta_{Q}=1.$
EQ kernel with median scaling.]{\begin{centering}
\includegraphics[scale=0.28]{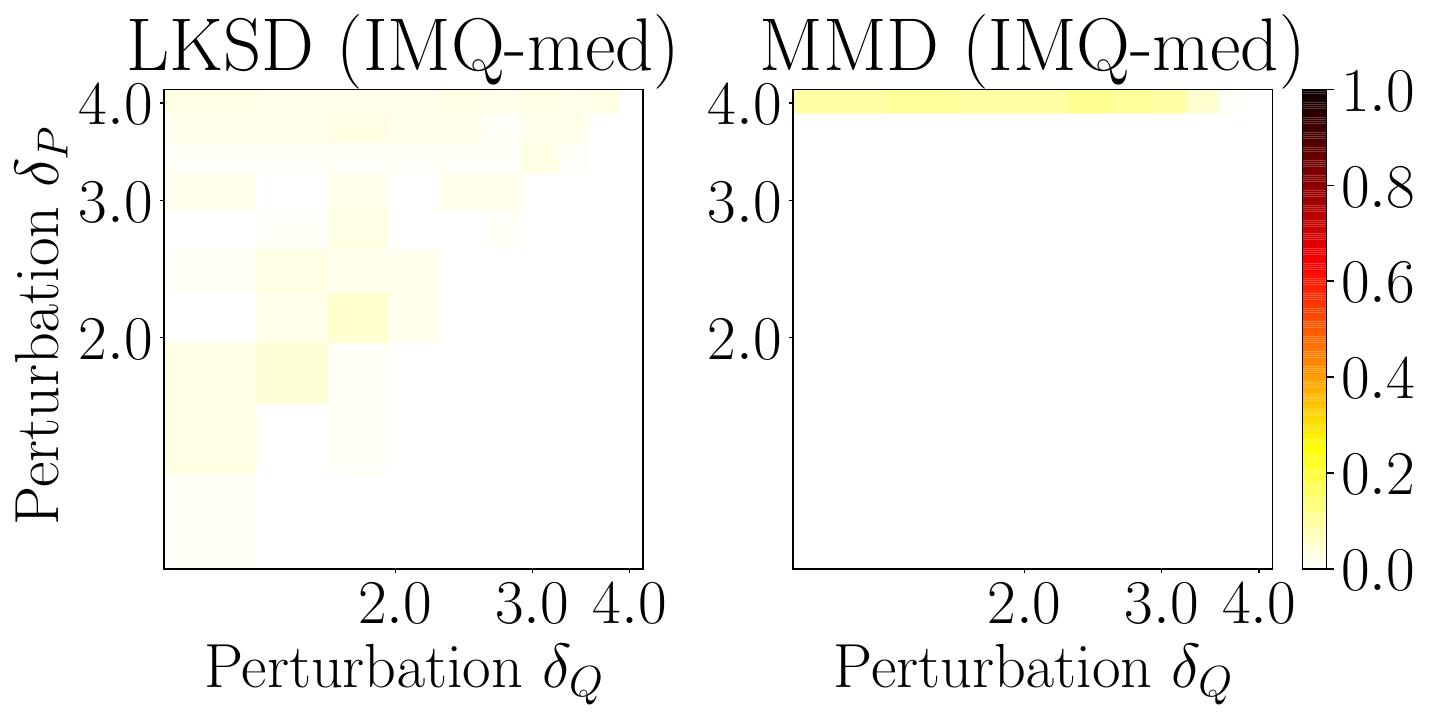}
\par\end{centering}
\centering{}}\hfill{}\subfloat[\label{fig:ppca_h1_p3_q1_IMQ-1}PPCA $\delta_{P}=3$, $\delta_{Q}=1.$
IMQ kernel with median scaling.]{\begin{centering}
\includegraphics[scale=0.28]{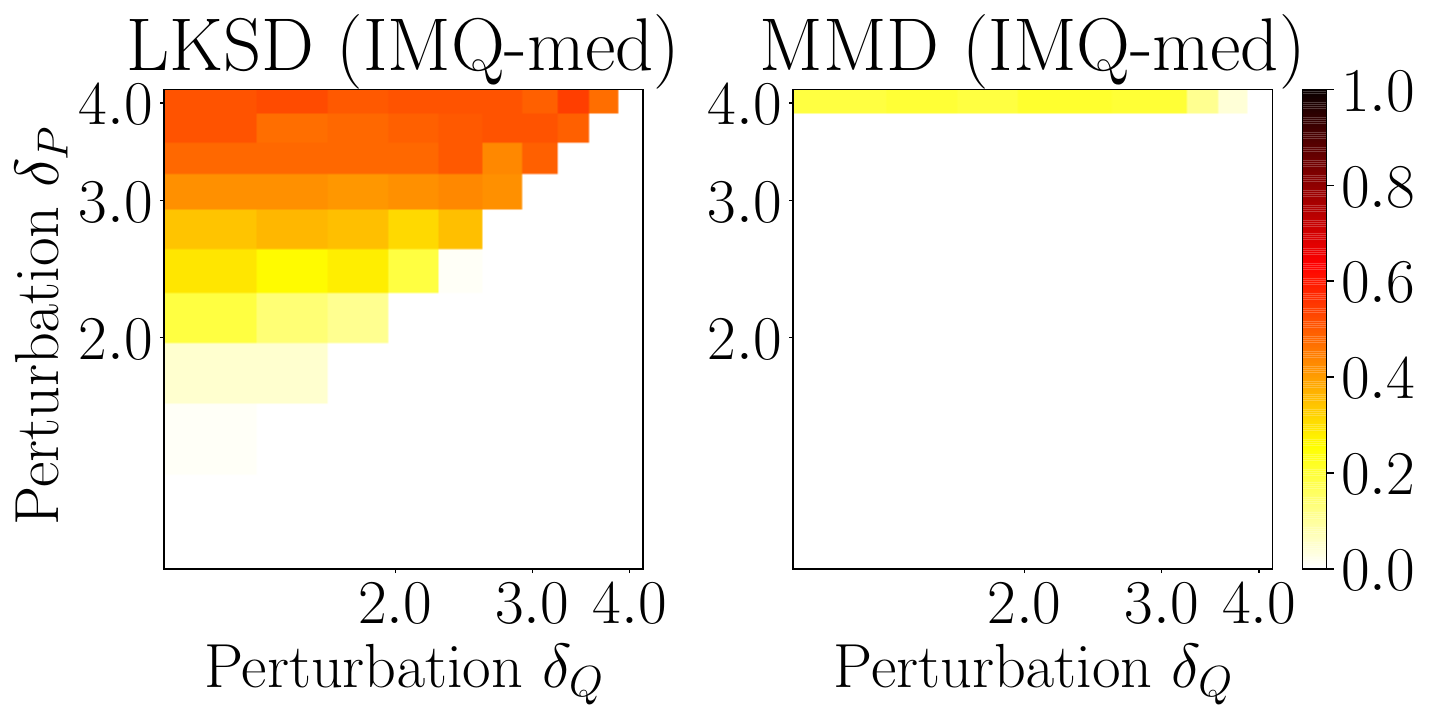}
\par\end{centering}
}\\
\subfloat[$n=300$]{\begin{centering}
\includegraphics[scale=0.28]{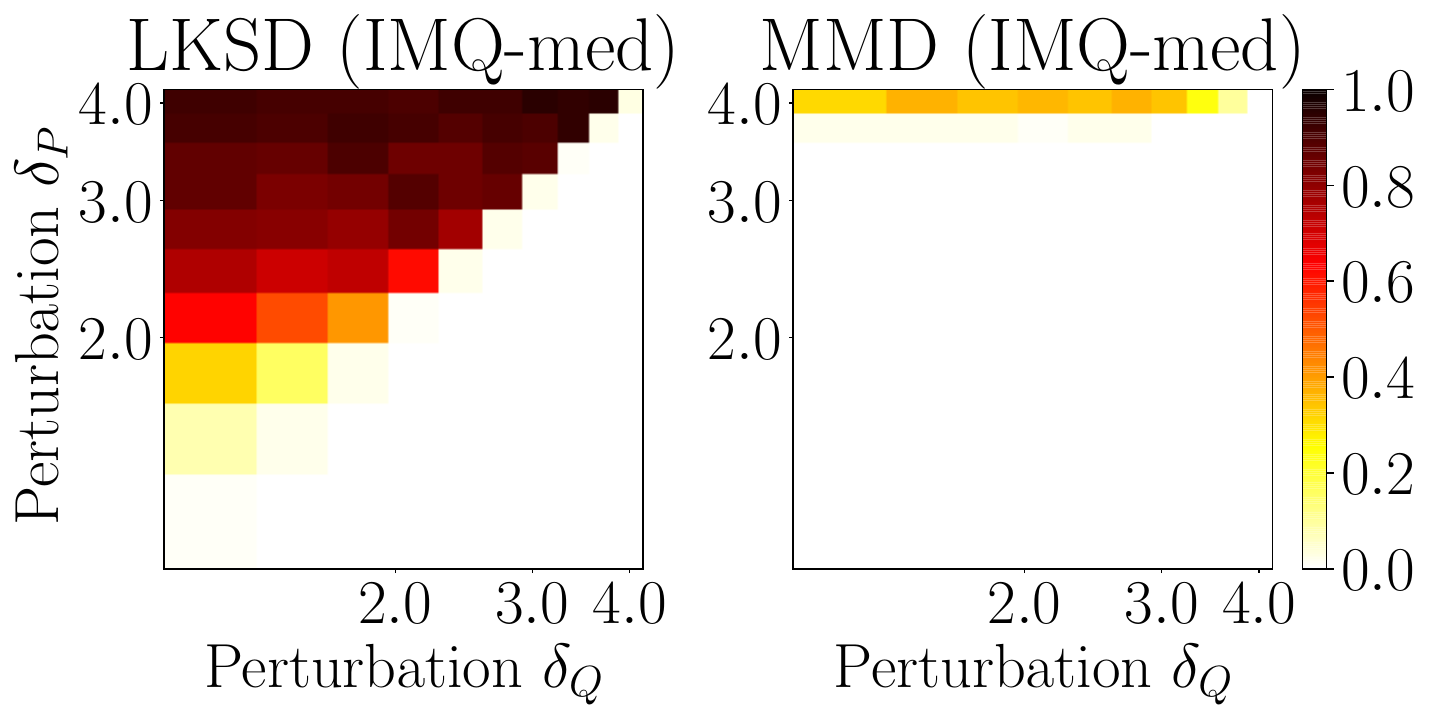}
\par\end{centering}
}\caption{Comparison between the MMD test of \citet{BouBelBlaAntGre2016}, the
proposed LKSD test using the PPCA model. The heat map represents the
estimated test power. Each result is computed on 100 trials. The significance
level $\alpha=0.05.$ }
\label{fig:ppca_vary_delta}
\end{figure}

\paragraph{Differing perturbation values}

To demonstrate the effect of perturbation parameters, we also compare
the two tests for differing configurations of $(\delta_{P},\delta_{Q}).$
We use the IMQ kernel with median scaling as in Section \ref{subsec:ppca-typeI-power}
in the main text. We choose a perturbation parameter from a grid of
size $10$ between $10^{-2}$ and $10^{1/2}$ in the $\log_{10}$
scale. Figure \ref{fig:ppca_vary_delta} demonstrates the test power
for different settings of $(\delta_{P},\delta_{Q})$ with varying
sample size $n.$ 

\subsection{Truncated Probabilistic Principal Component Analysis \label{subsec:boundedPPCA}}

As mentioned in Section \ref{subsec:Problem-setup}, our test enables
the comparison of models having likelihoods with unknown normalizing
constants. To demonstrate this feature, we consider a truncated PPCA
model. Following the notation in Section \ref{subsec:PPCA}, 
\[
p(x|z,A,\psi)\propto\varphi_{\rho}(x)\Normal{x;Az}{\psi^{2}I_{x}},\ P_{Z}=\Normal{\bfzero}{I_{z}},
\]
where $\varphi_{\rho}$ is a continuously differentiable bump function
satisfying $\varphi_{\rho}(x)=0$ if $\Verts x_{2}\geq\rho,$ and
${\cal N}(x;\mu,\Sigma)$ denotes the density of a Gaussian distribution
with mean $\mu$ and covariance $\Sigma.$ We may take smooth $\varphi_{\rho}$
such that for some $\rho_{0}>0,$ $\varphi_{\rho}(x)=1$ if $\Verts x_{2}\leq\rho_{0},$
and $0\leq\varphi_{\rho}(x)\leq1$ if $\rho_{0}<\Verts x_{2}\leq\rho$
\citep[Lemma 2.22]{Lee_2012}. We refer the reader to Appendix \eqref{subsec:bounded-domain}
for the KSD on bounded domains and the regularity condition required
for the marginal $p(x).$ 

Due to the truncation, the likelihood $p(x|z)$ is only available
up to an intractable constant $\int\varphi_{\rho}(x)\Normal{x;Az}{\psi^{2}I_{x}}\dd x$
(a function of $z$), making the latent posterior distribution doubly
intractable. Nonetheless, there are MCMC methods that circumvent the
double intractability \citep[see][for a review]{Park_2018}; these
methods typically apply if we can (approximately) sample from $p(x|z)$
for a given value of $z.$ For this truncated PPCA model, we may generate
samples from the likelihood using rejection sampling, allowing us
to use techniques such as the auxiliary variable algorithm \citep{Moeller2006}
or the exchange algorithm \citep{Murray2006}.

As in the previous section, we evaluate our test in terms of type-I
error rates and test power; we use the same notation. We set $D=100$
and $D_{z}=10,$ and use the same parameters for $R.$ We construct
a problem by perturbing the (1,1)-entry of $A.$ We compare the LKSD
and MMD tests with IMQ kernels with median- and covariance preconditioning;
these choices produced different outcomes in the previous section.
For the truncation function $\varphi_{\rho},$ we use $\rho_{0}=45$
and $\rho=50.$ We perform rejection sampling with proposal $\Normal{Az}{\psi^{2}I_{x}}.$
For some values of $z,$ we observed that the sampler never accepts
the proposal. We therefore decide to generate $300$ proposals; if
no sample has been accepted, we normalize a proposal $x$ such that
$\Verts x_{2}=0.99\rho_{0}$ and use it as a sample. Finally, we use
the exchange algorithm of \citet{Murray2006} for the LKSD test. We
employ the same proposal as the MALA sampler with a step size of $10^{-3},$
and use $m=2,000$ samples after $t=500$ burn-in steps. Although
the inexact rejection sampling induces bias to the Monte Carlo estimate,
we will see that this bias is negligible in terms of the size control. 

\paragraph*{Problem 1 (null)}

As in the previous section, we create a null scenario by setting $(\delta_{P},\delta_{Q})=(1,1+10^{-5}).$
We consider significance levels $\alpha=0.01,0.05.$ Table \ref{tab:tppca_type1}
reports the finite-sample size of the two tests for significance level
$\alpha=0.05.$ Again, the result for $\alpha=0.01$ is omitted as
no tests rejected the null hypotheses. It can be seen that the size
of both tests are controlled. 

\begin{table}[H]
\caption{Type-I errors the MMD test of \citet{BouBelBlaAntGre2016}, the proposed
LKSD test in Problem 1 for truncated PPCA models. Rejection rates
are computed on 300 trials with significance level $\alpha=0.05.$
The columns IMQ-med and IMQ-cov denote IMQ kernels with median- and
covariance preconditioning, respectively.}

\label{tab:tppca_type1}\centering\begin{threeparttable}
\begin{centering}
\begin{tabular}{rrr|rr}
\hline 
\headcell  \theadmd{Sample size $n$} & \multicolumn{4}{c}{\headcell  \theadmd{Rejection rates}}\tabularnewline
 & \multicolumn{2}{c|}{IMQ-med} & \multicolumn{2}{c}{IMQ-cov}\tabularnewline
 & \multicolumn{1}{l}{MMD} & \multicolumn{1}{l|}{LKSD} & \multicolumn{1}{l}{MMD} & \multicolumn{1}{l}{LKSD}\tabularnewline
$100$ & 0.003 & 0.000 & 0.003 & 0.003\tabularnewline
$200$ & 0.003 & 0.010 & 0.017 & 0.003\tabularnewline
$300$ & 0.000 & 0.000 & 0.013 & 0.010\tabularnewline
400 & 0.000 & 0.007 & 0.007 & 0.003\tabularnewline
$500$ & 0.000 & 0.010 & 0.007 & 0.010\tabularnewline
\hline 
\end{tabular}
\par\end{centering}
\end{threeparttable}
\end{table}

\paragraph*{Problem 2 (alternative)}

\begin{figure}[H]

\begin{centering}
\subfloat[\label{fig:tppca_h1_p2_q1_imq_med}: IMQ kernel with median scaling.]{\begin{centering}
\includegraphics[width=6.5cm]{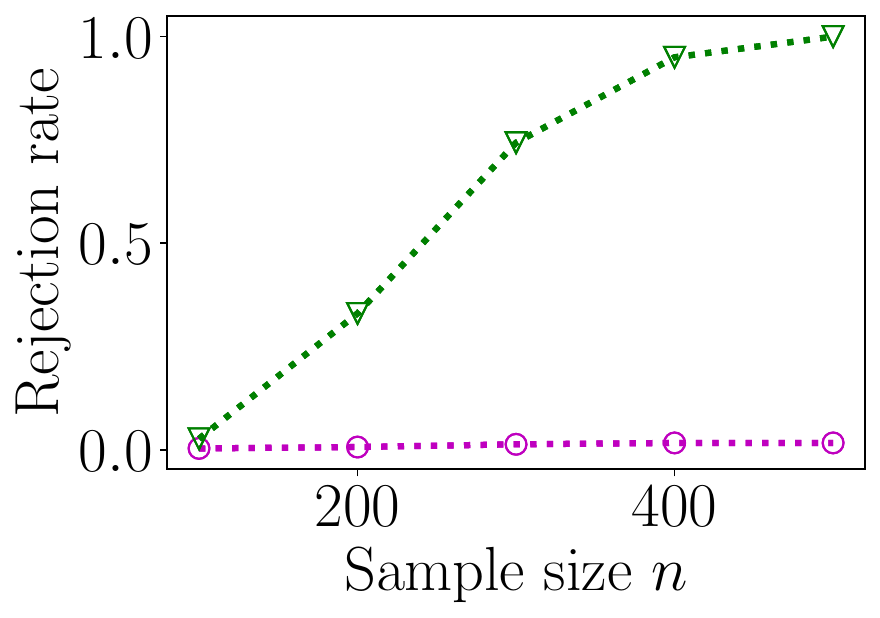}
\par\end{centering}
\centering{}}\hfill{}\subfloat[\label{fig:tppca_h1_p2_q1_IMQ-cov} IMQ kernel with covariance preconditioning.]{\begin{centering}
\includegraphics[width=6.5cm]{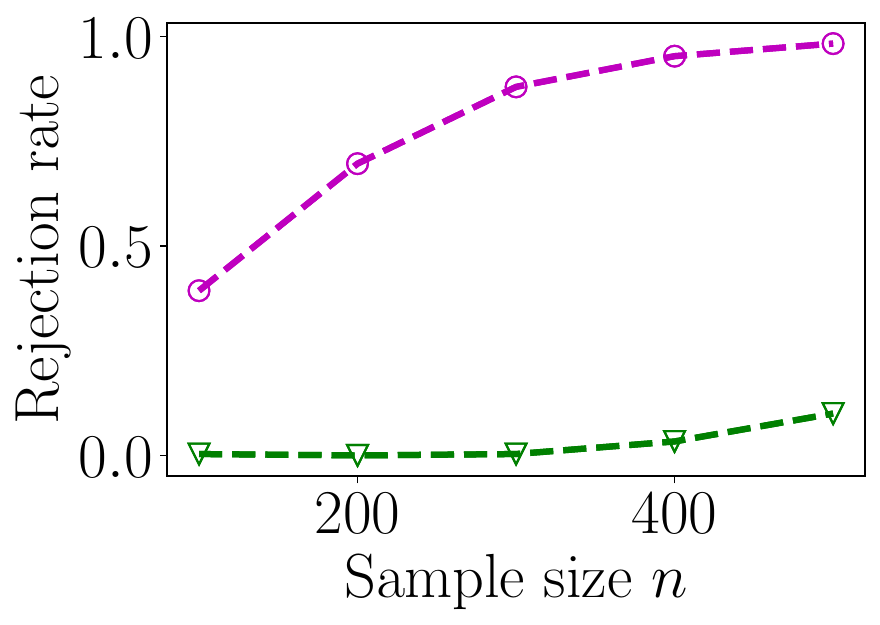}
\par\end{centering}
}
\par\end{centering}
\begin{centering}
\par\end{centering}
\caption{Power curves of the MMD test and the proposed LKSD
test in Problem 2 for truncated PPCA models. Dotted lines indicate
tests with the IMQ kernel with the median preconditioning; dashed
lines the covariance preconditioning. The perturbation parameters
are set as $(\delta_{P},\delta_{Q})=(2,1).$ Each result is computed
on 300 trials. The significance level $\alpha=0.05.$ TheMarkers:
\textifsymbol[ifgeo]{51} (the LKSD test); \textbigcircle{} (the relative
MMD test).}
\label{fig:tppca-h1-p2-q1}
\end{figure}

We next compare the power of the two tests. We construct an alternative
scenario using perturbation parameters $\delta_{P}=2$ for $P$ and
$\delta_{Q}=1$ for $Q.$ Figure \ref{fig:tppca-h1-p2-q1} shows the
estimated power curves. We observe a trend similar to the PPCA experiment:
the LKSD test with median preconditioning forms the MMD counterpart,
whereas the converse relation holds for covariance preconditioning.
The experiment shows that the LKSD test can also complement the MMD
test for models without intractable likelihood functions. 

\subsection{LDA \label{subsec:LDA-additional}}

\subsubsection{Type-I error and test power}

\paragraph{Differing perturbation values}

\begin{figure}[H]
\centering{}\subfloat[$n=100$]{\begin{centering}
\includegraphics[scale=0.28]{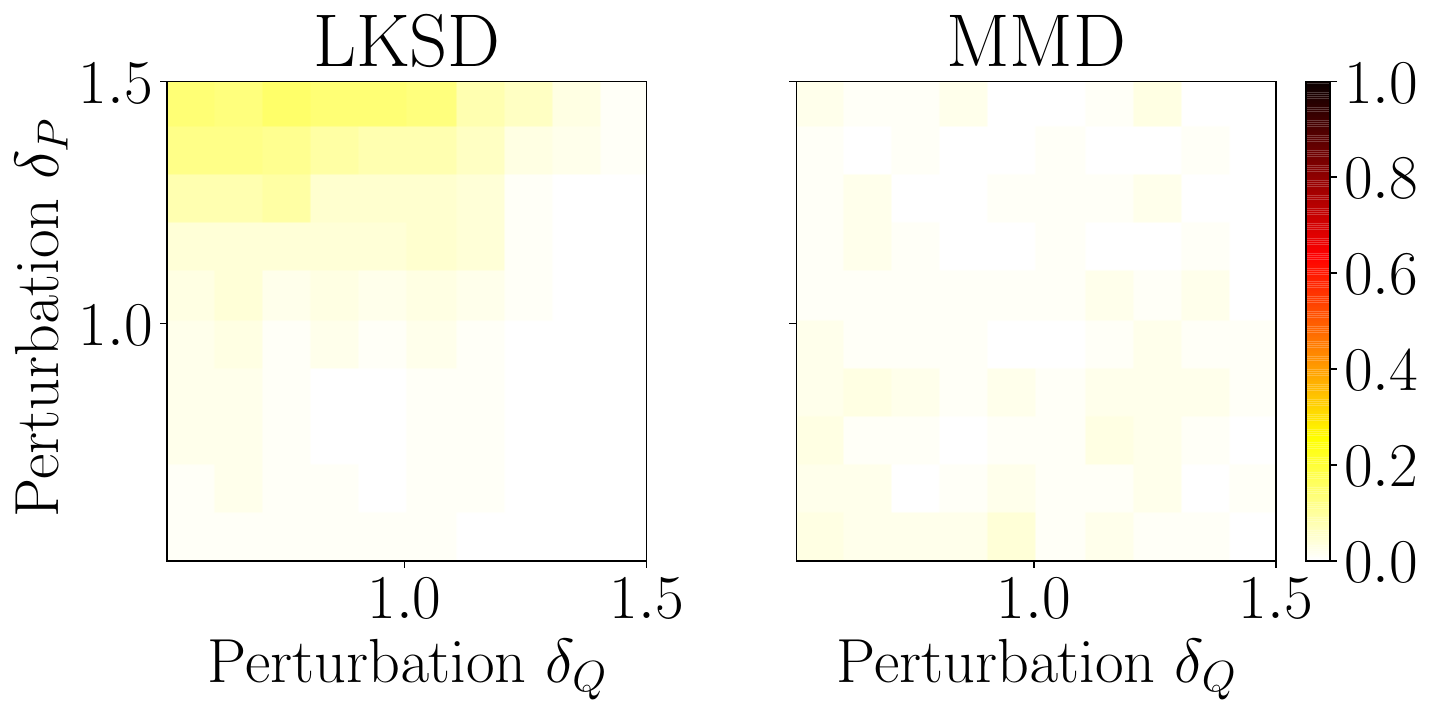}
\par\end{centering}
\centering{}}\hfill{}\subfloat[$n=200$]{\begin{centering}
\includegraphics[scale=0.28]{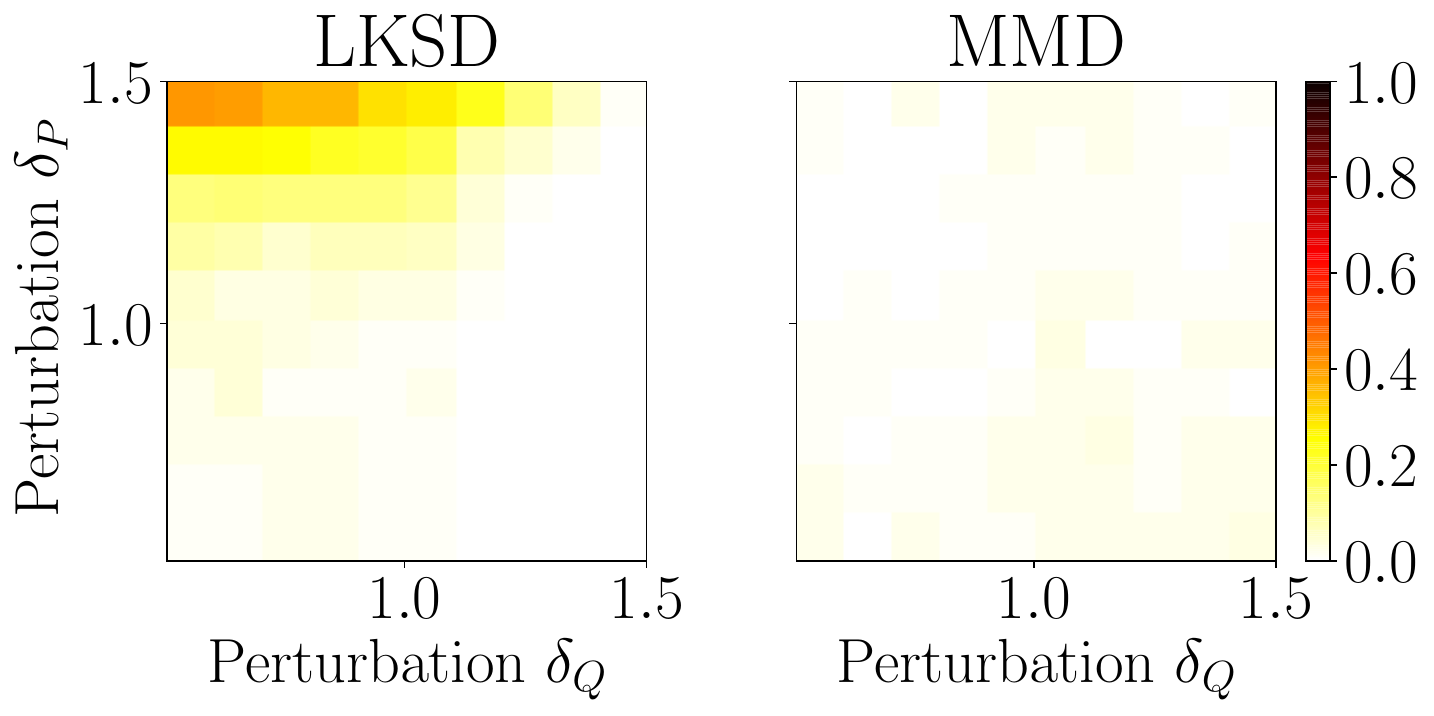}
\par\end{centering}
}\\
\subfloat[$n=300$]{\begin{centering}
\includegraphics[scale=0.28]{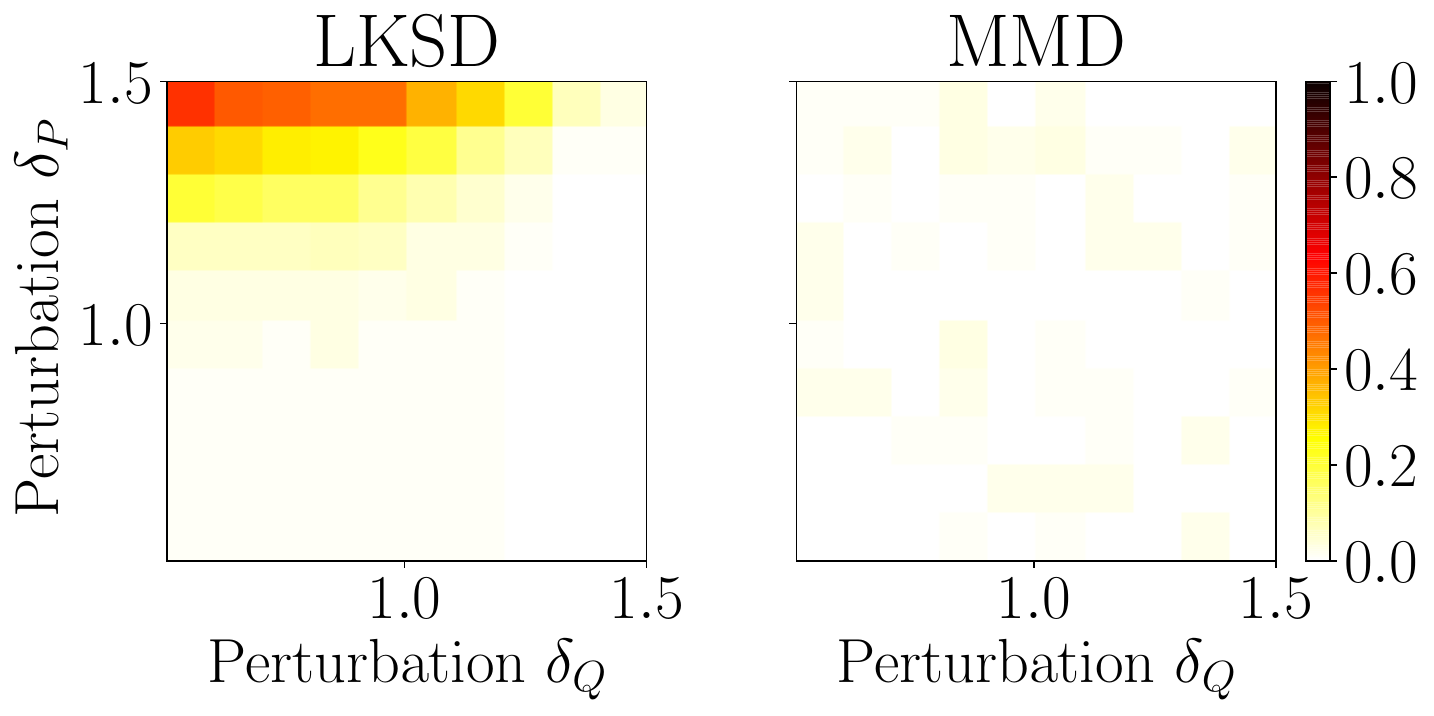}
\par\end{centering}
}\caption{Comparison between the MMD test of \citet{BouBelBlaAntGre2016}, the
proposed LKSD test using the LDA model. The heat map represents the
estimated test power. Each result is computed on 100 trials. The significance
level $\alpha=0.05.$}
\label{fig:lda_vary_delta}
\end{figure}

As in the previous section, we compare the LKSD and MMD tests using
differing configurations of $(\delta_{P},\delta_{Q}).$ All the model
configurations follow Section \ref{subsec:lda-prior-ptb} in the main
text, and we use the IMQ-BoW kernel. We choose a perturbation parameter
from a grid of size $10$ between $10^{-2}$ and $1$ in the $\log_{10}$
scale. Figure \ref{fig:lda_vary_delta} demonstrates the test power
for different settings of $(\delta_{P},\delta_{Q})$ with varying
sample size $n.$ 

\paragraph*{LDA models with more sparse topics}

We run the same experiment as in Section \ref{subsec:expLDA} except
that the topics $b$ are made sparse by sampling from the Dirichlet
distribution with all the concentration parameters $0.1.$ The results
are summarized in Tables \ref{tab:lda-prob1-sparse} and \ref{tab:lda-prob2-sparse}.
The LKSD test underperforms the MMD test in this case. As the topics
are more sparse, generated documents tend to have words from a particular
topic; this trend escalates as we increase the concentration parameter
of the topic proportion prior. Models thus observe compositions of
words that they would not generate, resulting in a high-variance test
statistic for the same reason as in Section \ref{subsec:lda-topic=002013ptb}
(In fact, the topics have probabilities as low as $10^{-40}$, which
comes close to violating the assumption on the density). 

\begin{table}[H]
\caption{Type-I error for LDA Problem 1 $(\delta_{P},\delta_{Q})=(0.5,0.6)$}
\label{tab:lda-prob1-sparse}
\begin{centering}
\begin{tabular}{rrr|rr}
\hline 
\headcell  \theadmd{Sample size $n$} & \multicolumn{4}{c}{\headcell  \theadmd{Rejection rates}}\tabularnewline
 & \multicolumn{2}{c|}{EQ BoW} & \multicolumn{2}{c}{IMQ BoW}\tabularnewline
 & \multicolumn{1}{l}{MMD} & \multicolumn{1}{l|}{LKSD} & \multicolumn{1}{l}{MMD} & \multicolumn{1}{l}{LKSD}\tabularnewline
$100$ & 0.007 & 0.013 & 0.007 & 0.013\tabularnewline
$200$ & 0.007 & 0.007 & 0.003 & 0.007\tabularnewline
$300$ & 0.007 & 0.017 & 0.000 & 0.017\tabularnewline
$400$ & 0.013 & 0.017 & 0.003 & 0.020\tabularnewline
$500$ & 0.020 & 0.010 & 0.003 & 0.010\tabularnewline
\hline 
\end{tabular}
\par\end{centering}
\end{table}

\begin{table}[H]
\caption{Power estimates for LDA Problem 2 $(\delta_{P},\delta_{Q})=(1.1,0.6)$}
\label{tab:lda-prob2-sparse}
\centering{}%
\begin{tabular}{rrrrrrrrr}
\hline 
\headcell  \theadmd{Sample size $n$} & \multicolumn{8}{c}{\headcell  \theadmd{Rejection rates}}\tabularnewline
 & \multicolumn{4}{c}{$\alpha=0.01$} & \multicolumn{4}{c}{$\alpha=0.05$}\tabularnewline
 & \multicolumn{2}{c}{EQ BoW} & \multicolumn{2}{c}{IMQ BoW} & \multicolumn{2}{c}{EQ BoW} & \multicolumn{2}{c}{IMQ BoW}\tabularnewline
 & MMD & LKSD & MMD & LKSD & MMD & LKSD & MMD & LKSD\tabularnewline
$100$ & 0.000 & 0.003 & 0.023 & 0.003 & 0.013 & 0.040 & 0.083 & 0.037\tabularnewline
$200$ & 0.000 & 0.000 & 0.030 & 0.000 & 0.017 & 0.040 & 0.170 & 0.043\tabularnewline
$300$ & 0.003 & 0.013 & 0.047 & 0.013 & 0.010 & 0.053 & 0.290 & 0.053\tabularnewline
$400$ & 0.003 & 0.007 & 0.093 & 0.007 & 0.013 & 0.080 & 0.373 & 0.077\tabularnewline
$500$ & 0.000 & 0.000 & 0.146 & 0.000 & 0.000 & 0.050 & 0.477 & 0.050\tabularnewline
\hline 
\end{tabular}
\end{table}

\subsubsection{Kernel parameter \label{subsec:lda-kernel-param}}

As in the PPCA experiment, we investigate the performance dependence
on the kernel choice. Using LDA problem 2 , we examine how the test
power is affected by the scaling parameter. We use the EQ and IMQ
BoW kernels as above, and choose their scaling parameter $\lambda^{2}$
from $\{10^{-6},10^{-5},\dots,10^{3}\}.$ For each $n\in\{100,300\}$
we run 300 trials and estimate the test power of the LKSD and MMD
tests. Figure \ref{fig:kernel_param_power} plots the power curves
of the tests. We can see that the MMD test fails for any choice of
the kernel. For the LKSD test, the IMQ kernel has a flat curve, indicating
its independence from the bandwidth (at least in this candidate range),
whereas the EQ kernel benefits from a small bandwidth value. 

\begin{figure}[H]
\subfloat[$n=100.$]{\centering{}\includegraphics[width=6cm]{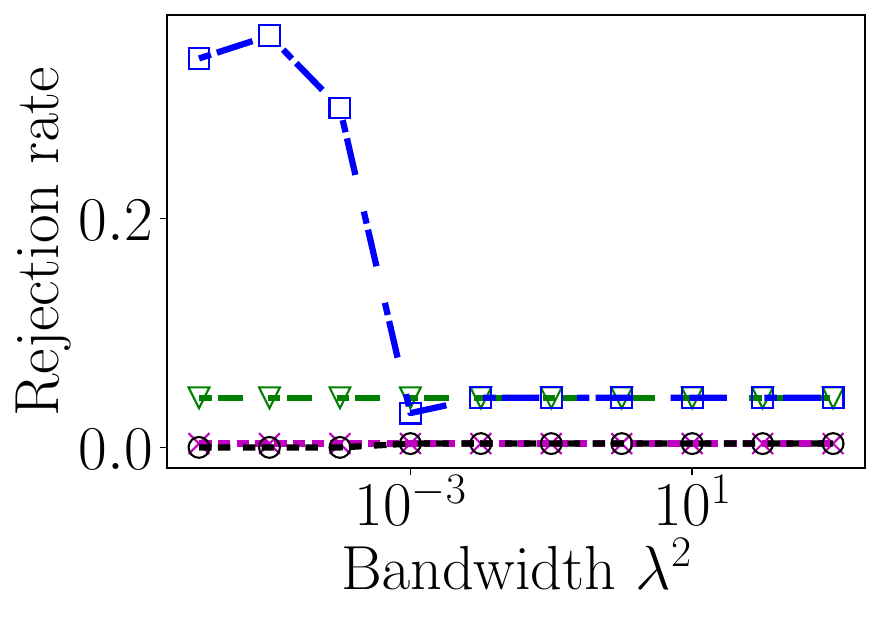}}\hfill{}\subfloat[$n=300.$]{\centering{}\includegraphics[width=6cm]{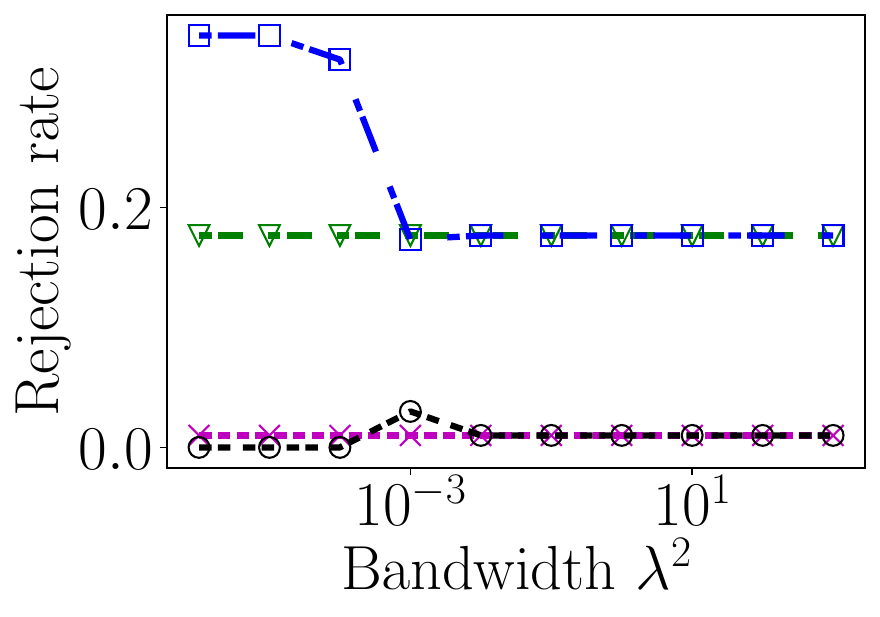}}

\caption{Power curves of the proposed LKSD test and the MMD test in LDA Problem
2. The perturbation parameters are set as $(\delta_{P},\delta_{Q}=2,1).$
Each result is computed on 300 trials. The significance level $\alpha=0.05.$
Markers: \textifsymbol[ifgeo]{51} (LKSD test with IMQ kernel); \textifsymbol[ifgeo]{48}
(LKSD test with EQ kernel); \textbigcircle{} (MMD test with IMQ kernel);
$\times$ (MMD test with EQ kernel).}

\end{figure}

\subsection{Experiment: close models and type-I errors\label{subsec:exp-closemodels}}

We investigate the behavior of the LKSD test when two models are close
to each other. In this case, the difference of the U-statistic kernels
defined by the models is small, which could therefore lead to the
degeneracy of the U-statistic; i.e., the normal approximation of the
test statistic is not appropriate. In the following, we investigate
the three variants of the LKSD test defined by different choices of
the variance estimator. We compare the jackknife estimator \eqref{eq:sigma-nt-definition}
with the following two estimators: (i) a U-statistic variance estimator
where $\zeta_{1}$ and $\zeta_{2}$ in \eqref{eq:ustat-var-decomposition}
are estimated by U-statistics, and (ii) a V-statistic variance estimator
where $\zeta_{1}$ is estimated by a V-statistic. The U-statistic
estimation was considered by \citet{BouBelBlaAntGre2016} and \citet{JitKanPatHayetal2018}.
 The issue with the U-statistic estimator is that it underestimates
the actual variance. In fact, we observed that the variance estimator
sometimes returns \textbf{negative} values. This can occur since the
statistic is given as a difference between unbiased estimates of quantities
close to zero (this issue applies to the V-statistic estimator). We
made the (arbitrary) choice to accept the null hypothesis when the
variance estimate was negative to avoid false rejections. For this
reason, we recommend against using the U-statistic estimator. 

\subsubsection{PPCA }

Our first experiment concerns PPCA models. Specifically, we choose
a PPCA model for the data distribution as in Section \ref{subsec:PPCA}
with $D=50.$ The difference is that we fix the perturbation parameter
$\delta_{P}$ for $P$ at $1,$ and vary the parameter $\delta_{Q}$
by choosing it from $\{10^{-i}:i\in\{2,3,\dots,9\}\}$ (this choice
yields null $H_{0}$ scenarios). We set the significance level $\alpha=0.05.$
For each $n\in\{100,200,300\},$ we run the tests for 300 trials and
examine the behavior of the tests under the null. 

\begin{figure}[H]
\begin{centering}
\par\end{centering}
\centering{}\subfloat[$n=100$]{\centering{}\includegraphics[scale=0.28]{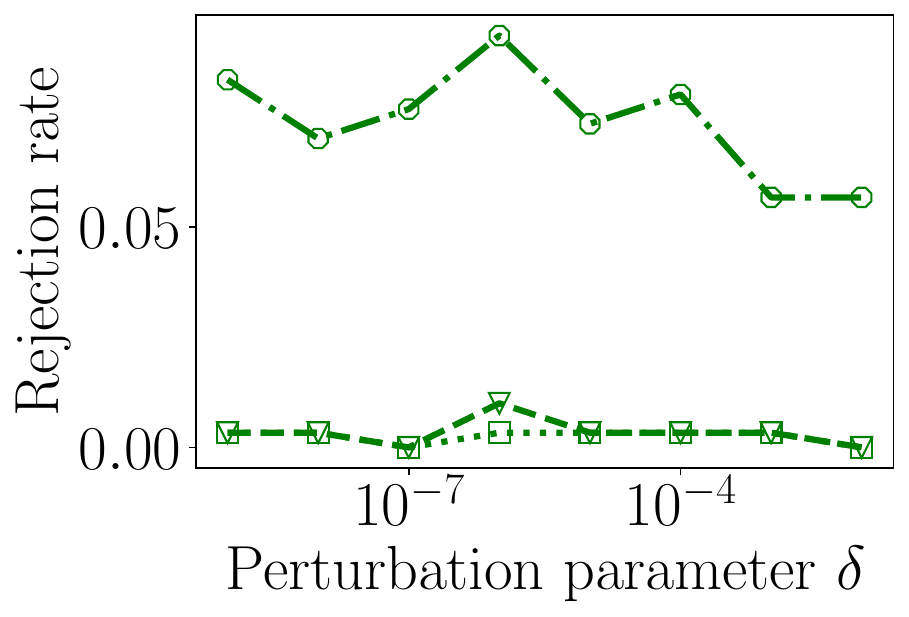}}\hfill{}\subfloat[$n=200$]{\centering{}\includegraphics[scale=0.28]{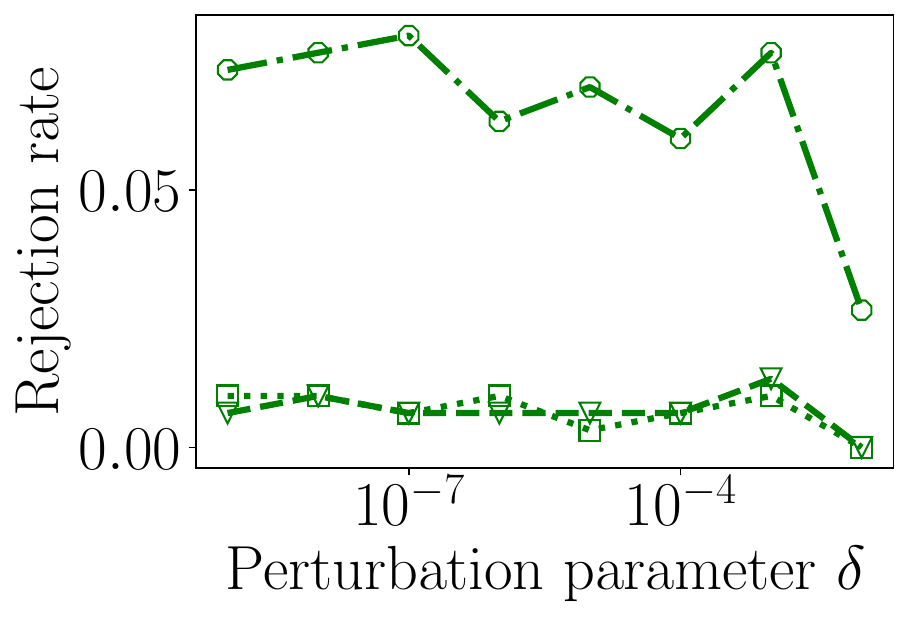}}\hfill{}\subfloat[$n=300$]{\centering{}\includegraphics[scale=0.28]{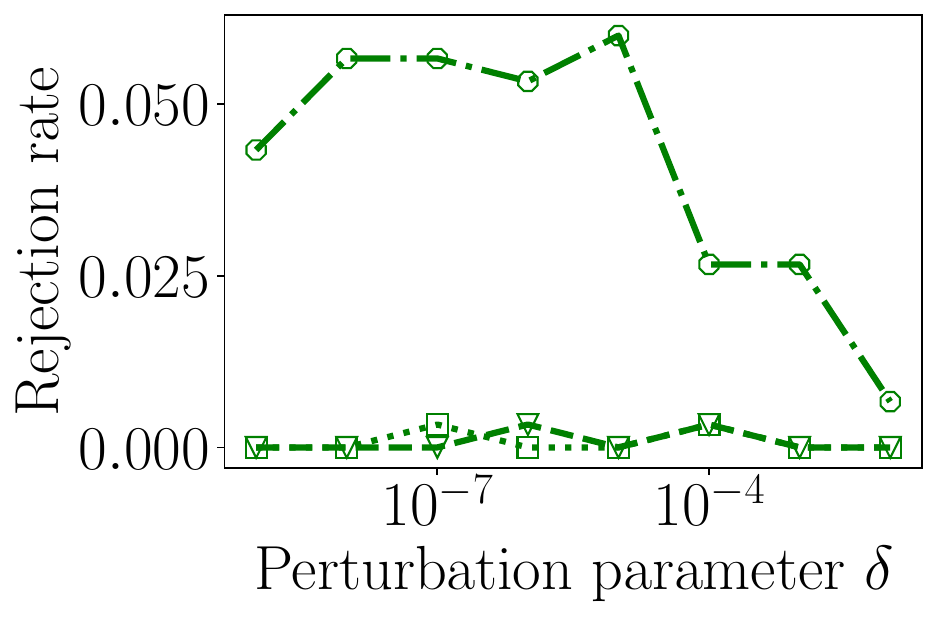}}\caption{The behaviors of the two LKSD tests under the null. The nominal level
$\alpha$ is set to $0.05.$ The test with the U-statistic variance
estimator has higher type-I errors as the models get closer to each
other. Markers: \textifsymbol[ifgeo]{51} (LKSD test with the jackknife
variance estimator); \textbigcircle{} (LKSD test with the U-statistic
variance estimator); \textifsymbol[ifgeo]{48} (LKSD test with the
V-statistic variance estimator).}
\label{fig:PPCA-close-p1}
\end{figure}

Figure \ref{fig:PPCA-close-p1} shows the tests' rejection rates.
We first note that as the perturbation parameter decays (the models
get closer to each other), the test with the U-statistic estimator
rejects more and has higher type-I errors than the nominal level $\alpha=0.05.$
These plots demonstrate that the jackknife and V-statistic versions
of the test are more robust in this setting. 

\pagebreak{}

\subsubsection{LDA}

We conduct a similar experiment with LDA models. The problem setup
is the same as in Section \ref{subsec:expLDA}, except that the vocabulary
size $L=100.$ We perturb the sparsity parameter of the Dirichlet
prior of an LDA model. We set $\delta_{P}=1$ and $\delta_{Q}=1+\delta,$
where $\delta$ is chosen from $\{10^{-2i}:i\in\{1,\dots,5\}\}.$
For each $n\in\{100,200,300\},$ we run the tests for 300 trials with
significance level $\alpha=0.05.$

\begin{figure}[H]
\begin{centering}
\par\end{centering}
\centering{}\subfloat[$n=100$]{\centering{}\includegraphics[scale=0.28]{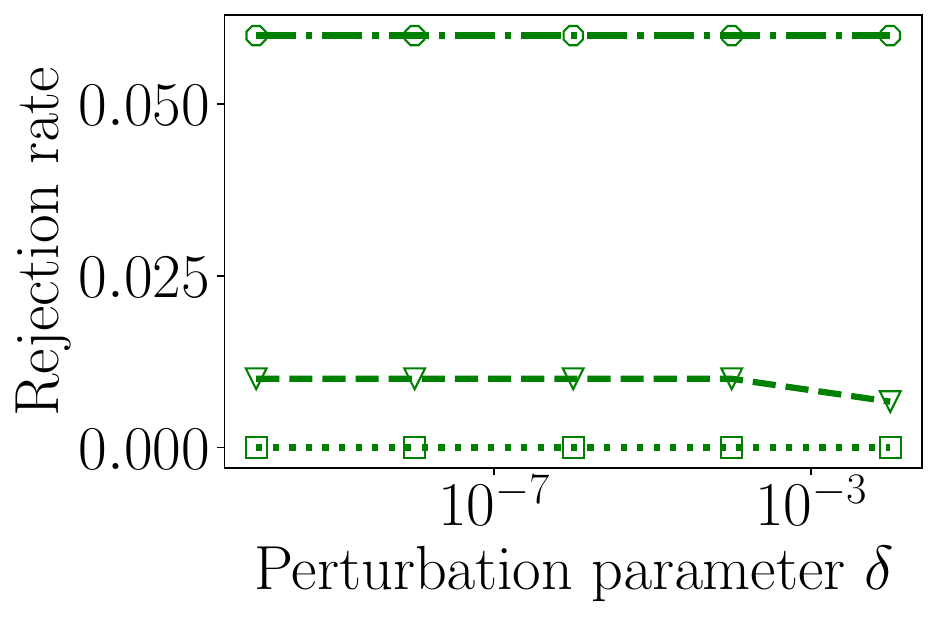}}\hfill{}\subfloat[$n=200$]{\centering{}\includegraphics[scale=0.28]{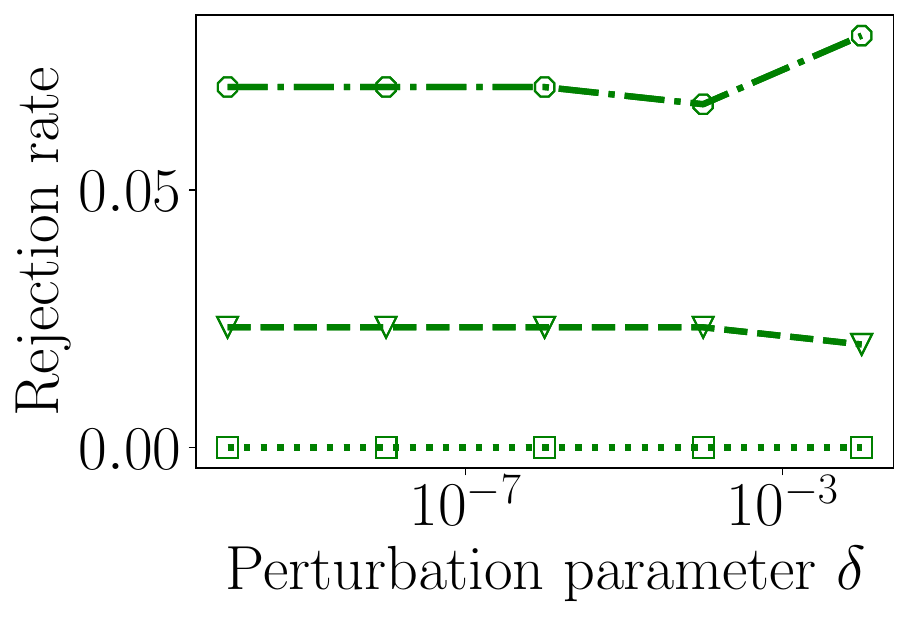}}\hfill{}\subfloat[$n=300$]{\centering{}\includegraphics[scale=0.28]{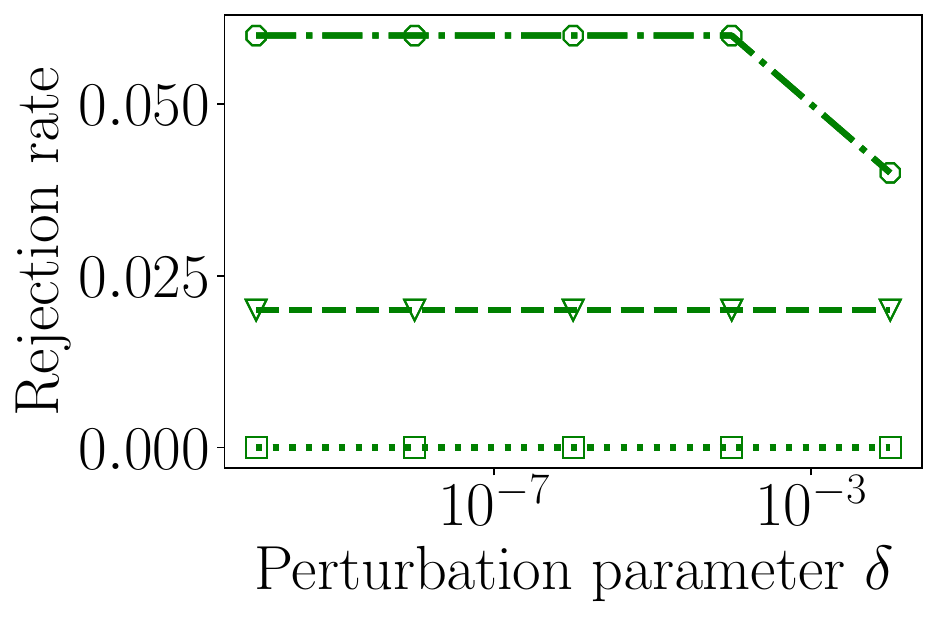}}\caption{The behaviors of the two LKSD tests under the null. The nominal level
$\alpha$ is set to $0.05.$ The test with the U-statistic variance
estimator has higher type-I errors as the models get closer to each
other. Markers: \textifsymbol[ifgeo]{51} (LKSD test with the jackknife
variance estimator); \textbigcircle{} (LKSD test with the U-statistic
variance estimator); \textifsymbol[ifgeo]{48} (LKSD test with the
V-statistic variance estimator).}
\label{fig:LDA-close-p1}
\end{figure}

Figure \ref{fig:LDA-close-p1} shows the rejection rate of each test.
The test with the V-statistic estimator is more conservative than
the other tests. The U-statistic variance appears to underestimate
the actual variance. 

\subsection{Experiment: identical models \label{subsec:Experiment:-identical-models}}

\begin{figure}[H]
\begin{centering}
\par\end{centering}
\centering{}\subfloat[Identical PPCA models. ]{\begin{centering}
\includegraphics[width=6cm]{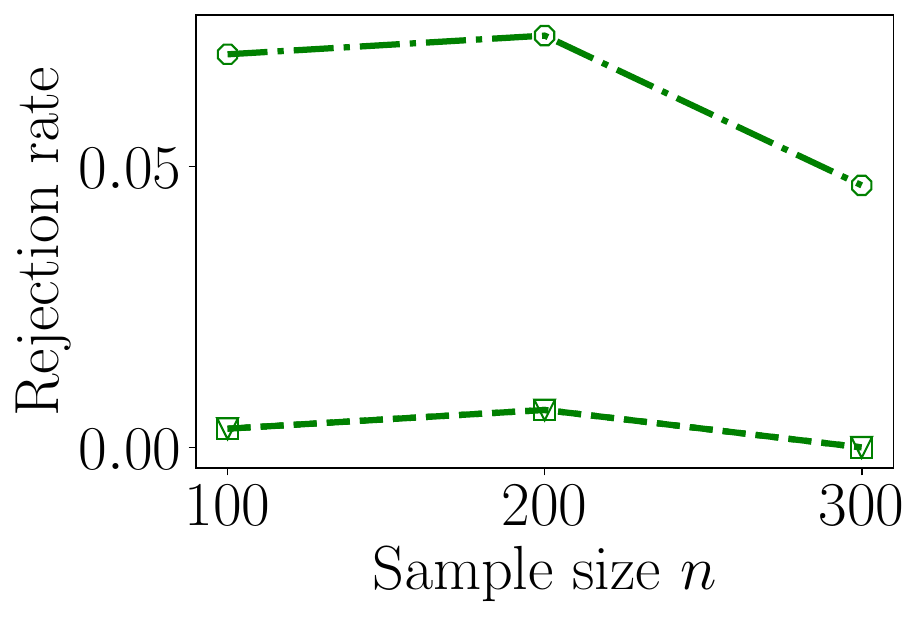}
\par\end{centering}
}\hfill{}\subfloat[Identical LDA models. ]{\begin{centering}
\includegraphics[width=6cm]{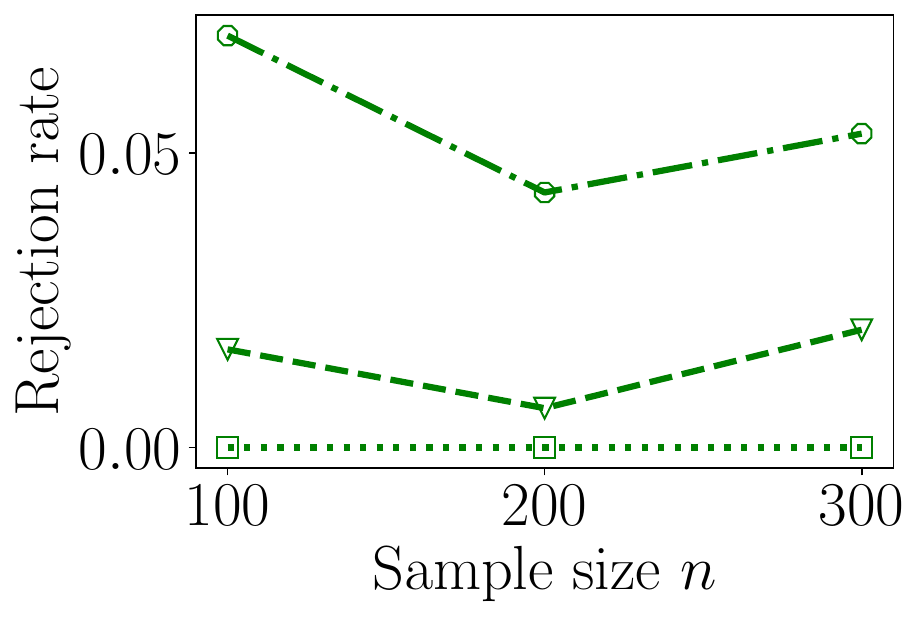}
\par\end{centering}
}\caption{Plots of type-I errors when two models are identical. Markers: \textifsymbol[ifgeo]{51}
(LKSD test with the jackknife variance estimator); \textbigcircle{}
(LKSD test with the U-statistic variance estimator); \textifsymbol[ifgeo]{48}
(LKSD test with the V-statistic variance estimator). The LKSD test
with the U-statistic variance estimator has higher errors than the
nominal level $\alpha=0.05.$}
\label{fig:identical-H0}
\end{figure}

We look into the behaviors of the LKSD test when the models are identical.
Our test procedure provides no guarantee in this case, as the asymptotic
distribution would deviate from the normal distribution. As in the
previous section, we compare the performance of the tests with the
three proposed variance estimators. In the following, we fix the significance
level $\alpha$ at $0.05.$ As in Sections \ref{subsec:PPCA}, \ref{subsec:expLDA},
we choose perturbation parameters for the candidate models, and run
300 trials for a differing sample size $n\in\{100,200,300\}.$ For
both PPCA and LDA models, we choose $\delta_{P}=\delta_{Q}=1.$ Figure
\ref{fig:identical-H0} shows the plot for each problem. We note that
the U-statistic test has higher type-I error rates in this setting,
although they are closed to the design level. Notwithstanding that
our test assumptions are violated, the jackknife and V-statistic approaches
reject the null at a rate well below $0.05,$ and remain conservative
in this example. 

